\documentclass{sigchi}

\RequirePackage{snapshot}

\toappear{}

\usepackage{graphics} 
\usepackage[T1]{fontenc}
\usepackage{txfonts}
\usepackage{mathptmx}
\usepackage[pdfpagelabels=false]{hyperref}
\usepackage{color}
\usepackage[usenames,dvipsnames,table]{xcolor}
\usepackage{booktabs}
\usepackage{textcomp}
\usepackage[all]{hypcap}  
\usepackage{ccicons}  

\usepackage{morefloats}
\usepackage{gensymb}
\usepackage{subcaption}
\usepackage{amsmath}
\usepackage{amsfonts}
\usepackage{amssymb}
\newcommand{\secref}[1]{the ``\nameref{sec:#1}'' section}
\newcommand{\figref}[1]{Fig.~\ref{fig:#1}}
\newcommand{\tabref}[1]{Table~\ref{tab:#1}}

\def\plaintitle{What Can Be Predicted from\\Six Seconds of Driver Glances?}

\def\emptyauthor{}
\def\plainkeywords{Gaze patterns; driver state prediction; naturalistic on-road study; hidden Markov models.}

\makeatletter
\def\url@leostyle{%
  \@ifundefined{selectfont}{
    \def\UrlFont{\sf}
  }{
    \def\UrlFont{\small\bf\ttfamily}
  }}
\makeatother
\def\pprw{8.5in}
\def\pprh{11in}

\definecolor{linkColor}{RGB}{6,125,233}
\hypersetup{%
  pdftitle={\plaintitle},
  pdfauthor={\emptyauthor},
  pdfkeywords={\plainkeywords},
  bookmarksnumbered,
  pdfstartview={FitH},
  colorlinks,
  citecolor=black,
  filecolor=black,
  linkcolor=black,
  urlcolor=linkColor,
  breaklinks=true,
}

\begin{document}

\title{\plaintitle}

\newcommand{\authorspace}{\hspace{0.2in}}
\author{
Lex Fridman$^1$ \authorspace
Heishiro Toyoda$^2$ \authorspace
Sean Seaman$^3$ \authorspace
Bobbie Seppelt$^3$ \authorspace\\
Linda Angell$^3$ \authorspace
Joonbum Lee$^1$ \authorspace
Bruce Mehler$^1$ \authorspace
Bryan Reimer$^1$\\
\affaddr{$^1$ Massachusetts Institute of Technology}\\
\affaddr{$^2$ Toyota Collaborative Safety Research Center}\\
\affaddr{$^3$ Touchstone Evaluations}
}


\maketitle


\begin{abstract}
  We consider a large dataset of real-world, on-road driving from a 100-car naturalistic study to explore the predictive
  power of driver glances and, specifically, to answer the following question: what can be predicted about the state of
  the driver and the state of the driving environment from a 6-second sequence of macro-glances? The context-based
  nature of such glances allows for application of supervised learning to the problem of vision-based gaze estimation,
  making it robust, accurate, and reliable in messy, real-world conditions. So, it's valuable to ask whether such
  macro-glances can be used to infer behavioral, environmental, and demographic variables? We analyze 27 binary
  classification problems based on these variables. The takeaway is that glance can be used as part of a multi-sensor
  real-time system to predict radio-tuning, fatigue state, failure to signal, talking, and several environment
  variables.
\end{abstract}



\section{Introduction}\label{sec:introduction}

As the level of vehicle automation continues to increase, the car is more and more becoming a multi-sensor computational
system tasked with understanding (1) the state of the driver \cite{yang2010driver} and (2) the state of the driving
environment \cite{mccall2006video}. From a computer vision perspective, both of these tasks have over a decade of active
research that proposes various methods for robust, real-time processing of inward-facing and outward-facing video to
extract actionable knowledge with which the car can assist the driver \cite{sivaraman2013looking}. Driver gaze
classification is one of the more successful recent outcomes of these computer vision efforts promising above 90\% gaze region
classification accuracy \cite{fridman2016owllizard} in the wild when simplifying the general gaze estimation problem by
considering broad segmentation of gaze base on attention allocation semantics: forward roadway, left mirror, right
mirror, rearview mirror, instrument cluster, center stack, and other regions. The promise of accurate real-time gaze
classification is what motivates the question posed in this work: once the system infers gaze region from video, what
can we predict about the state of the driver and the state of the environment? Put another way, a gaze classification
system can be seen as one of several sensors available in the vehicle, and so it is valuable to investigate what
actionable information can be inferred from this sensor in order to design a better interface between human and machine
in the driving context.

The driving task and the driving environment places the driver under a wide array of physical and cognitive
demands. Intuitively, visual demand can be predicted from gaze \cite{reimer2016multi}. However, glance allocation
strategies provide a window through which we can predict the more general mental and physical state of the driver
outside of just where they are looking (i.e., activity \cite{maciej2009comparison}, inattention \cite{d2007visual},
fatigue \cite{khushaba2011driver}). The possibility of inference about aspects of the external driving environment based
on macro-glances is an open question for which this paper provides promising results. In this paper, ``macro-glances''
refer to the discretization of driver gaze (see \secref{macro-micro-glances}).

The generalizability of our exploratory look at what can and cannot be predicted from driver macro-glances relies
inextricably on the characteristics of the dataset. We use 4,816 annotated six-second epochs of baseline driving from the
100-Car Naturalistic Driving Study database \cite{dingus2006100}. This dataset includes approximately 2,000,000 vehicle
miles, almost 43,000 hours of data, 241 primary and secondary drivers, 12 to 13 months of data collection for each
vehicle, and data from a highly capable instrumentation system including five channels of video and vehicle
kinematics. This data contains many extreme cases of driving behavior and performance, including severe fatigue,
impairment, judgment error, risk taking, willingness to engage in secondary tasks, aggressive driving, and traffic
violations. Therefore, we believe that conclusions derived from this data are applicable to general real-world driving.

The contribution, novelty, and validity of this work can be summarized most briefly as follows:

\begin{itemize}
\item \textbf{Contribution:} Show that just 6 seconds of coarse driver gaze regions can be used to predict a lot of
  things about the driver, the car, and the driving environment. This helps (1) provide a greater understanding of the
  ``human'' in human-to-vehicle interaction and (2) pave the way for a real-time HMI system in the car based on driver
  gaze that is robust to challenging real-world conditions. See \secref{framework} below.
\item \textbf{Novelty:} Drive gaze has been used to predict attention allocation, but not to predict everything else. We
  try to do just that for the first time and show when it works and when it doesn't.
\item \textbf{Validity:} The results are based on a large naturalistic on-road study with little to no constraints on
  the participants, so the data is representative of the general population and is extensive enough to provide a
  high degree of generalizability.
\end{itemize}

\section{Related Work}\label{sec:related-work}

The 100-Car Naturalistic Driving Study dataset has been extensively used to analyze various aspects of driver behavior
in the wild \cite{dingus2006100}. Much of the focus has been on the crashes and near-crashes in the data, and describing
the factors that lead to these crashes \cite{lord2010statistical} especially with regard to the long glances away from
the road \cite{liang2012dangerous}. We focus instead on the baseline driving epochs which are more representative of the
variability of driver behavior and driving environment.

\subsection{Macro-Glances and Micro-Glances}\label{sec:macro-micro-glances}

We define the terms ``macro-glances'' and ``micro-glances'' to help specify the distinction between context-dependent
and context-independent allocations of gaze: 

\begin{itemize}
\item \textbf{Micro-Glances:} Context-independent gaze allocation achieved by fixational eye movement (i.e., saccades)
  and changes in head orientation. The target ``location'' of micro-glances is defined by the exact 3D coordinates of
  the fixation point. Example: driver looking at a stop sign.
\item \textbf{Macro-Glances:} Gaze allocation categorized into discrete regions that are defined by the context. The target
  ``location'' of macro-glances is one of these pre-defined regions and not the exact 3D coordinates of the fixation
  point. Example: driver looking at the forward roadway.
\end{itemize}

In this work, we analyze the sequence of driver macro-glances which contains both spatial and temporal information. In
particular, the temporal characteristics of the transition between glance regions is the main feature being
utilized. This is in contrast to the traditional method of measuring driver state, such as measuring the total time
visual attention is directed away from the forward roadway \cite{national2012visual} or to elements specific to the
operation of the HMI \cite{driver2006statement}.  While these measures are intuitive and have some demonstrated utility,
a shortcoming common to both approaches is that they aggregate behavioral information over a given time span to a single
number, disregarding temporal dynamics that has to be considered in making predictions about the driver's physical and
mental state. A good example of where temporal information is very important but has not been investigated as much as
the aggregate measure is in using blink for fatigue classification \cite{martins2015eye}. Furthermore, work studying a
driver's situational awareness of the driving environment has shown the complexity and context-dependent nature of a
driver's gaze dynamics \cite{endsley1995toward,bolstad2008measurement}.

It is intuitive that high-resolution micro-glances such as blinks and individual saccades could be used to predict the
behavioral and environmental variables in this work. For example, glances patterns have been correlated with lane change
behavior \cite{pech2014head}. The open question is whether short-windowed macro-glances can be used for these
classification problems. This is an important question because detection and tracking of driver micro-glances in
in-the-wild on-road data is much less accurate than detection of macro-glances. The ability to rely on macro-glances
alone for predictive tasks allows for the design of robust, real-time driver assistance systems that modify the behavior
of the vehicle based on the detected states. 

\subsection{Prediction from Driver Glances}

Driver glances have been used to to predict several aspects of driver state including cognitive load
\cite{liang2007real}, secondary activity \cite{munoz2016distinguishing}, and drowsiness \cite{wang2006driver}, as
covered in this section. The key novel contribution of our work is that we are using glance patterns to predict aspects
of driving that are not obviously related to gaze and thus have not been analyzed in prior literature. These aspects
include demographics (e.g., age, gender), behavior (e.g., failure to signal, talking), and environment (intersection
proximity, lighting conditions, road type). In other words, this work serves as a new and useful exploration of what
broad macro eye-movement reveal about the state of the driver and the state of the driving environment.

\paragraph{Cognitive Load and Secondary Tasks} 

As reviewed in \cite{angell2015identification}, some initial work has been conducted that may be useful in the
development of algorithms for identifying periods of driving during which different types of task loading occur. Much of
this work was initially focused on identifying visual demand, or periods of visual or visual-manual task loading (e.g.,
\cite{victor2005sensitivity,liang2009detecting,kircher2009comparison,munoz2016distinguishing}). Additional work been
directed at developing algorithms to identify cognitive load using eye glance behavior and driving performance metrics
as inputs. Using data collected in a driving simulator, Zhang et al. explored a decision tree approach in
\cite{zhang2004driver} to estimate drivers' cognitive workload. Also working with simulator based data, Liang et
al. used similar measures in a support vector machine (SVM) approach using a 40-second window with 95\% overlap between
windows in \cite{liang2007real} to detect cognitive distraction, and obtained 91.6\% accuracy in the structured
predictions to which the model was applied. In \cite{liang2007nonintrusive} Liang et al. worked with Bayesian network
models and found that they could identify cognitive load reliably with an average accuracy of 80.1\%. They also found
that dynamic Bayesian networks (DBNs) gave a better performance than static Bayesian network models. Further, blink
frequency and eye fixation measures were particularly indicative of cognitive task workload in structured experimental
data. Building on the previous simulator based work, Liang used a hierarchical layered algorithm in
\cite{liang2009detecting}, which incorporated both a DBN and a supervised clustering algorithm, to identify feature
behaviors when drivers were in different cognitive states. Three groups of performance measures were used at the lowest
level of this algorithm: (1) eye movement temporal measures (blink frequency, fixation duration, etc.), (2) eye movement
spatial measures (spatial location of gaze in x, y, z), and (3) driving performance measures (steering error, steering
wheel standard deviation, lane position standard deviation) that were summarized across 30-second time windows, with no
overlap between windows. Liang interpreted a sequential analysis as indicating that from a risk state identification
perspective, it is not necessary to detect cognitive distraction if visual distraction is present as the latter
dominates. A recent work \cite{zhang2016evaluation} compared alternate SVM based classification approaches in a
simulation context with experimentally defined periods of visual-manual, cognitive, and combined distraction. A
``two-stage'' classifier first considered visual-manual distraction and then detecting dual or cognitive distraction
states was evaluated against a ``direct-mapping'' classifier developed to identify all distraction states at the same
time. Advantages and limitations to both approaches appeared. Liang's \cite{liang2009detecting} work is also relevant to
the current effort in that it considers issues related to applying detection algorithms to naturalistic data.

\paragraph{Drowsiness and Impairment} Detection of driver arousal from blink rates, eye movement, and gaze patterns has
received considerable attention in the simulated context and on small on-road datasets over two decades. A 10 year old
survey paper \cite{wang2006driver} on driver fatigue detection covers the features of eye and eyelid movement that have
continued to be used in papers that followed it. To the best of our knowledge, these features have not yet been proven
to be robust to the highly variable naturalistic driving conditions, perhaps due to the costs and challenges associated
with evaluating algorithms that require the collection of large driver-facing video datasets. The drowsiness detectors
that have been implemented in many commercial vehicles have relied instead on measures of vehicle dynamics and driving
performance \cite{eskandarian2007evaluation,sandberg2011detecting}.

\section{Glance Model and Prediction Approach}\label{sec:classification}

\newcommand{\demogtype}{\color[RGB]{191,0,0}Demographic}
\newcommand{\behavtype}{\color[rgb]{0,0,1}Behavior/State}
\newcommand{\envirtype}{\color[RGB]{0,105,31}Environment}

\definecolor{lightBvf}{gray}{0.7}
\newcommand{\headerBvf}[1]{\rule[-1.2em]{0em}{3em}\renewcommand{\arraystretch}{1}\begin{tabular}[c]{@{}l@{}}#1\end{tabular}}
\renewcommand{\arraystretch}{1.5}
\begin{table*}[h!]
  \centering
  \begin{tabular}{llllll}
    \hline
    \headerBvf{Binary Classification Problem} &
    \headerBvf{Type} &
    \headerBvf{Accuracy\\(Average)} &
    \headerBvf{Accuracy\\(St. Dev.)} &
    \headerBvf{Class 1 Size\\(Epochs)} &
    \headerBvf{Class 2 Size\\(Epochs)}\\
    \hline
    Weather (Clear vs Raining) & \envirtype & 51.6\% & 2.7\% & 4,261 & 345\\\arrayrulecolor{lightBvf}\hline
    Behavior (Speeding) & \behavtype & 52.3\% & 7.4\% & 3,497 & 101\\\arrayrulecolor{lightBvf}\hline
    Seatbelt (Yes vs No) & \envirtype & 55.8\% & 2.4\% & 4,101 & 601\\\arrayrulecolor{lightBvf}\hline
    Traffic Density (Low vs Medium) & \envirtype & 56.2\% & 0.4\% & 2,385 & 2,255\\\arrayrulecolor{lightBvf}\hline
    Traffic Divider (Present vs Not Present) & \envirtype & 56.6\% & 1.8\% & 3,102 & 1,423\\\arrayrulecolor{lightBvf}\hline
    Travel Lanes (2 or Less vs 3 or More) & \envirtype & 57.7\% & 1.2\% & 2,725 & 1,975\\\arrayrulecolor{lightBvf}\hline
    Alignment (Straight vs Curve) & \envirtype & 57.7\% & 2.6\% & 4,186 & 519\\\arrayrulecolor{lightBvf}\hline
    Age (Young vs Mature) & \demogtype & 58.3\% & 1.9\% & 1,657 & 1,188\\\arrayrulecolor{lightBvf}\hline
    Traffic Density (Medium vs High) & \envirtype & 58.6\% & 4.0\% & 2,255 & 176\\\arrayrulecolor{lightBvf}\hline
    Behavior (Following Too Closely) & \behavtype & 59.1\% & 2.3\% & 3,497 & 871\\\arrayrulecolor{lightBvf}\hline
    Lighting (Day vs Night with Light) & \envirtype & 59.2\% & 1.7\% & 3,257 & 1,059\\\arrayrulecolor{lightBvf}\hline
    Gender (Male vs Female) & \demogtype & 59.7\% & 0.6\% & 2,514 & 1,773\\\arrayrulecolor{lightBvf}\hline
    Surface Condition (Wet vs Dry) & \envirtype & 60.3\% & 1.9\% & 4,309 & 452\\\arrayrulecolor{lightBvf}\hline
    Age (Young vs Middle) & \demogtype & 60.7\% & 0.9\% & 1,657 & 1,442\\\arrayrulecolor{lightBvf}\hline
    Traffic Density (Low vs High) & \envirtype & 61.2\% & 3.9\% & 2,385 & 176\\\arrayrulecolor{lightBvf}\hline
    Locality (Rural vs Interstate) & \envirtype & 61.9\% & 0.8\% & 1,362 & 1,298\\\arrayrulecolor{lightBvf}\hline
    Locality (City vs Interstate) & \envirtype & 62.6\% & 2.6\% & 1,555 & 1,298\\\arrayrulecolor{lightBvf}\hline
    Lighting (Night with vs without Light) & \envirtype & 63.8\% & 2.2\% & 831 & 456\\\arrayrulecolor{lightBvf}\hline
    Age (Middle vs Mature) & \demogtype & 63.8\% & 1.5\% & 1,442 & 1,188\\\arrayrulecolor{lightBvf}\hline
    Locality (City vs Rural) & \envirtype & 63.8\% & 0.9\% & 1,555 & 1,362\\\arrayrulecolor{lightBvf}\hline
    Traffic Light/Sign (Present vs Not Present) & \envirtype & 64.0\% & 1.9\% & 4,252 & 377\\\arrayrulecolor{lightBvf}\hline
    Lighting (Day vs Night without Light) & \envirtype & 66.6\% & 1.9\% & 3,257 & 456\\\arrayrulecolor{lightBvf}\hline
    Near an Intersection (Yes vs No) & \envirtype & 70.9\% & 3.6\% & 4,025 & 764\\\arrayrulecolor{lightBvf}\hline
    Distraction (Talking) & \behavtype & 75.4\% & 2.0\% & 1,330 & 575\\\arrayrulecolor{lightBvf}\hline
    Behavior (Failed to Signal) & \behavtype & 75.5\% & 1.5\% & 3,497 & 247\\\arrayrulecolor{lightBvf}\hline
    Distraction (Fatigue) & \behavtype & 80.4\% & 3.1\% & 1,330 & 181\\\arrayrulecolor{lightBvf}\hline
    Distraction (Adjusting Radio) & \behavtype & 88.3\% & 2.1\% & 1,330 & 201\\\arrayrulecolor{black}\hline
  \end{tabular}
  \caption{This table answers the central question posed by this work: what aspects of the driver and driving
    environment can be predicted using a short sequence macro-glances? Each row specifies the binary classification
    problem, the variable type, accuracy mean and standard deviation, and the number of 6-second epochs
    associated with each glance. The rows are sorted according to average classification accuracy in ascending order.}
  \label{tab:affdiyqonqeghauhfzvi}
\end{table*}



Each six-second driving epoch contains the gaze region and a timestamp at the beginning of the epoch. Following this
tuple is an arbitrary number of similar tuples marking the macro-glance transitions and their associated
timestamps. These ``glance transitions'' refer to the moments in time when, based on the frame-by-frame annotations, the
driver's gaze changed from one region to another. ``Glance transitions'' are event-based (see discrete event simulation
\cite{fishman2013discrete}) in that they do not contain any self-transitions and only include changes of state. The
duration of a glance is encoded in the difference of the timestamps of adjacent transitions.

For the purpose of modeling both glance transitions and durations as a Hidden Markov Model (HMM), we discretize the
sequence of ``glance transitions'' into 25 state samples (spaced 250 milliseconds apart). By definition, the resulting
sequence of states allow for self-transitions. The probability of such self-transitions form a simple model of state
duration that was evaluated to be sufficient in this context. Explicit modeling of state duration for HMMs is an active
area of research \cite{yu2010hidden} and would be an effective extension to the model used in this work if epochs of
longer and non-uniform durations were considered. The sampling rate of 4 Hz for the 6-second epochs was determined to be
the lowest-resolution sampling that had below 1\% information loss over the original data. The result is that each
six-second epoch of ``glance transitions'' is reduced to a sequence of 25 macro-glance states and induced state transitions.

For classification, a fixed-length sequence of discrete values can be viewed as a categorical feature vector input to a
traditional classifier. We investigated this approach using parameter grid search of Random Forest and SVM classifiers,
both of which resulted in worse performance than what is reported in \secref{results}. The better performing approach
was to model the temporal structure of the sequence using a classic hidden Markov model (HMM). Each of the gaze regions
in the sequence are modeled as the discrete observation of the HMM. These observations can take on 8 values: (1)
rearview mirror, (2) center stack, (3) eyes closed, (4) interior object, (5) right, (6) forward, (7) instrument cluster,
and (8) left. An important point about this approach is that distinct macro-glance duration is not explicitly
modeled. The explicit-duration hidden semi-Markov model (HSMM) \cite{johnson2013hdphsmm} was evaluated for its ability
to model the sticky dynamics of each state. However, this approach did not perform well. We believe that this is due to
the limited and uniform length of each training sequence (see \secref{conclusion} for discussion of future work that
proposes further investigation of this kind of explicit duration modeling).

As described in \secref{results}, each prediction question is modeled a binary classification problem. One HMM is
trained per class. The number of hidden states in the HMM is set to 8, which does not correspond to any directly
identifiable states in the driving context. Instead, this parameter was programmatically determined to maximize
classification performance. One HMM is constructed for each of the two classes in the binary classification problem. The
HMM model parameters are learned using the GHMM implementation of the Baum-Welch algorithm
\cite{schliep2004general}. This process uses 80\% of the sequences from each of the two classes. As shown in
\tabref{affdiyqonqeghauhfzvi}, the classes are often unbalanced. In order to balance the training set, the minority
class is over-sampled using the SMOTE algorithm \cite{chawla2002smote}.

The result of the training process are two HMM models. Each model can be use to provide a log-likelihood of an observed
sequence. The HMM-based binary classifier then takes a 25-observation sequence, computes the log-likelihood from each of
the two HMM models, and returns the class associated with the maximum log-likelihood.

\section{Dataset and Results}\label{sec:results}

The 100-Car Naturalistic Driving Study dataset includes approximately 2,000,000 vehicle miles, almost 43,000 hours of
data, 241 primary and secondary drivers, 12 to 13 months of data collection for each vehicle, and data from a highly
capable instrumentation system including five channels of video and vehicle kinematics \cite{dingus2006100}. Our work
uses 4,816 six-second baseline driving epochs randomly selected from this dataset. Each epoch was manually annotated for
macro-glances based on the video of the driver's face. This annotation serves as the training and evaluation variables
for each of the binary classification tasks in \secref{binary-classification}.

\subsection{Baseline Epoch Dataset}\label{sec:dataset}

The 100-car study was the first large-scale naturalistic driving study of its kind \cite{dingus2006100,klauer2006impact}
and the forerunner of the much larger and subsequent SHRP2 naturalistic study. As such, the 100-car study was intended
to develop the instrumentation, methods, and procedures for the SHRP2 and to offer an opportunity to begin to learn
about how crashes develop, arise, and culminate based on recording of the pre-crash period (which had not been possible
prior to the development of methods used in the naturalistic 100-car study).

From the data that were acquired during the 100-car study, two databases were constructed: (1) an event database, and
(2) a baseline database.  The event database was comprised of epochs of driving that ended with a conflict.  Conflicts
were classified at four levels of severity: crash, near-crash, crash-relevant, and proximity-type conflicts.  These
event epochs were each 6 seconds long – (consisting of 5 seconds prior to a precipitating event and 1 second after).  After
data-acquisition, human analysts performed detailed extraction and coding of data that had been recorded during the
study for each 6 s period (including frame-by-frame analysis of glance behavior).  In addition, if a secondary task was
underway by a driver during this period, analysts coded it, and information about it.

The baseline database was constructed of 20,000 epochs -- also each 6 seconds long.  These baseline epochs were ones in
which the vehicle maintained a velocity over 5 mph – and in which driving occurred without incident (without any
conflict occurring).  Eye glance analyses were conducted on 5,000 of these baseline epochs.  Baseline epochs were
selected at random from all recorded data (excluding event data) -- and this selection did not make use of any kinematic
triggers.  The 6-second length of these epochs was chosen to match the length of event epochs.  Event variables such as
``precipitating factor'' and ``evasive maneuver'' (which were coded for event epochs) were not coded for baseline epochs
– since no conflict occurred within them. While the baseline epochs are free from safety-critical events (i.e., do not
contain crashes, near-crashes, or incidents), these epochs of ``just driving'' nevertheless are rich records of behaviors
that are undertaken by ordinary drivers on real roads during everyday driving.  This makes them an excellent source of
data for the work reported here.

The number of baseline epochs selected from each vehicle for the baseline database was determined by each vehicle's
involvement in crash, near-crash, and incident epochs in the event database. A stratified proportional sample of
baseline epochs was constructed such that vehicles which were involved in more conflicts, also contributed more baseline
epochs to the baseline database.  This was done to create the required basis for a case-control design needed for
odds-ratio calculations that were planned for subsequent analyses on the dataset.

From the 100-car study, a specific dataset was prepared and made accessible to the scientific community for analysis. It
may be downloaded (along with documentation) from \cite{vtti100cardata}. This database contains only de-identified data
(i.e., no video data are available).

\subsection{Binary Classification Performance}\label{sec:binary-classification}

\newcommand{\transfig}[3]{
  \begin{subfigure}[t]{#2}
    \includegraphics[height=2in]{images/transitions/#1.pdf}
    \caption{#3}
  \end{subfigure}
}
\begin{figure*}
  \centering
  \transfig{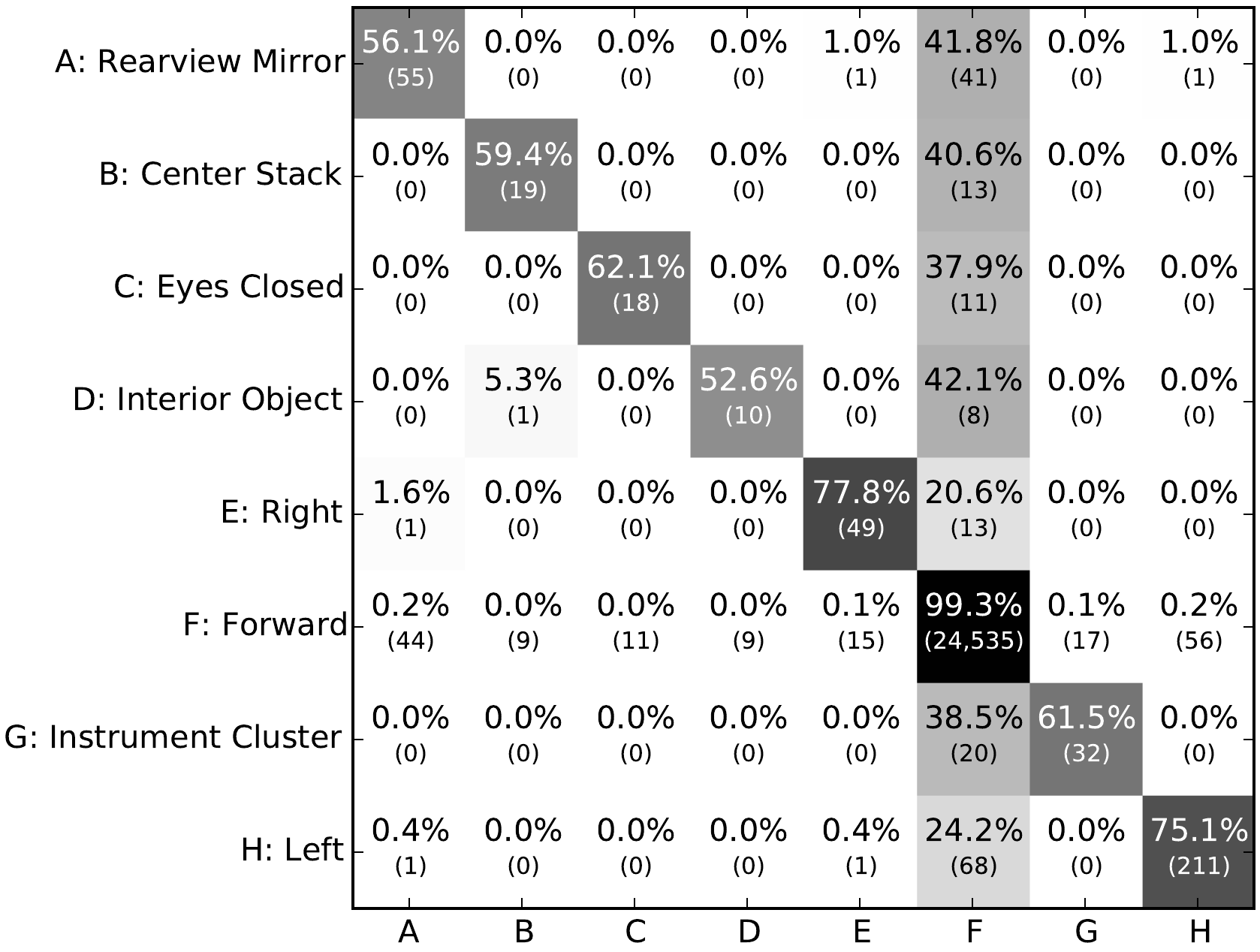}{0.37\textwidth}{Glance transition matrix for the first binary prediction class: ``not distracted''.}
  \hspace{0.1in}
  \transfig{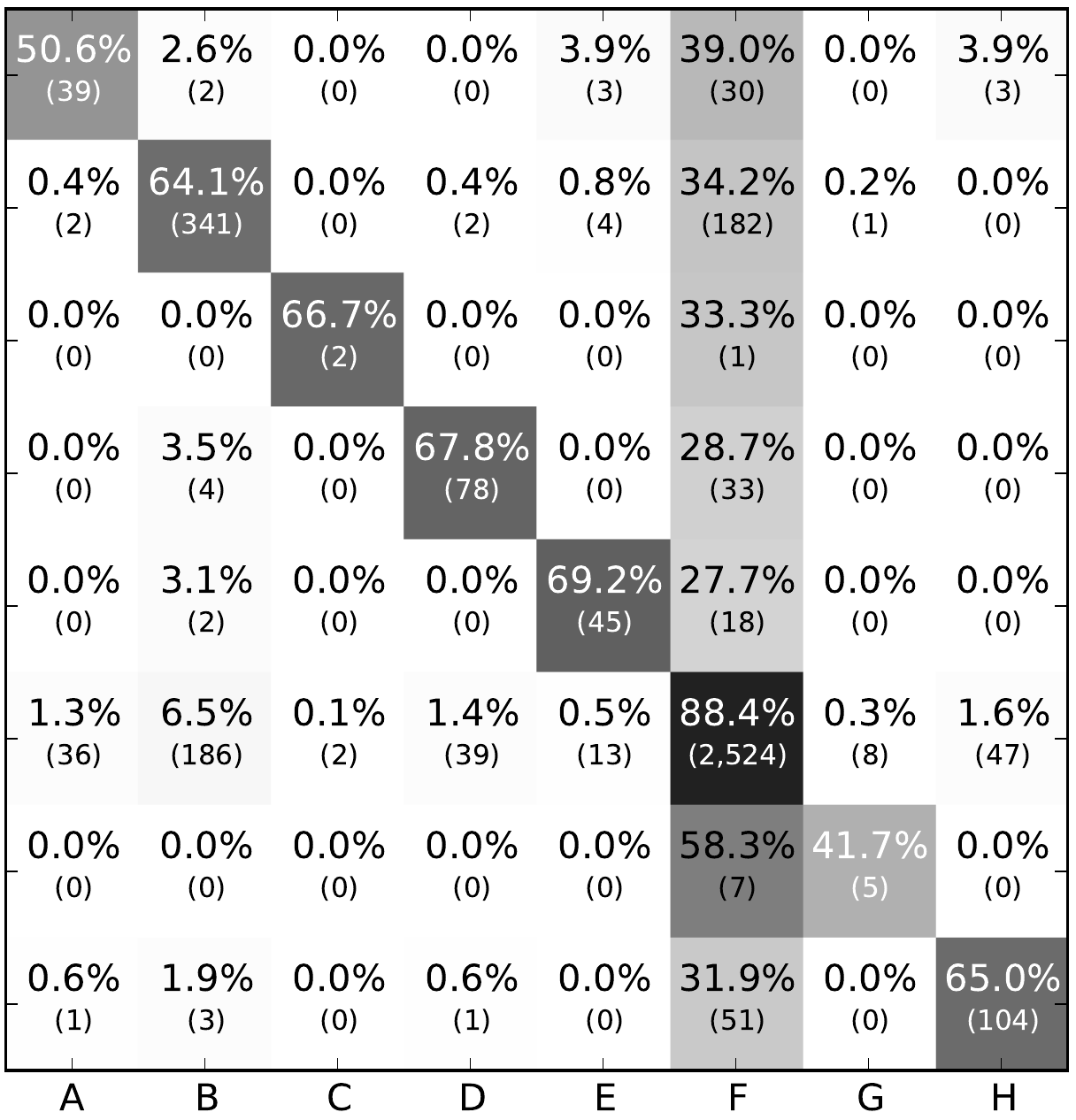}{0.27\textwidth}{Glance transition matrix for the first binary prediction class:
    ``adjusting radio''.}
  \hspace{0.1in}
  \transfig{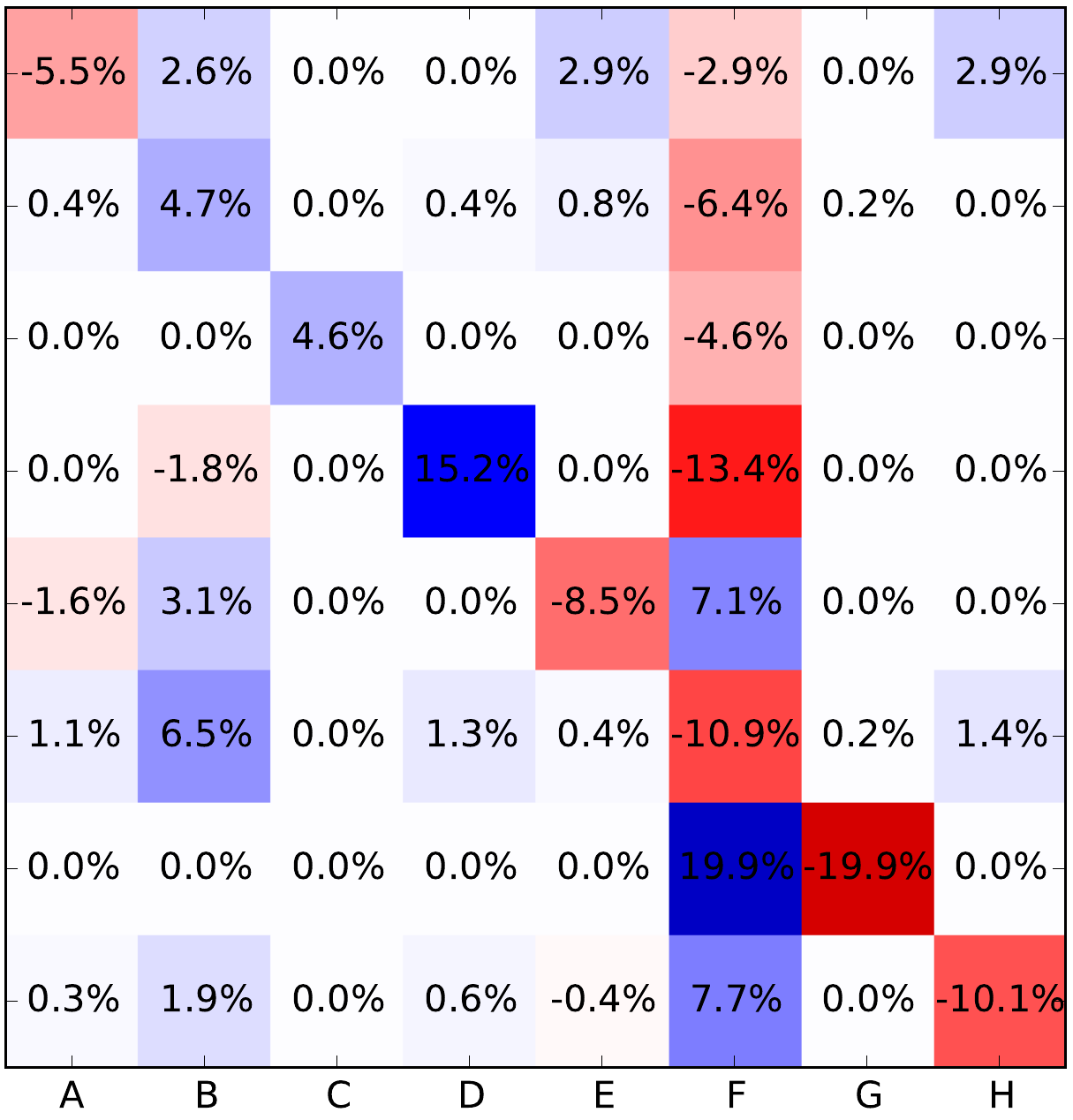}{0.27\textwidth}{The difference between the transition matrix for the two classes of ``not distracted'' and ``adjusting radio''.}
  \caption{An example of two glance transition matrices (first two subfigures) and their difference (third subfigure)
    that illustrates the discriminating characteristics of the glance dynamics based on which the two HMM models can
    make binary predictions. The y-axis is the ``transition from'' region and the x-axis is the ``transition to''
    region. In the first two subfigures, the percentages designate the probability of transitions and the values in
    parentheses show the absolute number of times those transitions appear in the dataset. The third subfigure shows the
    difference in probabilities between the first two subfigures. Blue indicates a positive difference, red indicates a
    negative difference. The intensity of color fill in each matrix cell, across all three subfigures, indicates
    magnitude of the values associated with the transition in that cell. The transition matrices and their differences
    for each of the binary classification problems considered in this paper are provided in the appendix.}
  \label{fig:transitions}
\end{figure*}

We evaluate the degree to which discriminative signal is present in 6-second bursts of macro-glances for the purpose of
predicting the following variables. We provide a brief description of each variable and the number of categorical values considered.

\begin{itemize}
\item Driving Environment
  \begin{itemize}
  \item \textbf{Proximity to an Intersection (2 values):} A vehicle is at or close to an intersection.
  \item \textbf{Lighting (3 values):} Daylight or evening, with the latter case considering with and without light.
  \item \textbf{Traffic Sign (2 values):} Presence of a traffic light or stop sign.
  \item \textbf{Locality (3 values):} Rural, interstate, and city.
  \item \textbf{Traffic Density (3 values):} Low, medium, or high. This level is based entirely on number of vehicles, and the
    ability of the driver to select the driving speed.
  \item \textbf{Surface Condition (2 values):} Wet or dry.
  \item \textbf{Weather (2 values):} Clear or rain.
  \item \textbf{Alignment (2 values):} Geographic curvature of the road: straight or curved.
  \item \textbf{Travel Lanes (2 values):} 2.5 is the threshold. The two categories are ``$\leq 2$'' and ``$\geq 3$''.
  \item \textbf{Traffic Divider (2 values):} Presence or absence of a median divider.
  \item \textbf{Seatbelt (2 values):} Wearing or not wearing a seatbelt.
  \end{itemize}
\item Driver Demographics
  \begin{itemize}
  \item \textbf{Age (3 values):} Young, middle, or mature. 23.5 is the threshold between young and middle. 40.5 is the
    threshold between middle and mature. Selected for dataset balance not behavioral profile.
  \item \textbf{Gender (2 values):} Female or male.
  \end{itemize}
\item Driver State and Behavior
  \begin{itemize}
  \item \textbf{Behavior (4 values):} Following too closely, failed to signal, speeding, or none.
  \item \textbf{Distraction (4 values):} Adjusting radio, fatigue, talking, or not distracted.
  \end{itemize}
\end{itemize}

These variables took on more values than those listed above, but the values were pruned in two ways. First, values for
distraction that were directly related to glance were removed. Obviously, 100\% accuracy can be achieved in predicting
glance region from glance region, so we are only interested in predicting driver state that does not
directly relate to glance. Second, we only considered values that were well-represented in the data. The threshold was
100 epochs. In some cases, the categorical values were combined. For example, age was collapsed into three groups:
young, middle, and mature. The partitioning was performed in a way that the numbers of epochs associated with each value
was balanced.

For variables that take on more than 2 values, we reduce the problem to a binary classification one between all pairs of
values. This allows us to explore the discriminating potential of macro-glances with regard to variables that take on 2,
3, or 4 values. For behavior and distraction variables, we only consider the pairing of values with the baseline state
of ``none'' and ``not distracted'', respectively.

\tabref{affdiyqonqeghauhfzvi} shows the result of applying the HMM-based classification method in
\secref{classification} to the binary classification problems associated with the variables listed above. This table
contains the answer to the question posed in the title of this paper. As expected, the problem of predicting anything
about the driver or driving environment from just 6 seconds of macro-glances is very difficult, because both the
duration of the sequence is too short and the resolution of the captured dynamics is too coarse. Nevertheless, accuracy
of above 75\% is achieved for the prediction of the following four driver behaviors and states: talking, fatigue,
radio-tuning activity, and failure to signal. Also, 70.9\% accuracy is achieved in predicting whether or not the vehicle
is approaching an intersection. To provide intuition as to why this prediction works at all, \figref{transitions}
provides a visualization for the dynamics of the radio-tuning activity classification problem. The first two subfigures show
the glance transition matrices for the ``not distracted'' and ``adjusting radio'' classes. The third subfigure shows the
difference in transition probabilities between the first two subfigures. These matrices show that there are
significant aggregate differences in macro-glance transitions for the two classes. The results in \tabref{affdiyqonqeghauhfzvi}
show that these differences can be detected from six-second epochs for several variables critical for the design of
intelligent driver assistance systems.

\begin{figure*}[ht!]
  \centering
  \includegraphics[width=\textwidth]{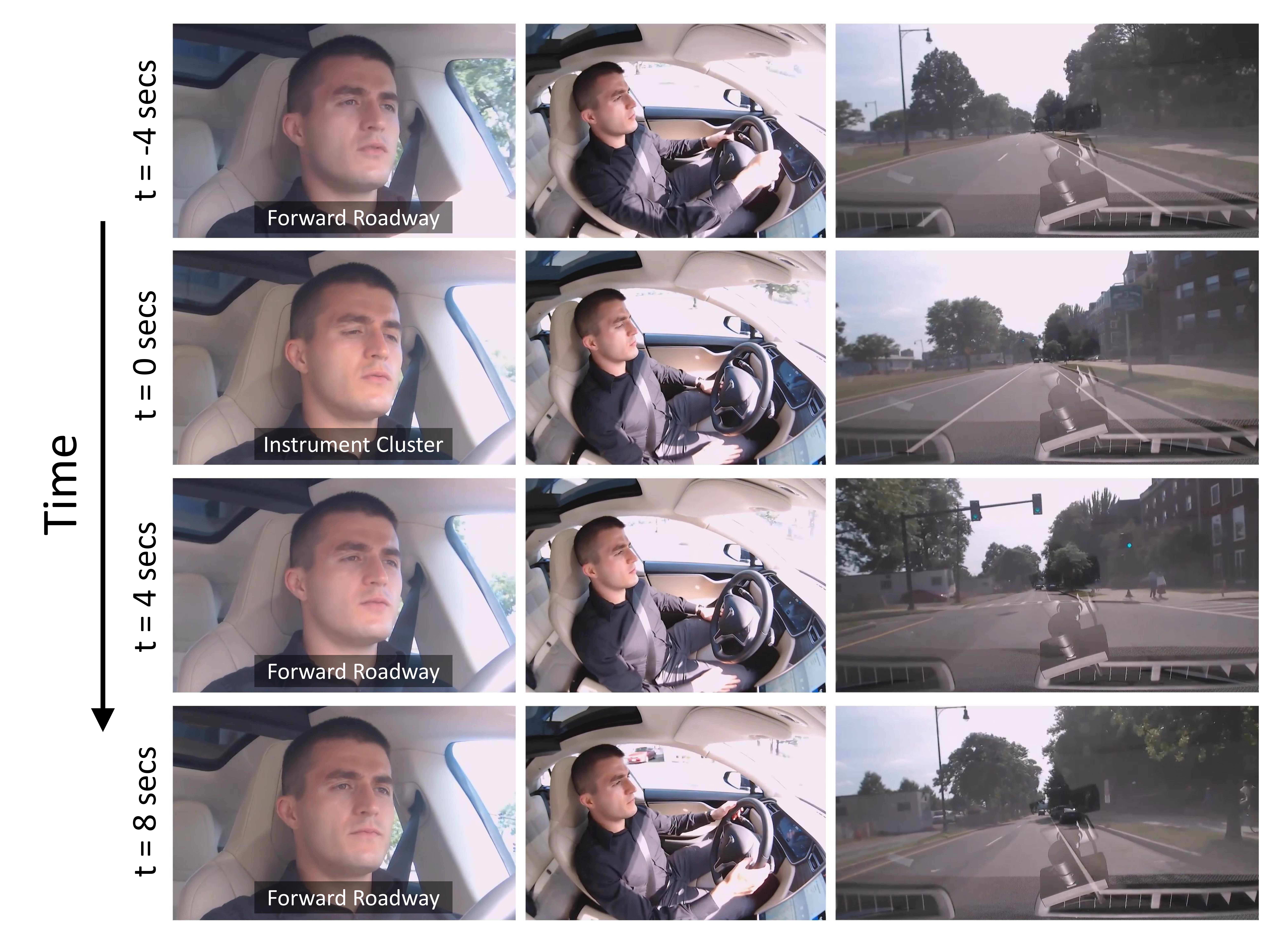}
  \caption{Illustrative sequence of snapshots collected in a Tesla vehicle of a driver's macro-glances. A time of $t=0$
    is marked as the point at which the driver can be annotated as being near an intersection.}
  \label{fig:intersection}
\end{figure*}

\begin{figure*}[ht!]
  \centering
  \includegraphics[width=0.8\textwidth]{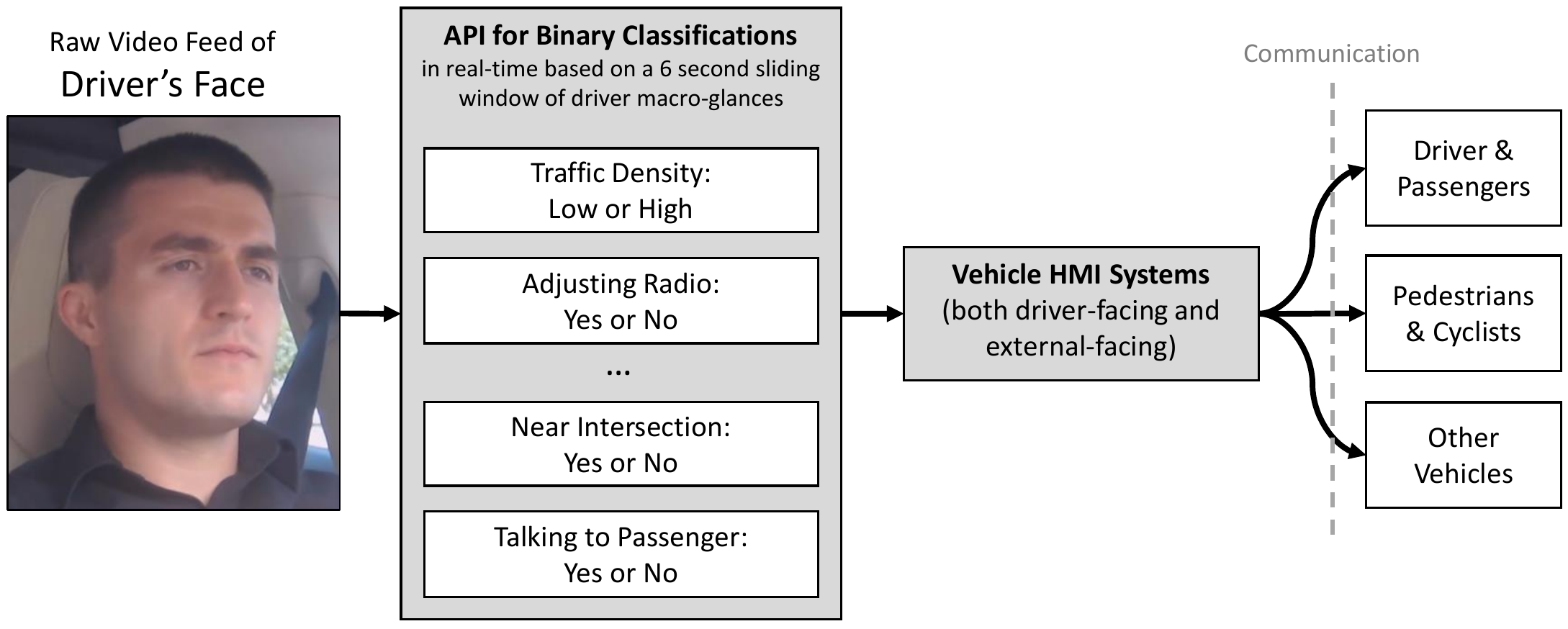}
  \caption{A framework for a system that processes (in real-time) the video of the driver's face and exposes an API to
    the HMI systems operating in the car that can appropriately alter their communication strategy with the driver,
    vehicle passengers, and the external driving environment based on the binary sensor signals.}
  \label{fig:framework}
\end{figure*}

\subsection{Framework for Gaze-Based HMI in the Car}\label{sec:framework}

The focus of this paper is to begin to answer the open question whether temporal patterns of macro-glances in the car
contain sufficient discriminative signal to predict the state of the driver, the car, and the world. Our approach
naturally leads to a real-time application of its various detectors in a vehicle. \figref{intersection} shows an
illustrative example of a driver approaching an intersection while operating a Tesla vehicle in Autopilot
mode. \figref{framework} proposes how such a sequence of videos would be processing through time. In this proposed
system, a sliding window of 6 seconds is used to make a series of binary predictions. Each of the predictions along with
an estimated confidence are exposed through a CAN network API. HMI systems in the car are then able to alter their
communication strategy with the driver and the external environment based on the predictions each system listens to for
cases when the prediction confidence exceeds a minimum threshold.

In fact, an HMI system may utilize multiple sensor streams, only part of which would be the sensors derived from
macro-glances. This is grounded in the design of joint cognitive systems envisioned and developed over the previous 3
decades \cite{woods1985cognitive,hollnagel1983cognitive,miller2015joint}. For example, for the case in
\figref{intersection}, the prediction may be that the driver's macro-glances are not indicative of proximity to an
intersection even though based on the GPS coordinates reported on the CAN network, the car is in fact approaching
intersection. This, in combination with the fact that the Tesla is operating under Autopilot and is thus driving itself,
can be used by the car's external signaling system to infer that the driver is not paying attention and is unaware of
the intersection. This information can then be conveyed to pedestrians and other vehicles, so that they make their
movement decisions with a higher degree of caution.

\section{Conclusion}\label{sec:conclusion}

This work asks what can and cannot be predicted from short bursts of driver macro-glances. We consider a representative
sample of 4,816 annotated six-second epochs of driving from a 100-car naturalistic study. The variables under
consideration fall into three categories: driving environment, driver behavior/state, and driver demographic
characteristics. We form binary-classification problems from all the well-represented variables available in the dataset
and model regularly-sampled macro-glances as a hidden Markov model for each class to make the binary prediction. The
results show that radio-tuning activity, fatigue state, failure to signal, talking, and proximity to an intersection can
be predicted with 70.9\% to 88.3\% accuracy. Based on these results, the general conclusion of this work is that
macro-glances can be part of a multi-sensor system for predicting external environment factors, but on its own is only
sufficient to predict a limited but important set of variables related to driver behavior and state. Nevertheless,
significant improvements in accuracy may be achievable through further development of the underlying algorithmic
approach. To this end, future work will investigate whether other approaches that capture temporal dynamics in the data,
such as Hidden Semi-Markov Models (HSMM) \cite{johnson2013hdphsmm} or Recurrent Neural Networks (RNN)
\cite{schmidhuber2015deep}, may perform better than HMMs, in which case macro-glances alone may be used as the basis for
environment, behavior, state, and demographic prediction in future real-time driver assistance systems. Furthermore,
using macro glance epochs of heterogeneous duration for training and evaluation may result in significant increases in
prediction accuracy due to the fact that some environmental or behavioral factors may reveal themselves on different
time-scales. For example, detection of talking may only need 1-2 seconds of macro-glances, while the detection of rural versus
urban environmental conditions may requires an epoch of 10-20 seconds.

\section*{Acknowledgment}

Support for this work was provided by the Santos Family Foundation, the New England University Transportation Center,
and the Toyota Class Action Settlement Safety Research and Education Program. The views and conclusions being expressed
are those of the authors, and have not been sponsored, approved, or endorsed by Toyota or plaintiffs' class
counsel. Data was drawn from studies supported by the Insurance Institute for Highway Safety (IIHS).

%

\bibliographystyle{sigchi}
\bibliography{bib-hundred,bib-lex-fridman,bib-agelab}

\clearpage

\onecolumn

\appendix\label{sec:transition-matrices}

Each of the binary classification problems considered in this paper (see \tabref{affdiyqonqeghauhfzvi}) has two
non-overlapping classes. That is, for each problem, there is a set of 6-second epochs associated with either the first
class, the second class or neither class. For each of these epochs, there is a sequence of 25 discrete states (glance
regions) spaced evenly in time. Transition probabilities shown in this appendix are referring to state-transition within
this sequence of discrete state.

\figref{transitions} in the main body of the paper visualizes the transition probabilities for epochs associated with
each of the two classes for the binary classification problem of ``not distracted'' versus ``distracted while adjusting
radio.'' It also shows the difference between these two transition matrices. In this appendix, we perform the same
visualization for all of the binary classification problems considered in this paper. The problems are presented in the
same order as they appear in \tabref{affdiyqonqeghauhfzvi}, that is in the order from lowest to highest average
classification accuracy.

A key observation to make from the visualizations that follow is that the problems with lower classification accuracies
generally show less differences in the third subfigure, while problems with higher classification accuracies generally
show greater differences. In other words, these visualizations reveal the aggregate disciminative characteristics of
each problem. The HMM approach used for binary classification in this work exposes these discriminative characteristics
at the individual glance sequence level. The average accuracy achieved are listed in the caption of each figure set.

\newcommand{\transallfig}[1]{
   \includegraphics[height=2.15in]{images/transall/#1.pdf}
}

    \begin{figure}[!h]
      \transallfig{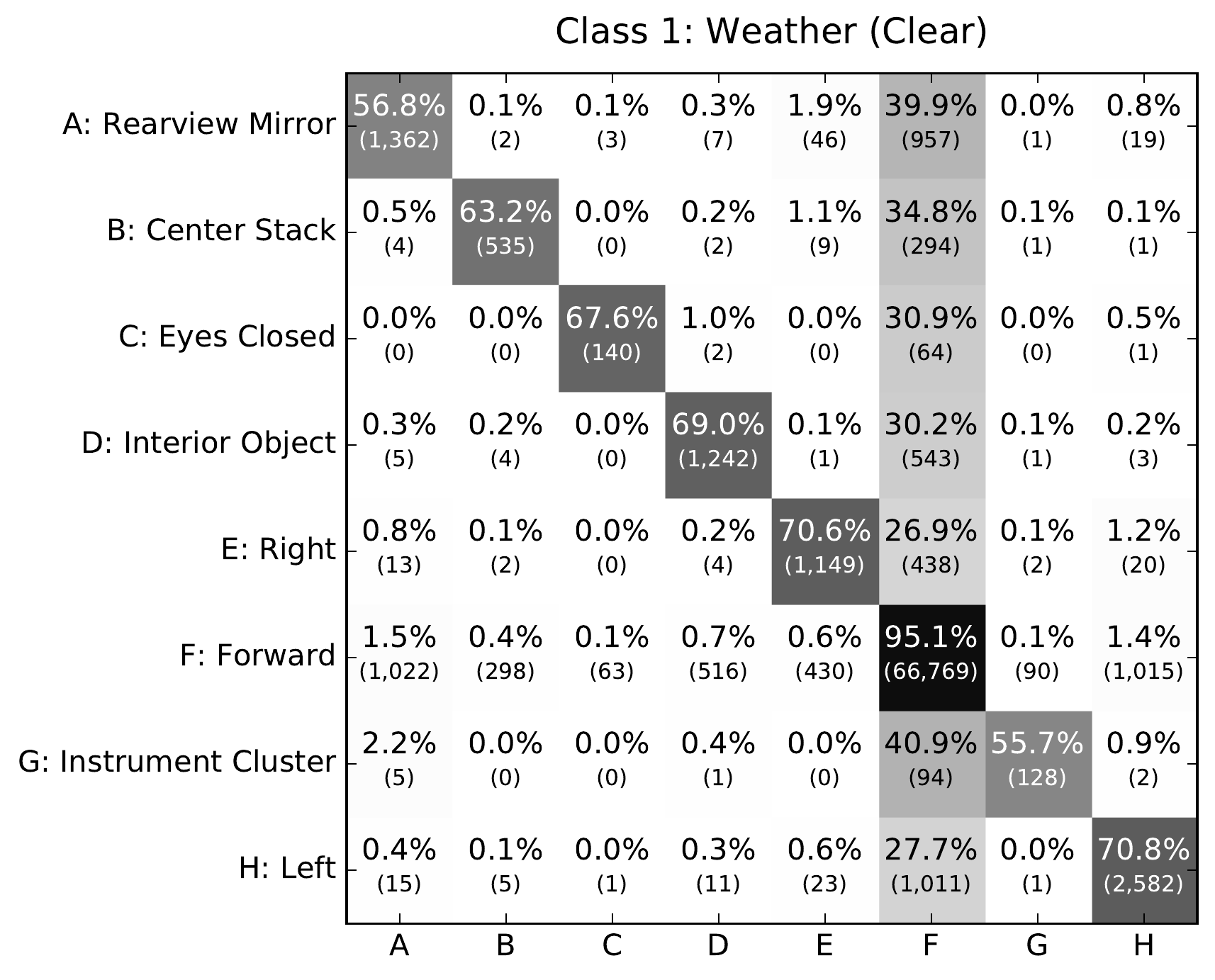}
      \transallfig{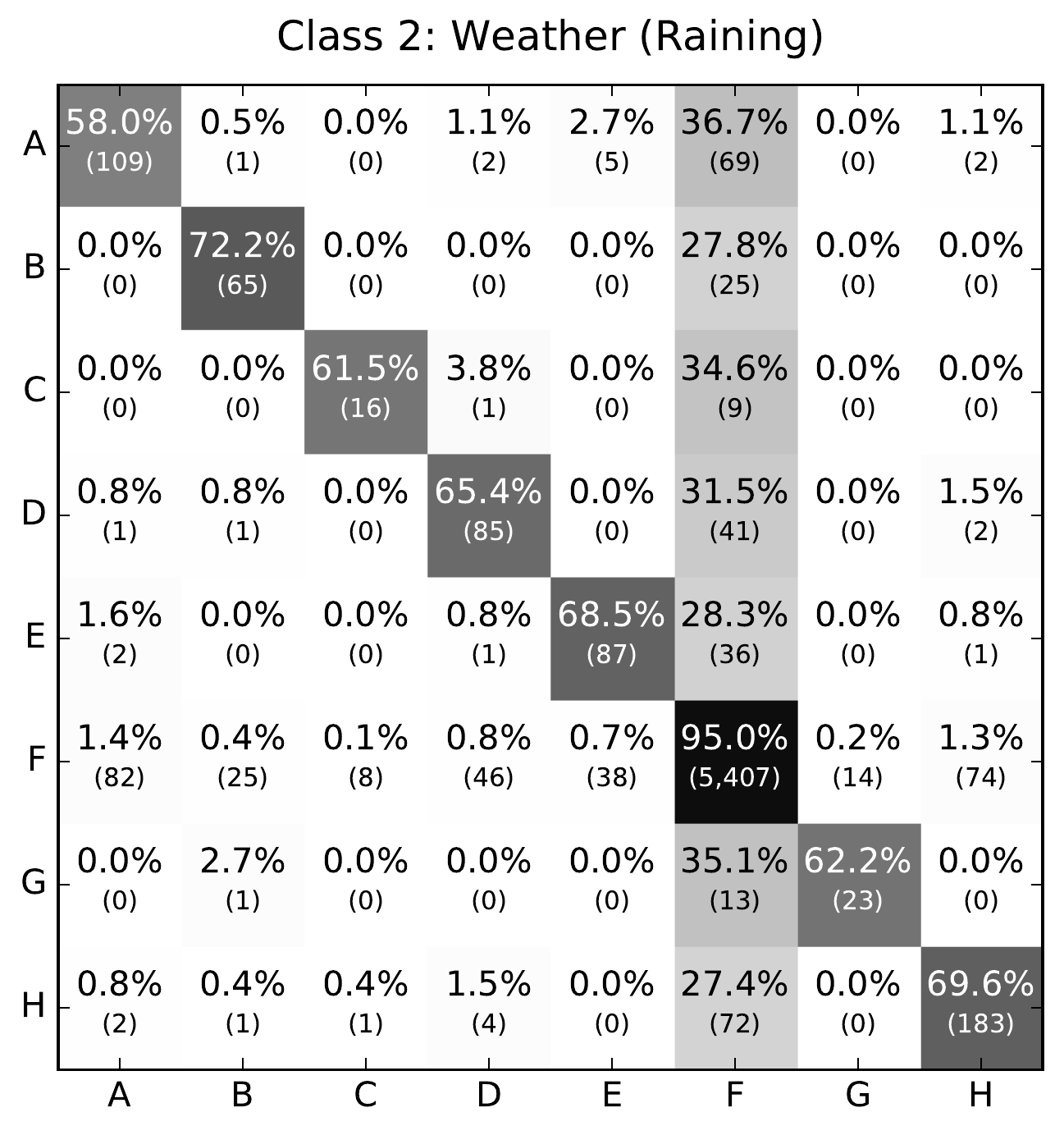}
      \transallfig{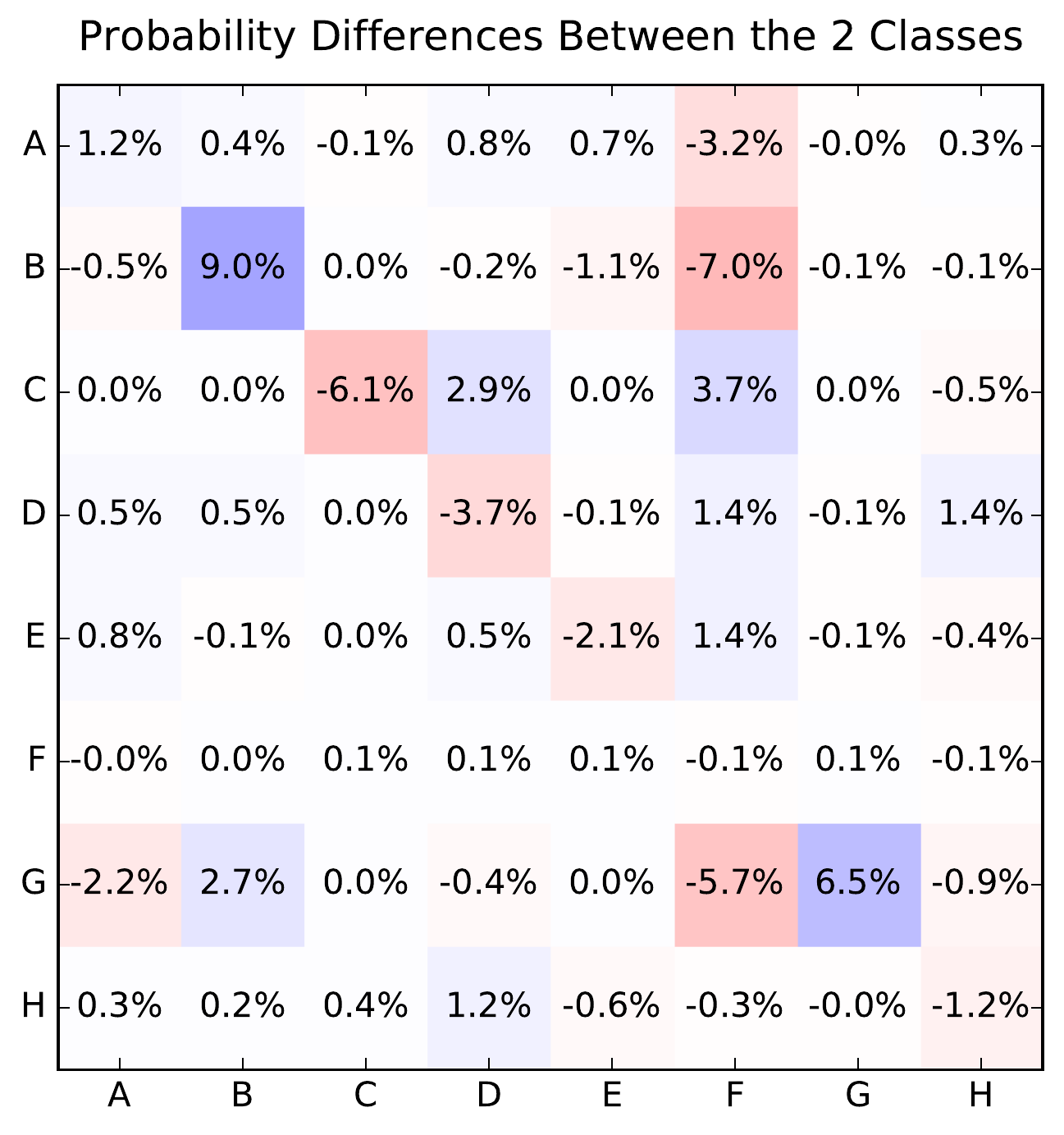}
      \caption{Transition matrices for Weather (Clear vs Raining). Classification accuracy: 51.6\%.}
      \label{fig:p00_weather_01}
    \end{figure}
    
    \begin{figure}[!h]
      \transallfig{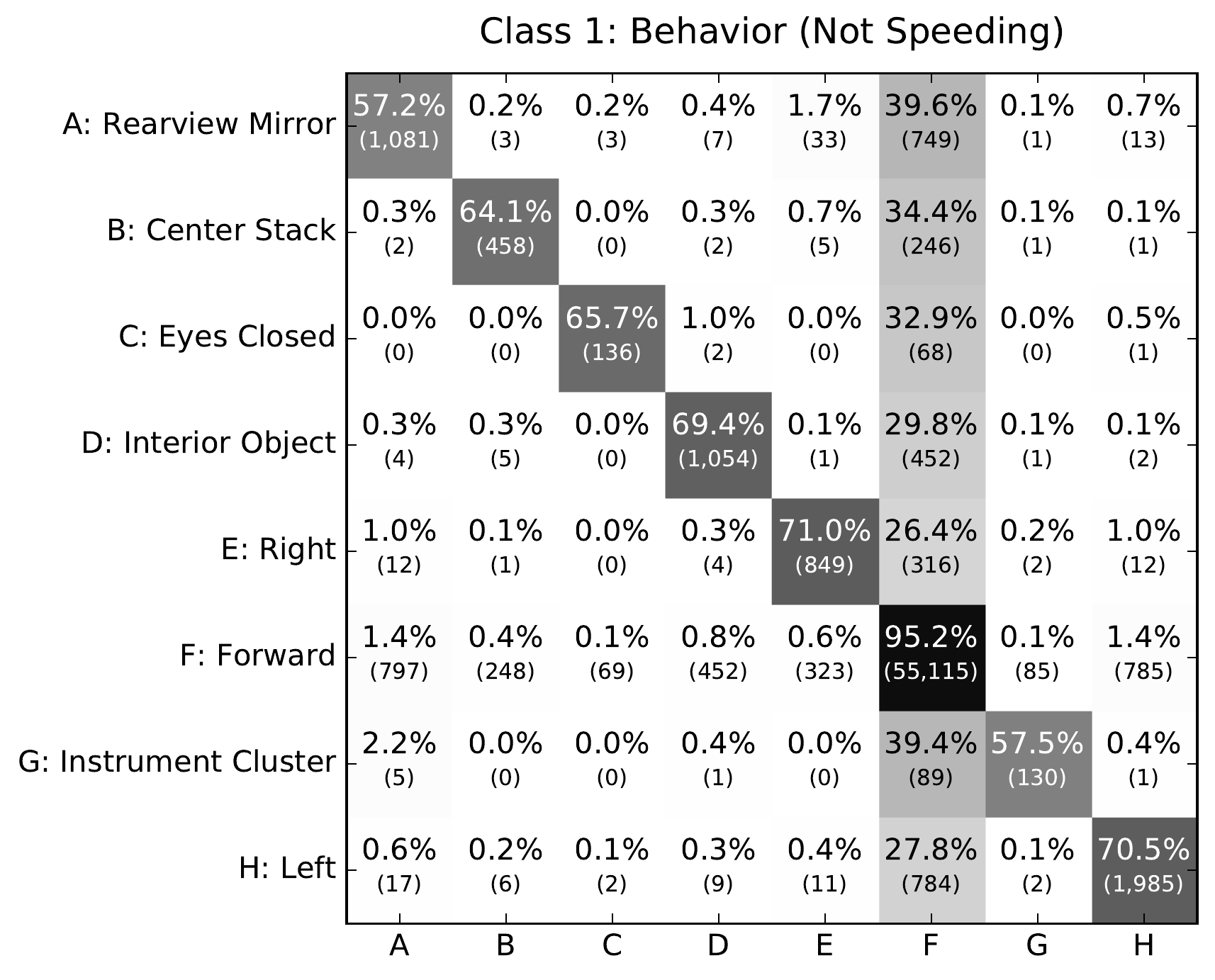}
      \transallfig{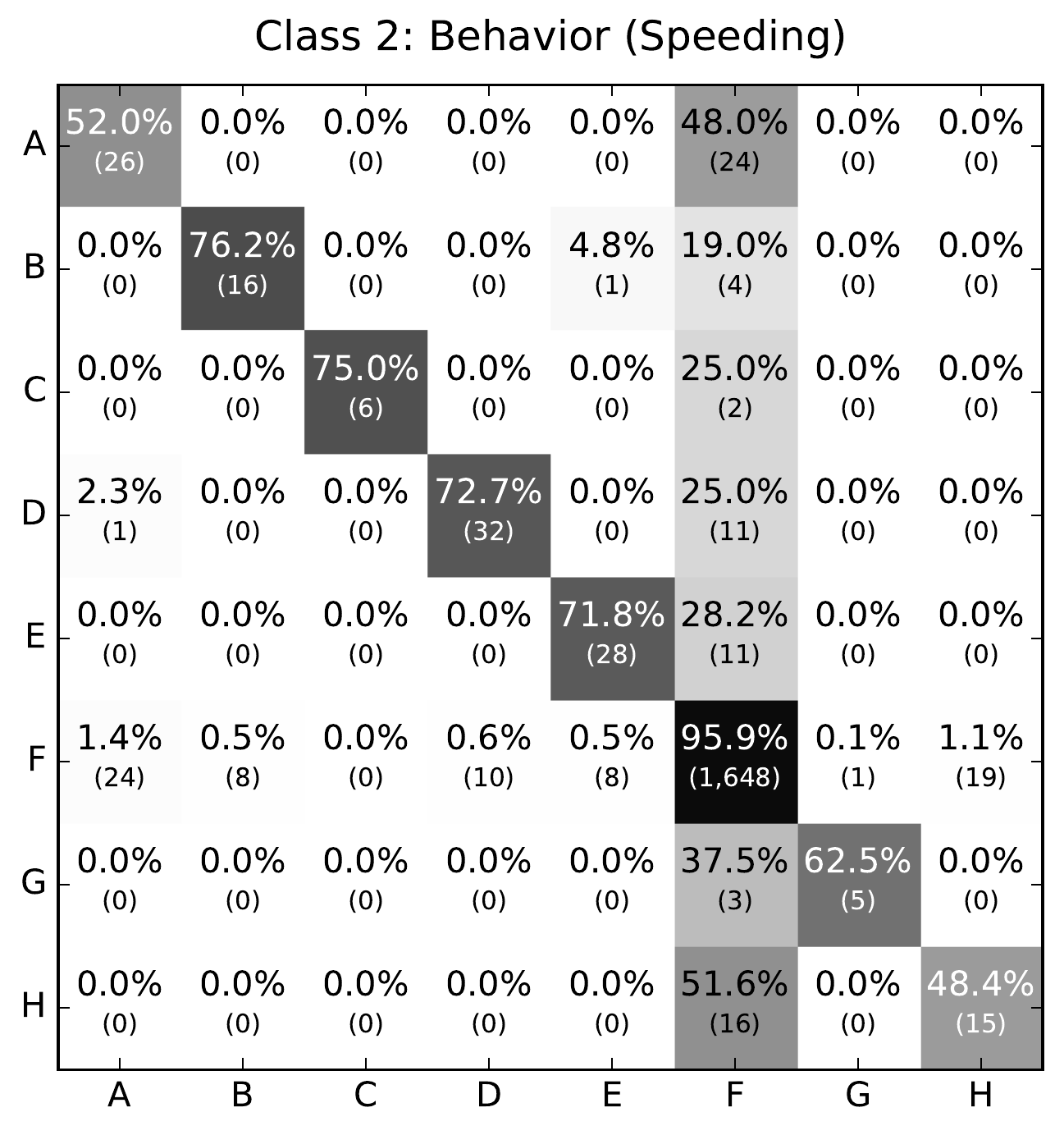}
      \transallfig{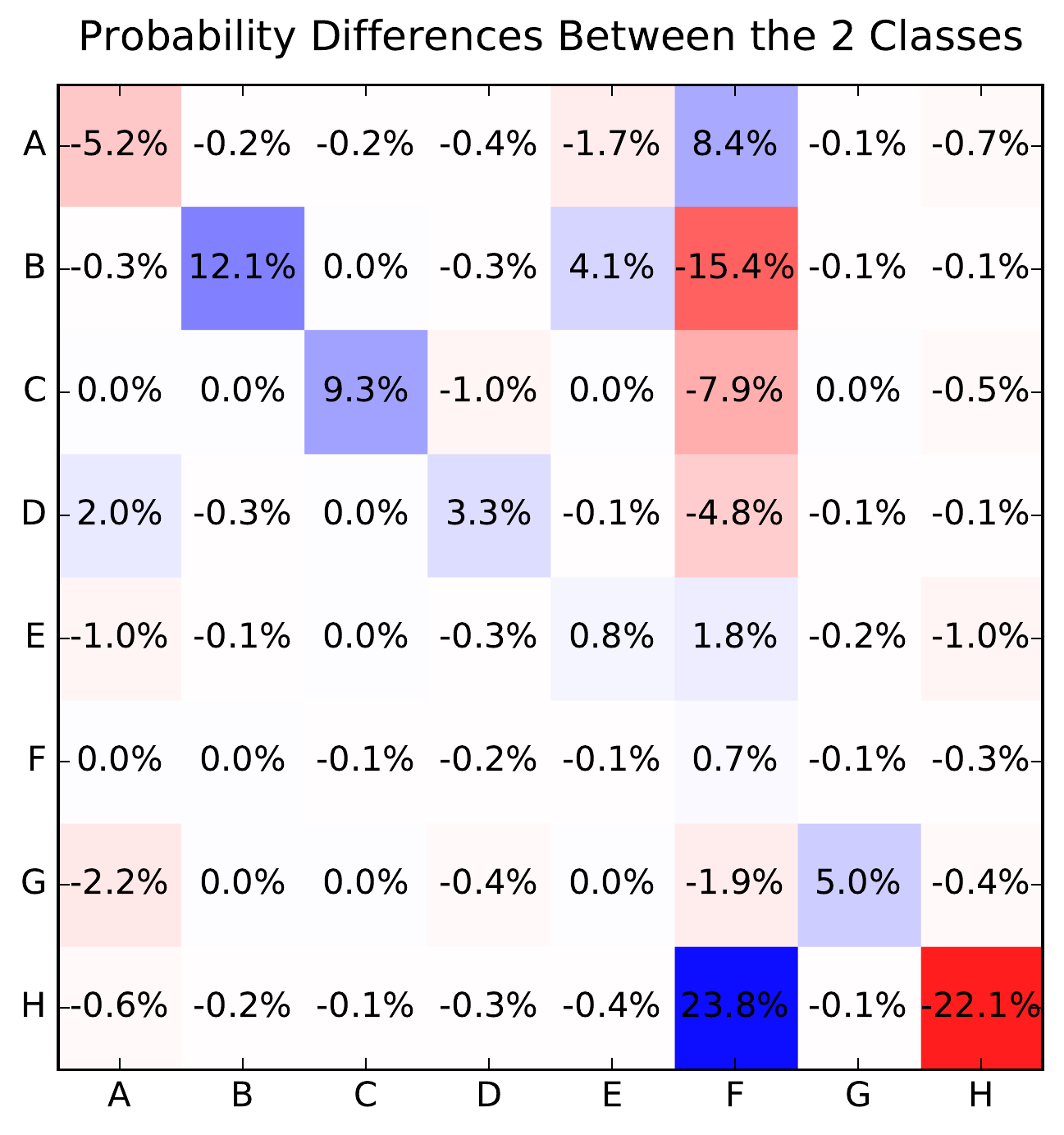}
      \caption{Transition matrices for Behavior (Speeding). Classification accuracy: 52.3\%.}
      \label{fig:p01_behavior_03}
    \end{figure}
    
    \begin{figure}[!h]
      \transallfig{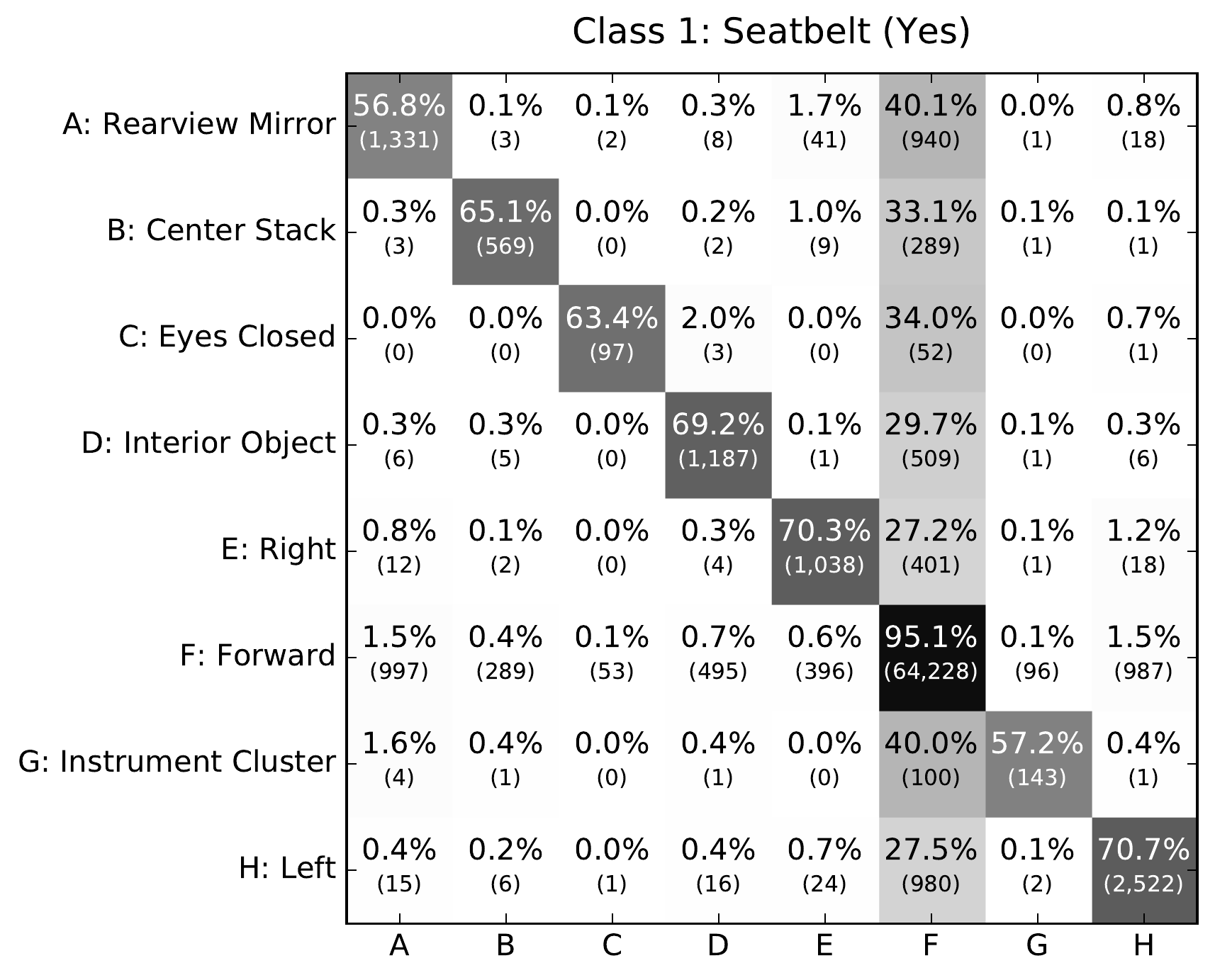}
      \transallfig{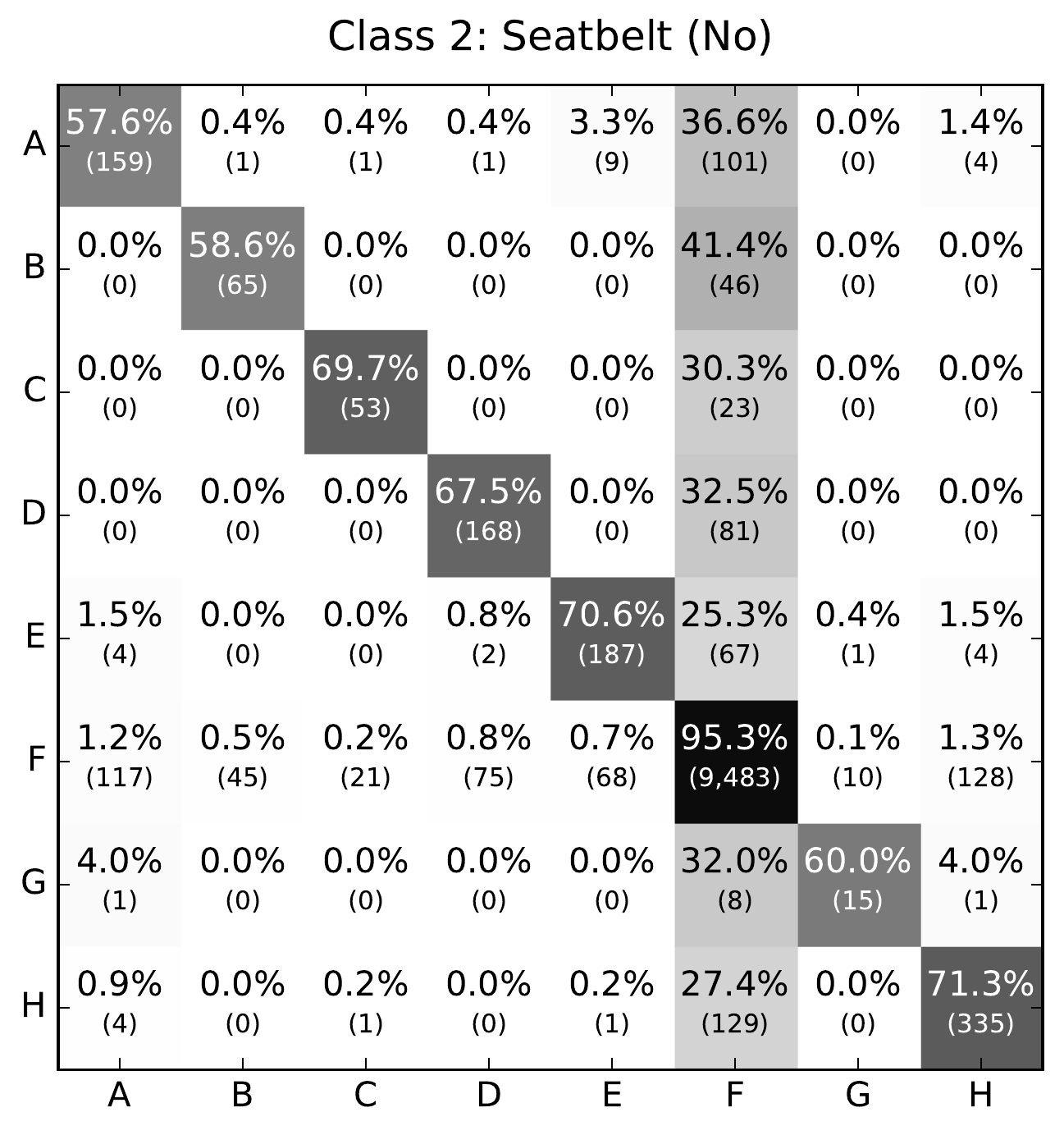}
      \transallfig{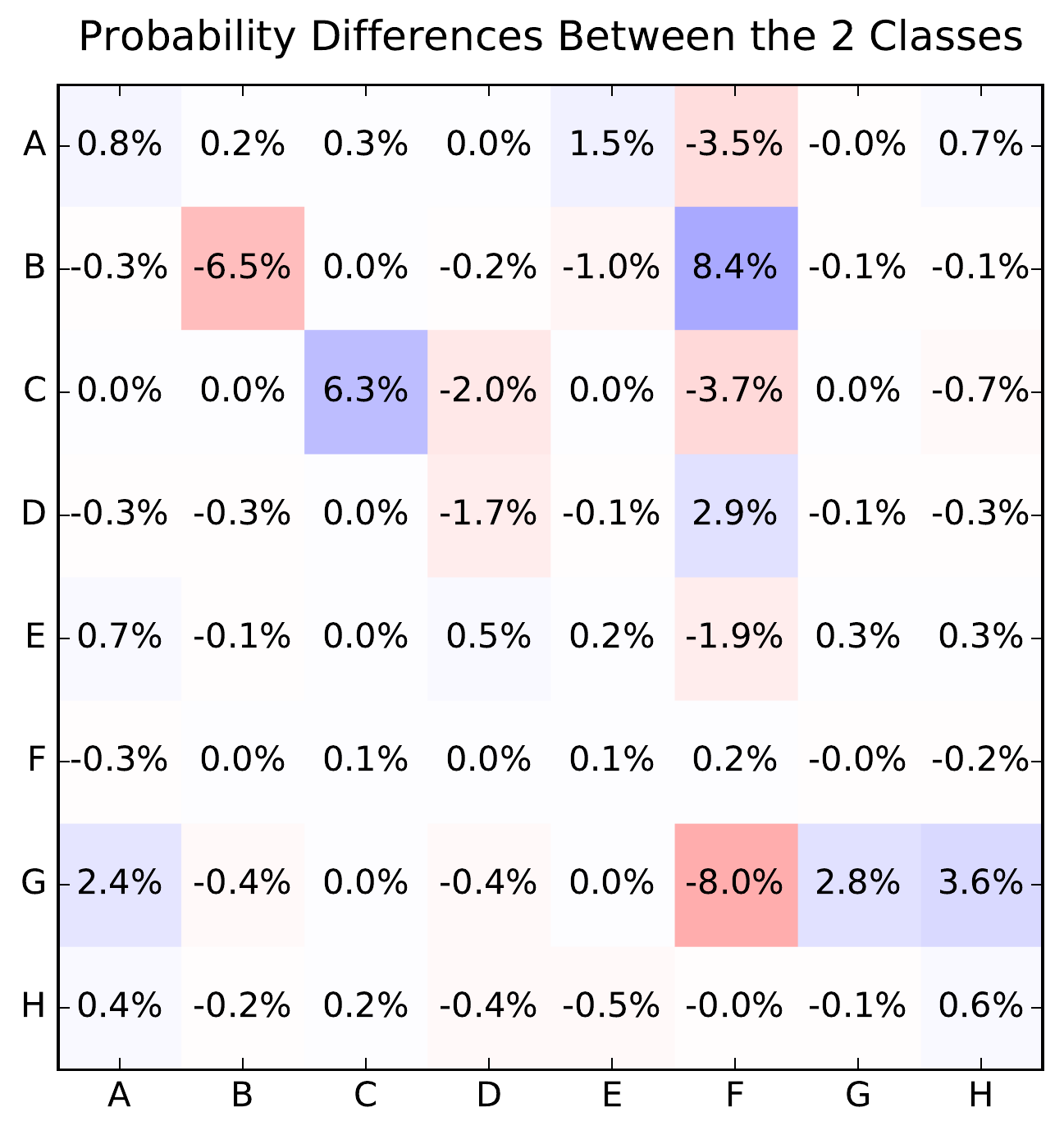}
      \caption{Transition matrices for Seatbelt (Yes vs No). Classification accuracy: 55.8\%.}
      \label{fig:p02_seatbelt_01}
    \end{figure}
    
    \begin{figure}[!h]
      \transallfig{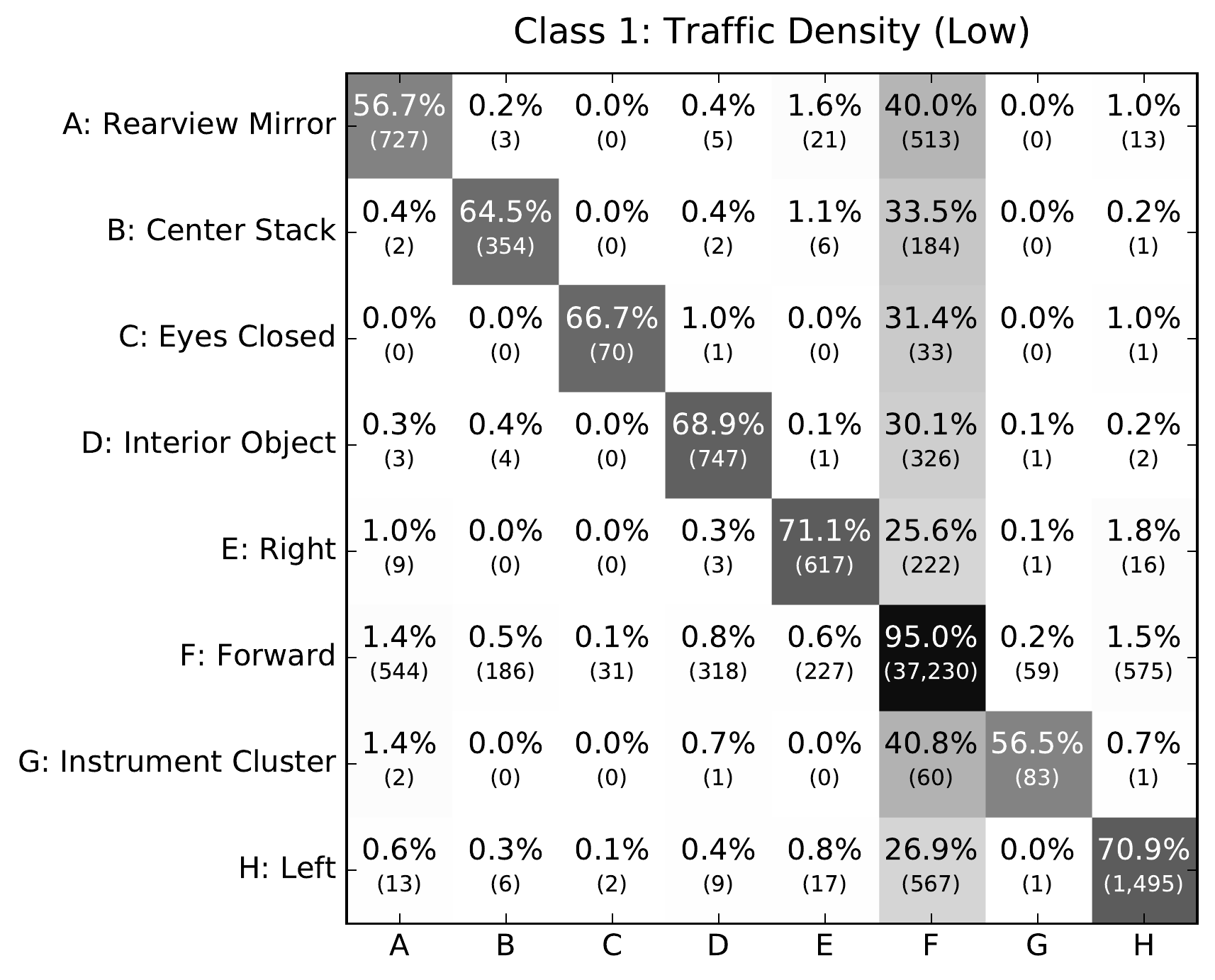}
      \transallfig{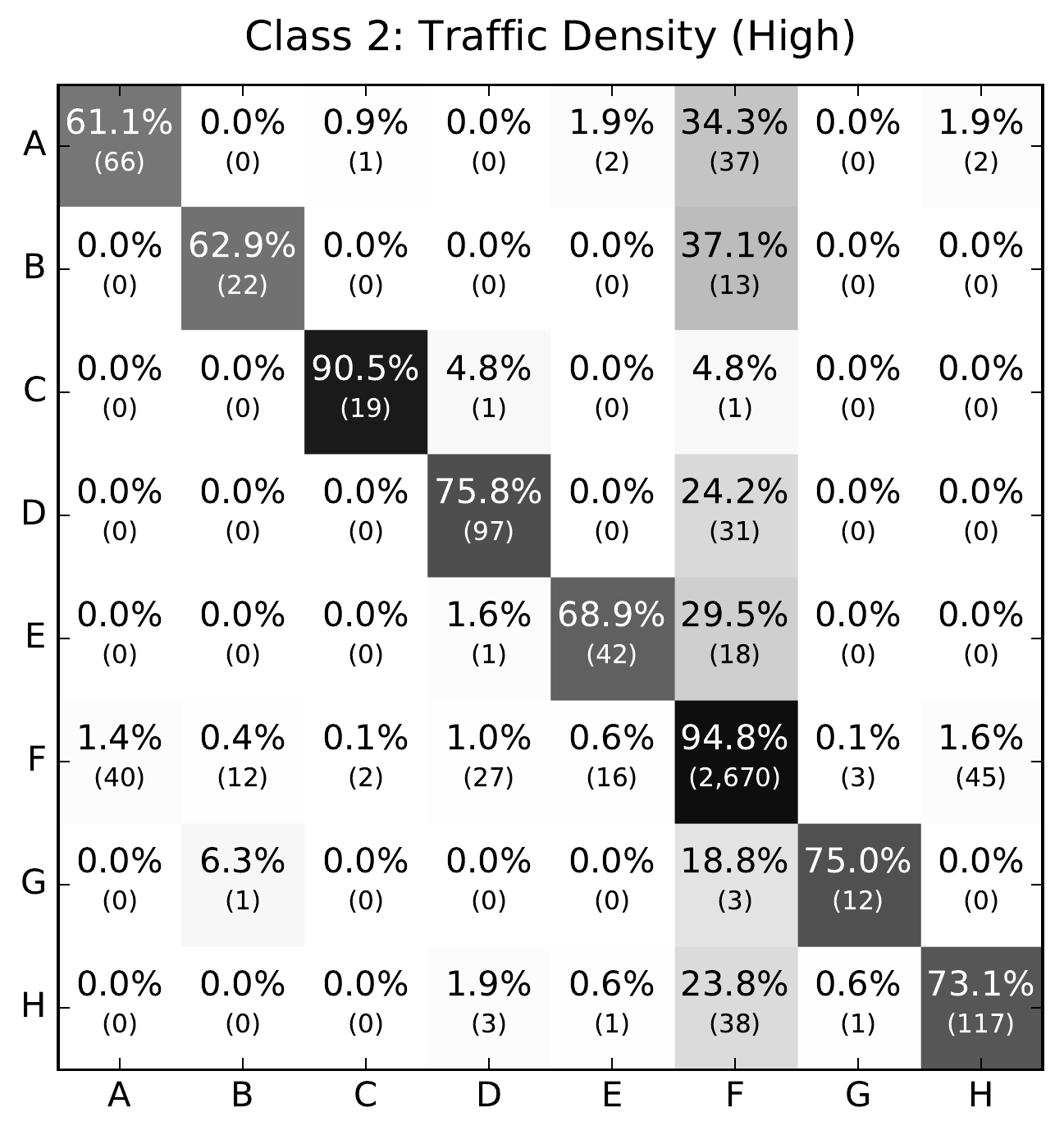}
      \transallfig{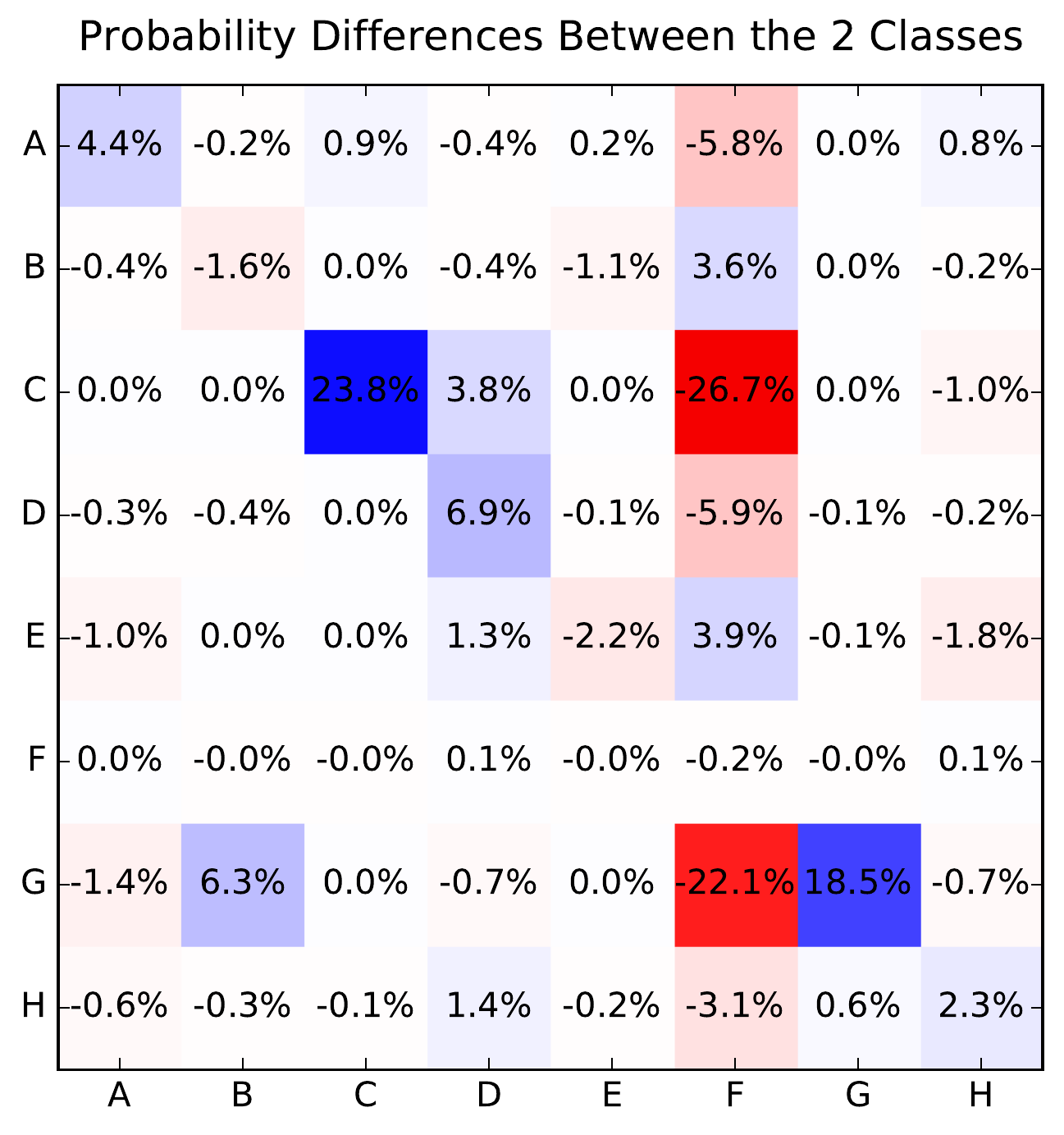}
      \caption{Transition matrices for Traffic Density (Low vs High). Classification accuracy: 56.2\%.}
      \label{fig:p03_traffic_density_02}
    \end{figure}
    
    \begin{figure}[!h]
      \transallfig{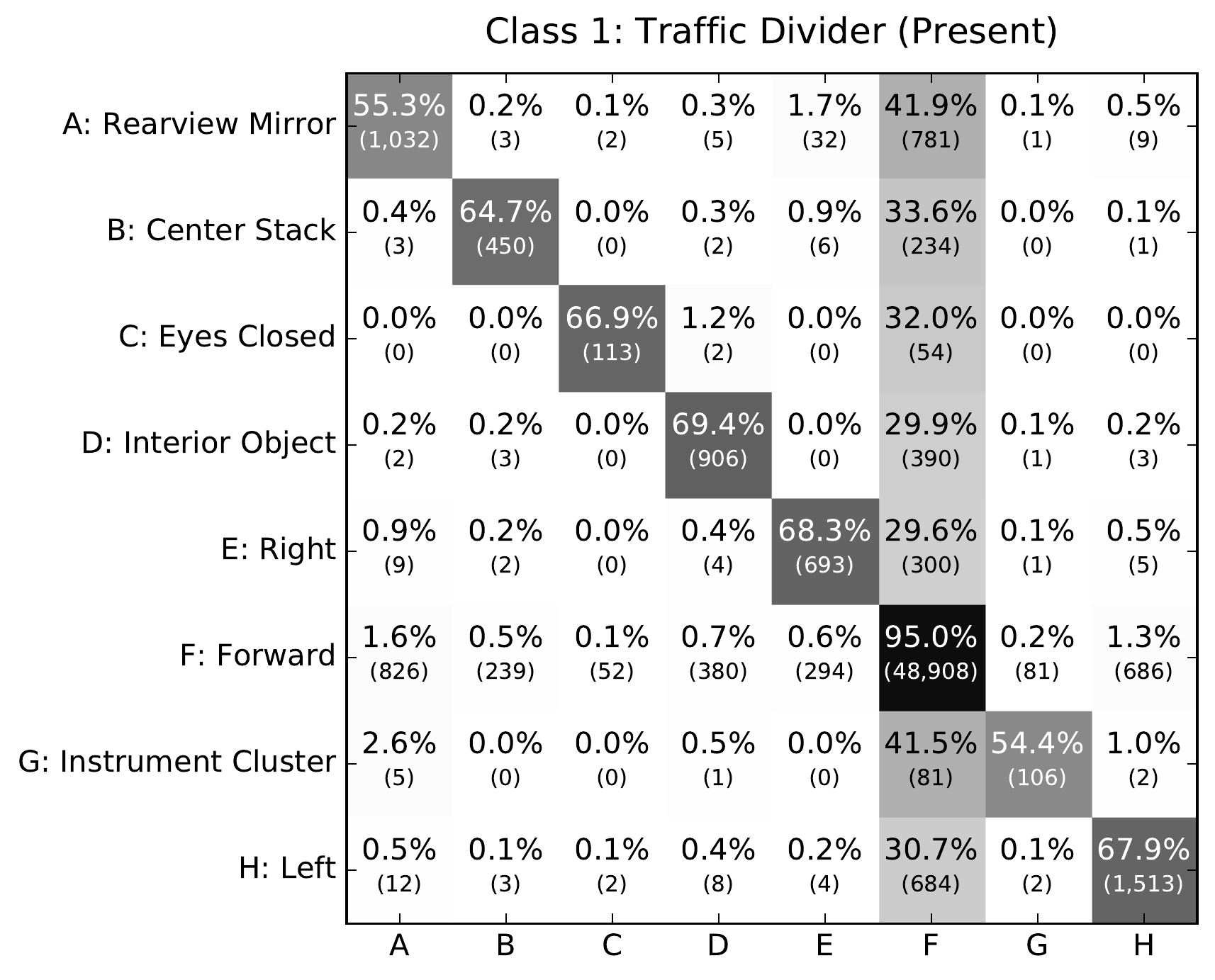}
      \transallfig{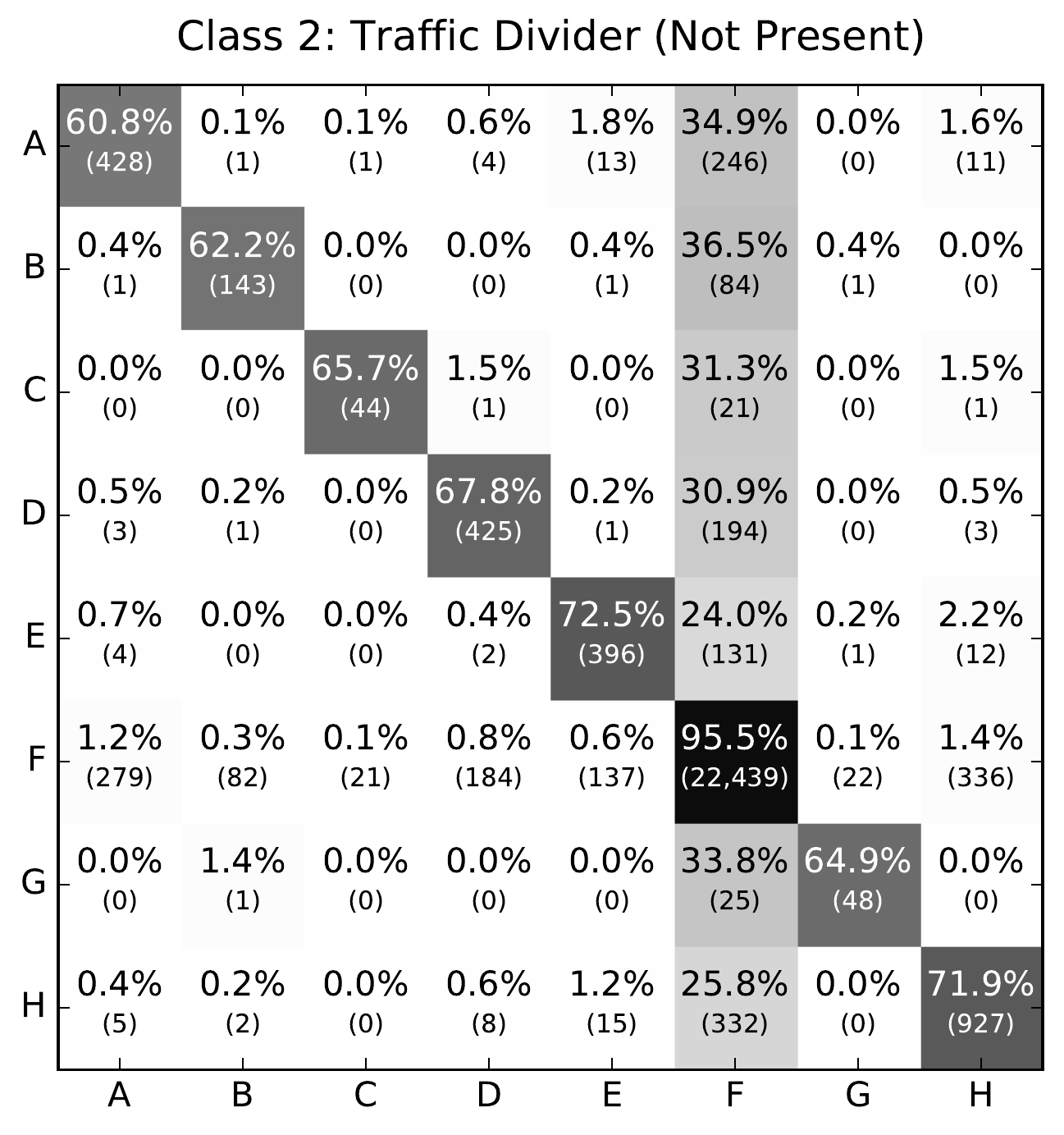}
      \transallfig{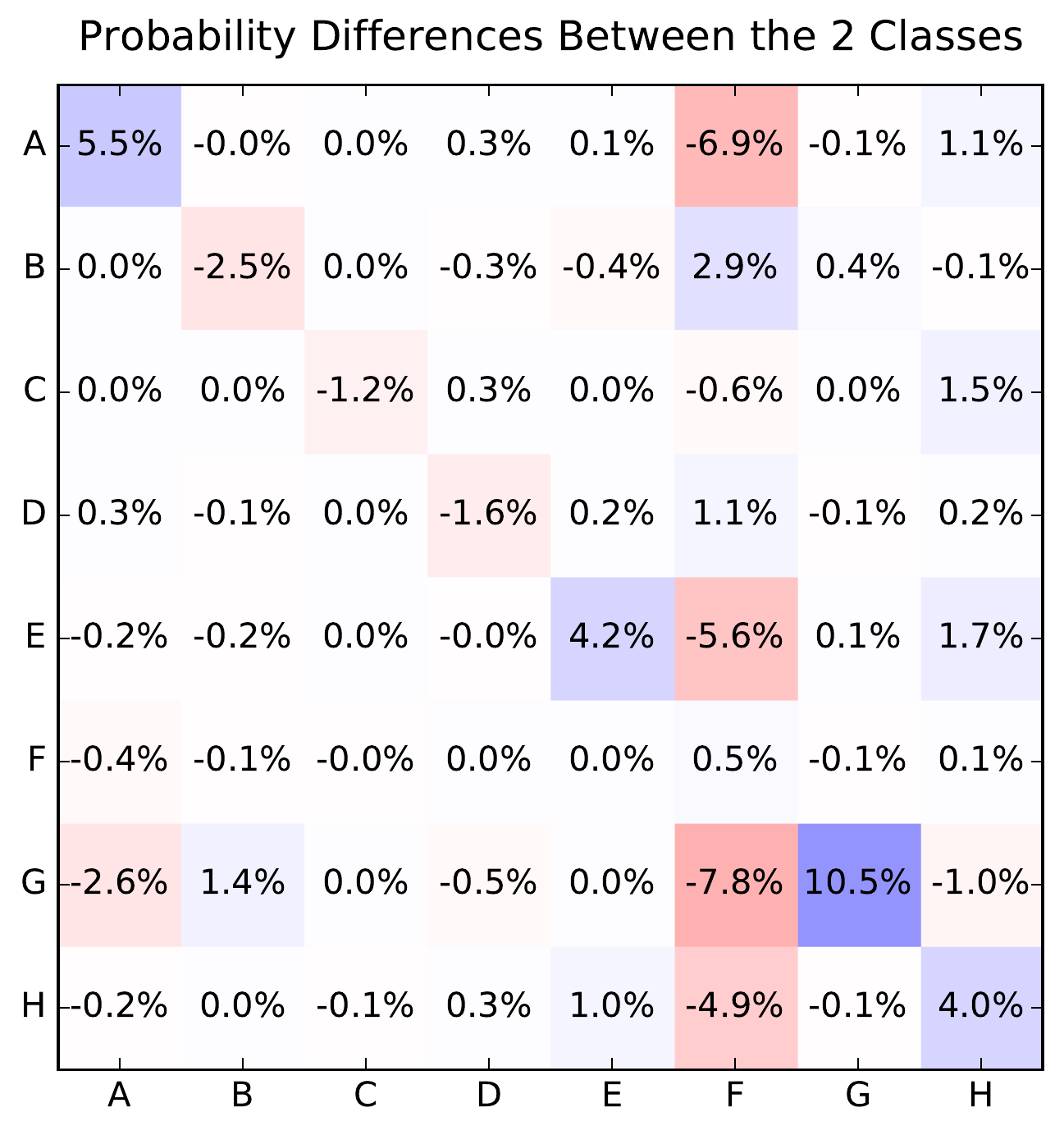}
      \caption{Transition matrices for Traffic Divider (Present vs Not Present). Classification accuracy: 56.6\%.}
      \label{fig:p04_traffic_flow_01}
    \end{figure}
    
    \begin{figure}[!h]
      \transallfig{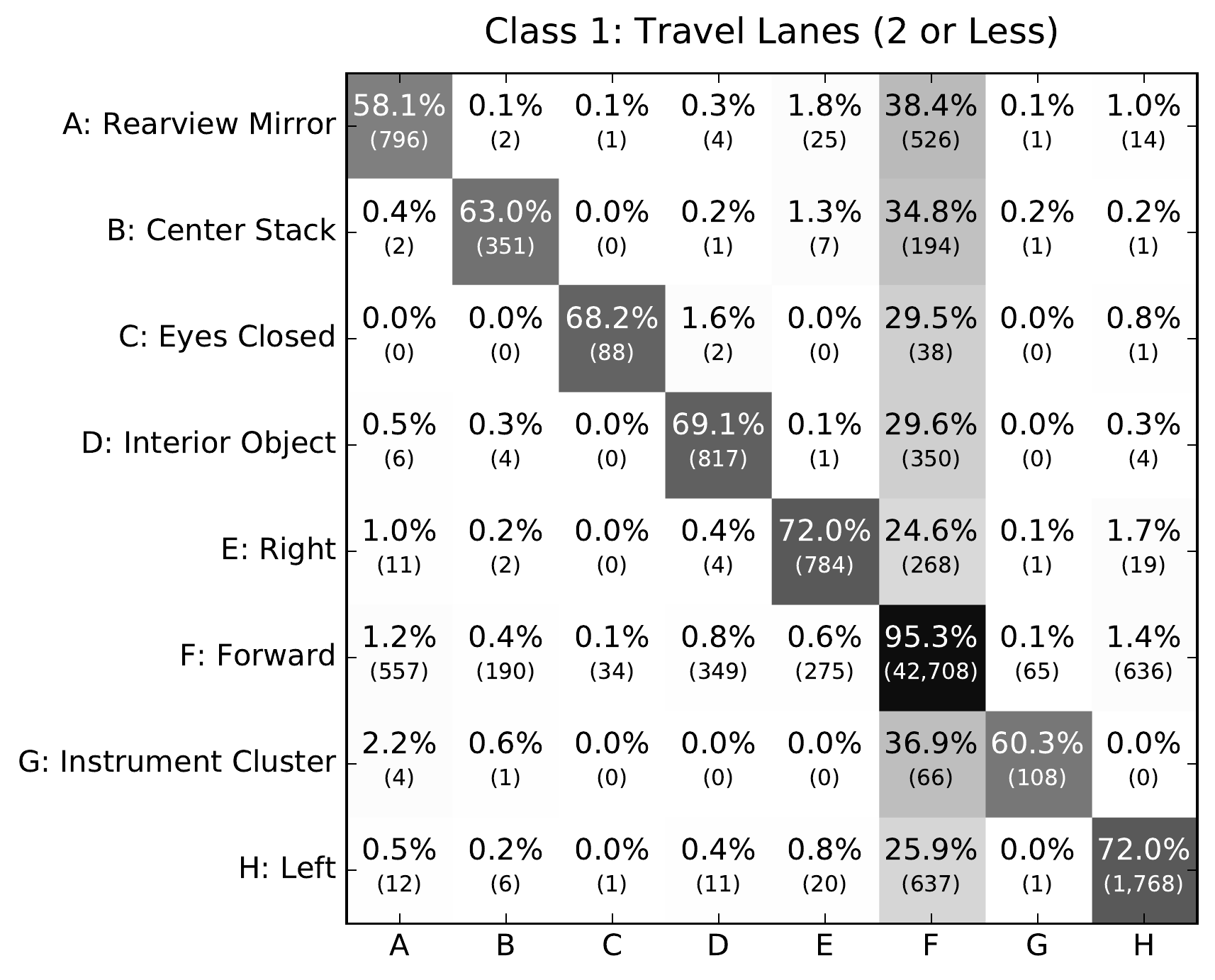}
      \transallfig{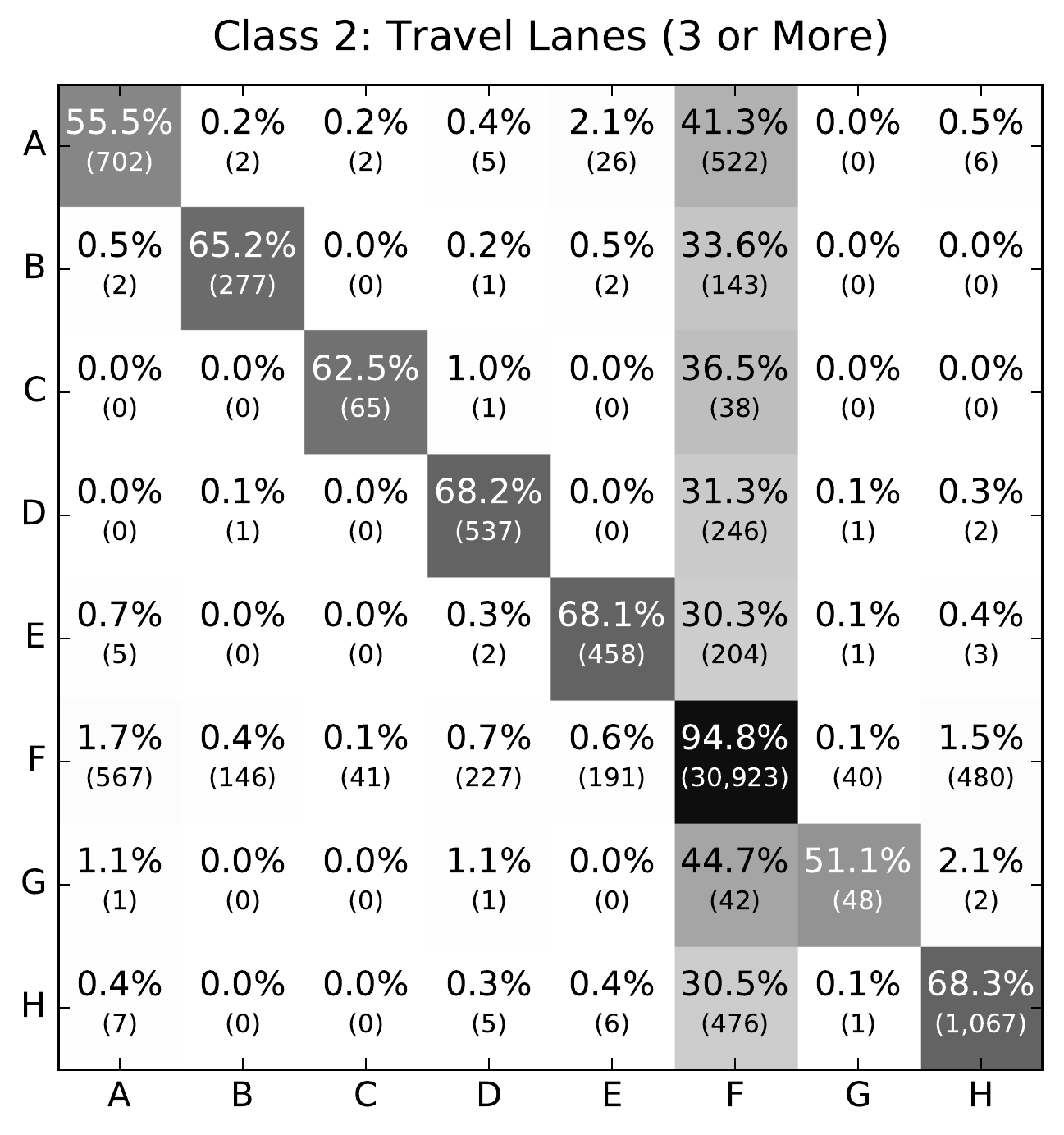}
      \transallfig{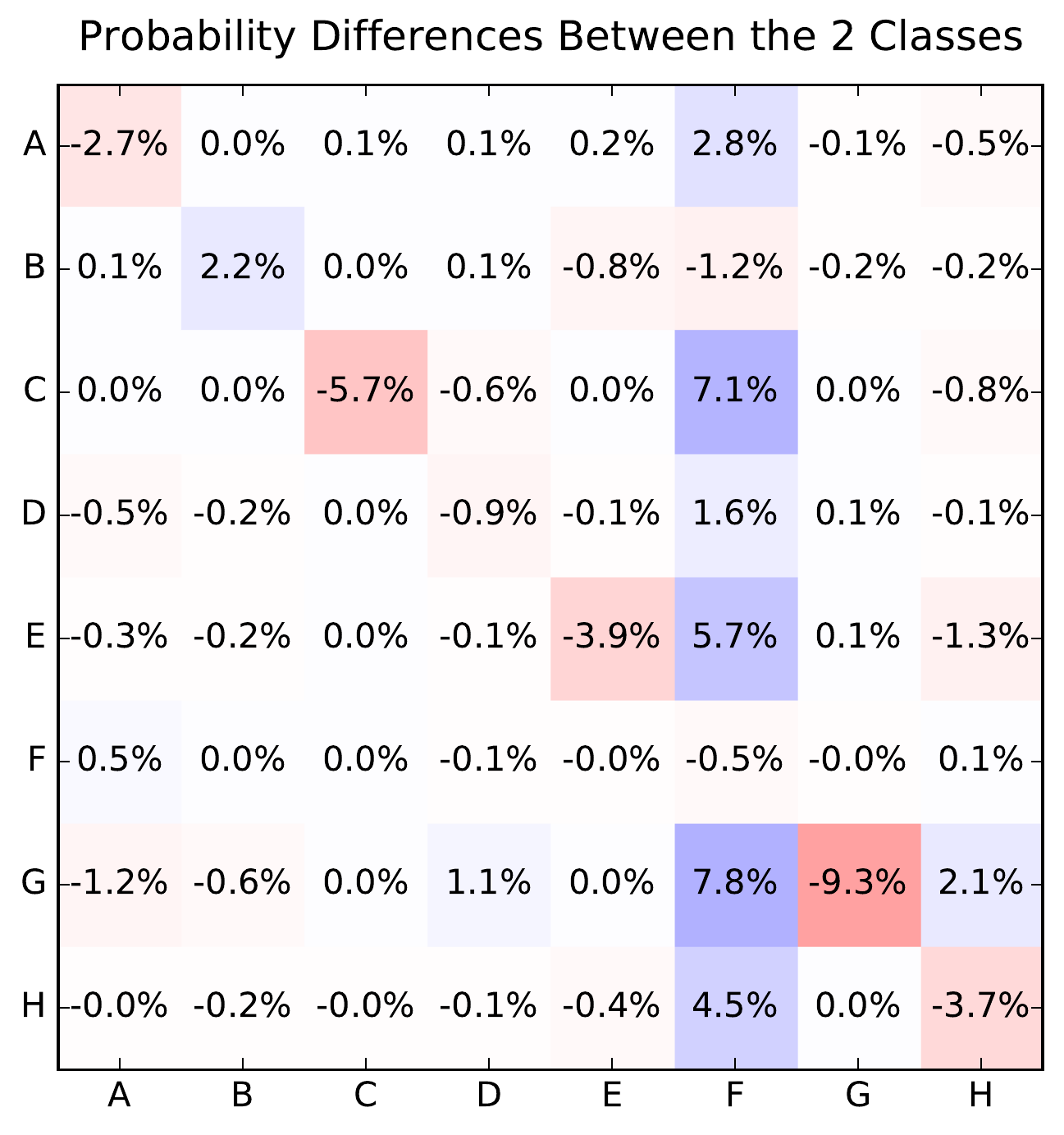}
      \caption{Transition matrices for Travel Lanes (2 or Less vs 3 or More). Classification accuracy: 57.7\%.}
      \label{fig:p05_travel_lanes_01}
    \end{figure}
    
    \begin{figure}[!h]
      \transallfig{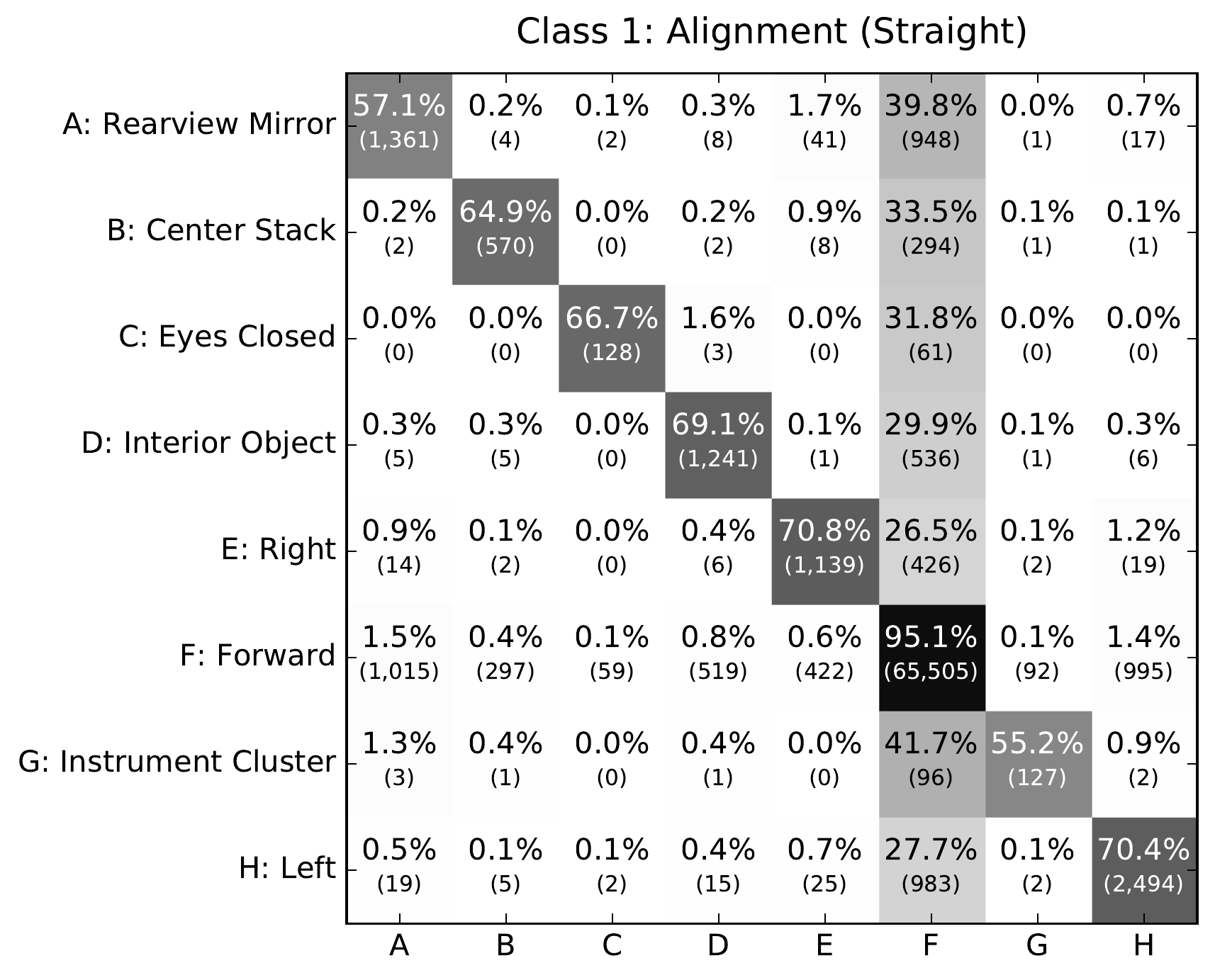}
      \transallfig{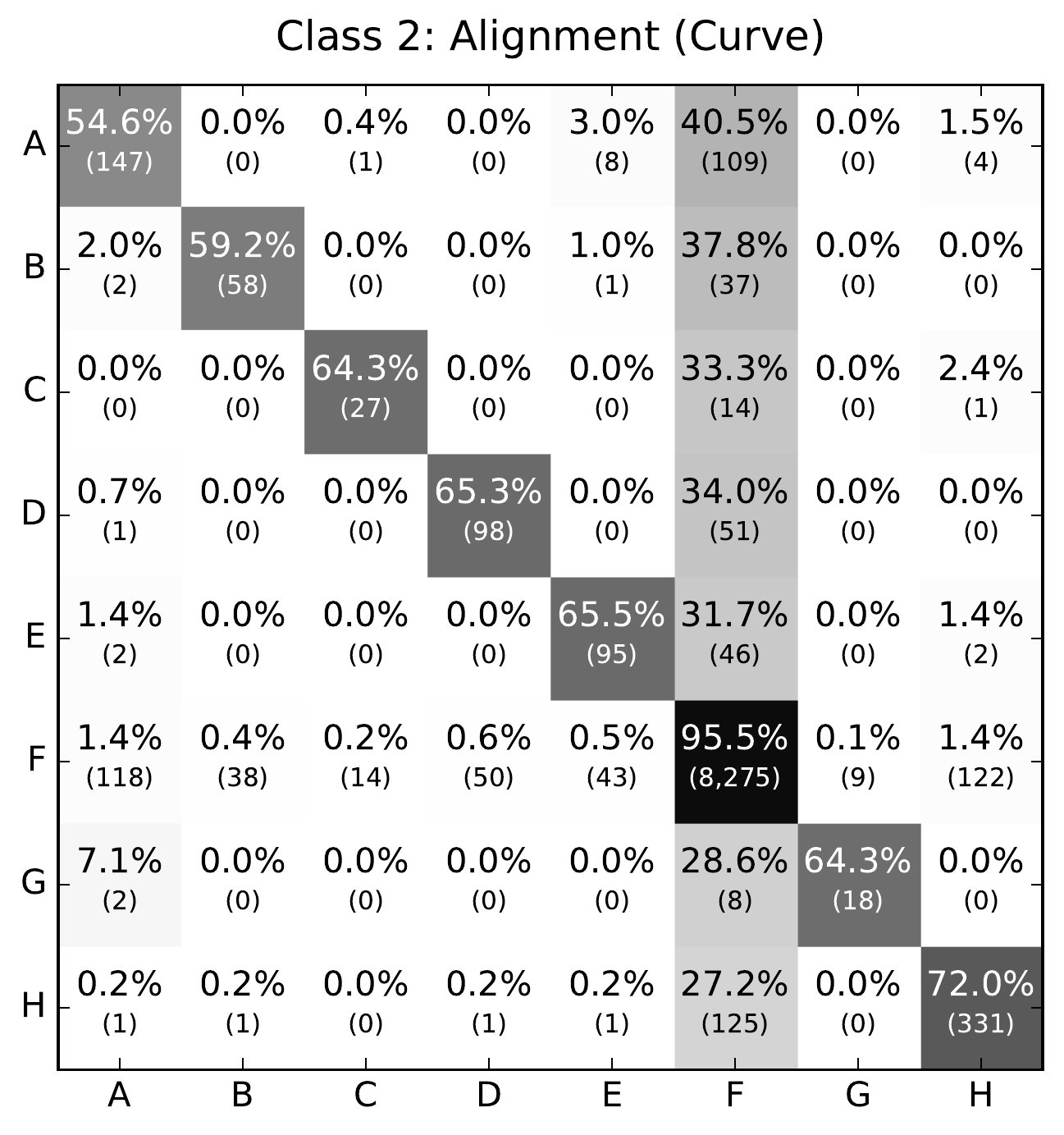}
      \transallfig{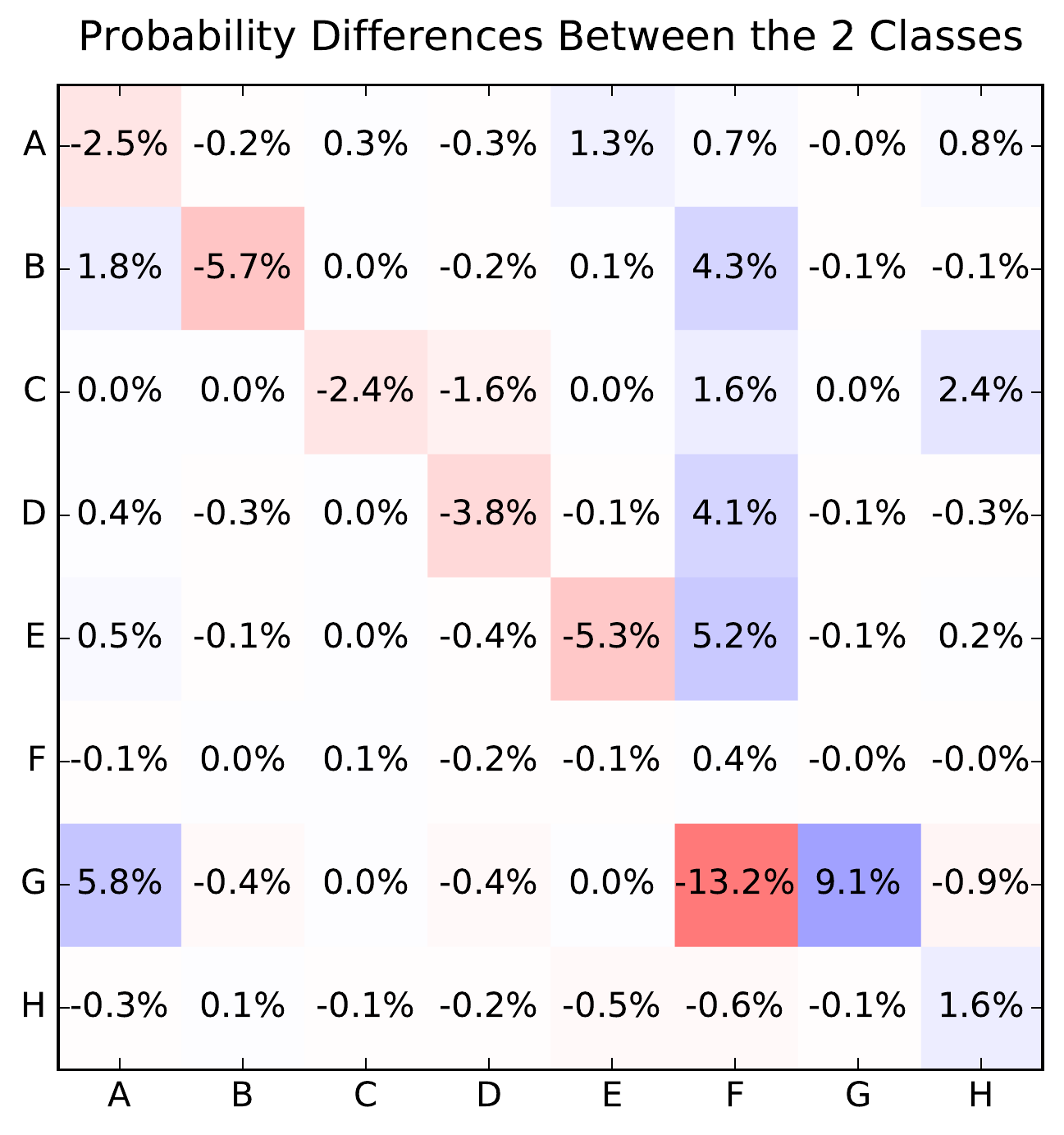}
      \caption{Transition matrices for Alignment (Straight vs Curve). Classification accuracy: 57.7\%.}
      \label{fig:p06_alignment_01}
    \end{figure}
    
    \begin{figure}[!h]
      \transallfig{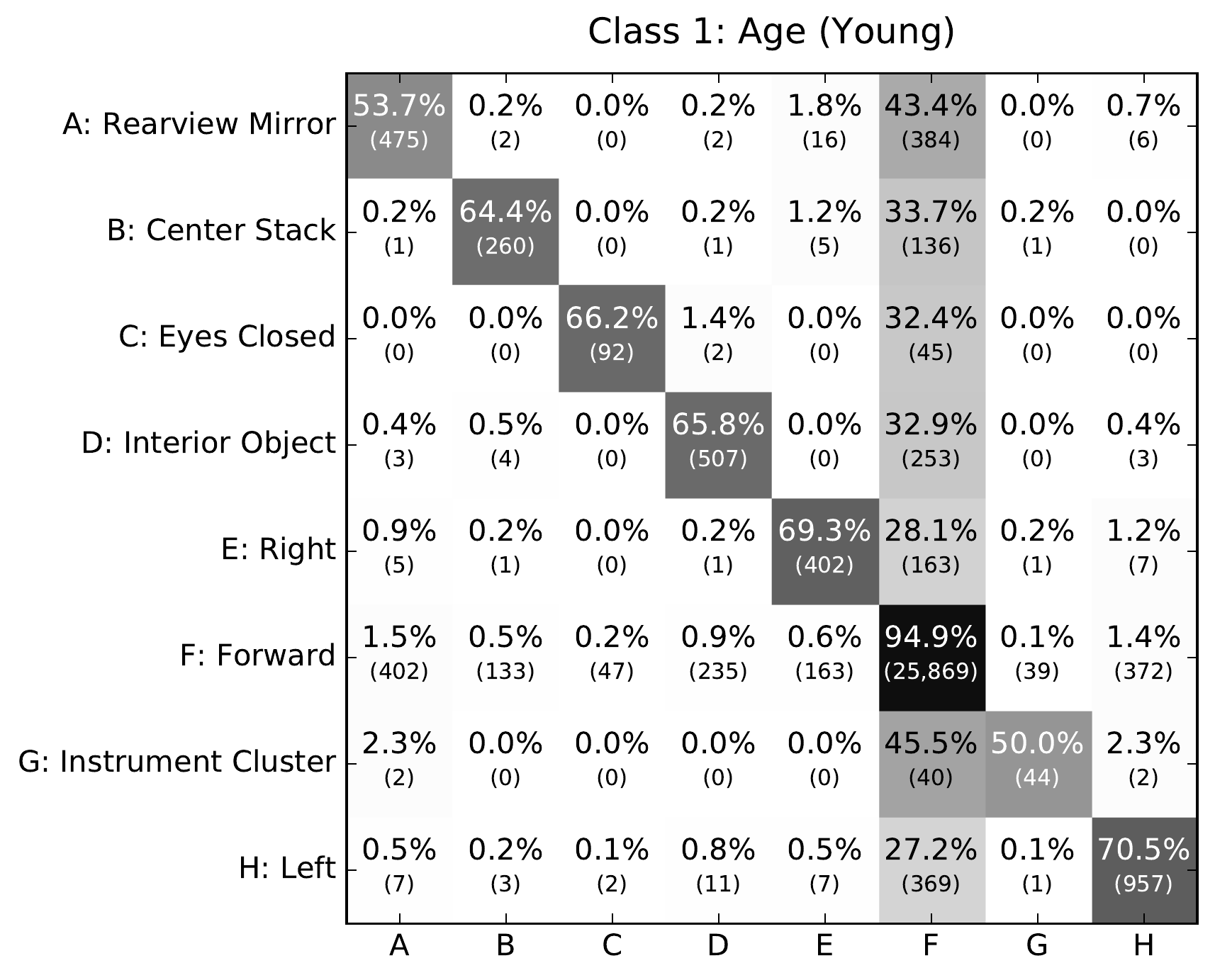}
      \transallfig{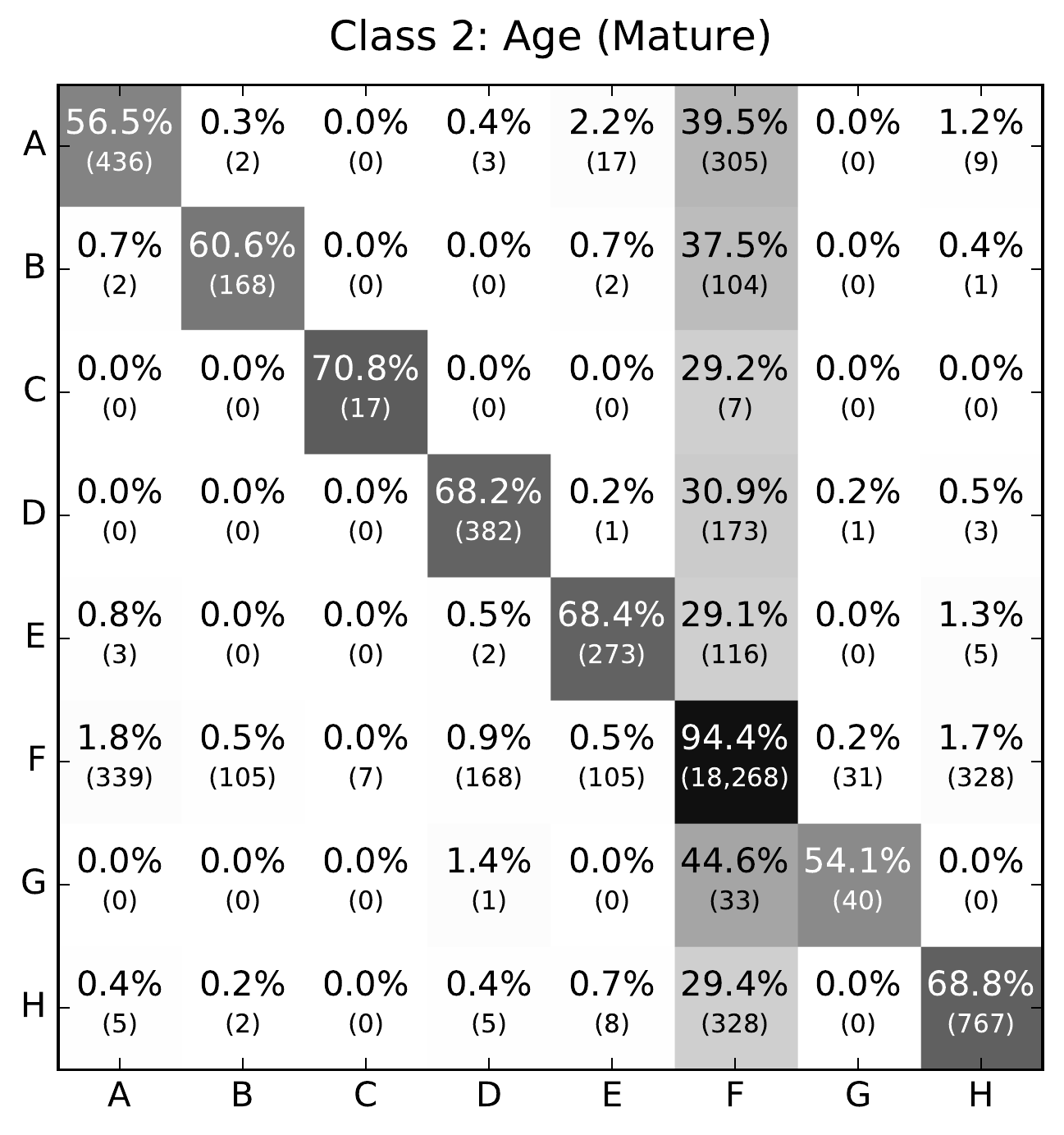}
      \transallfig{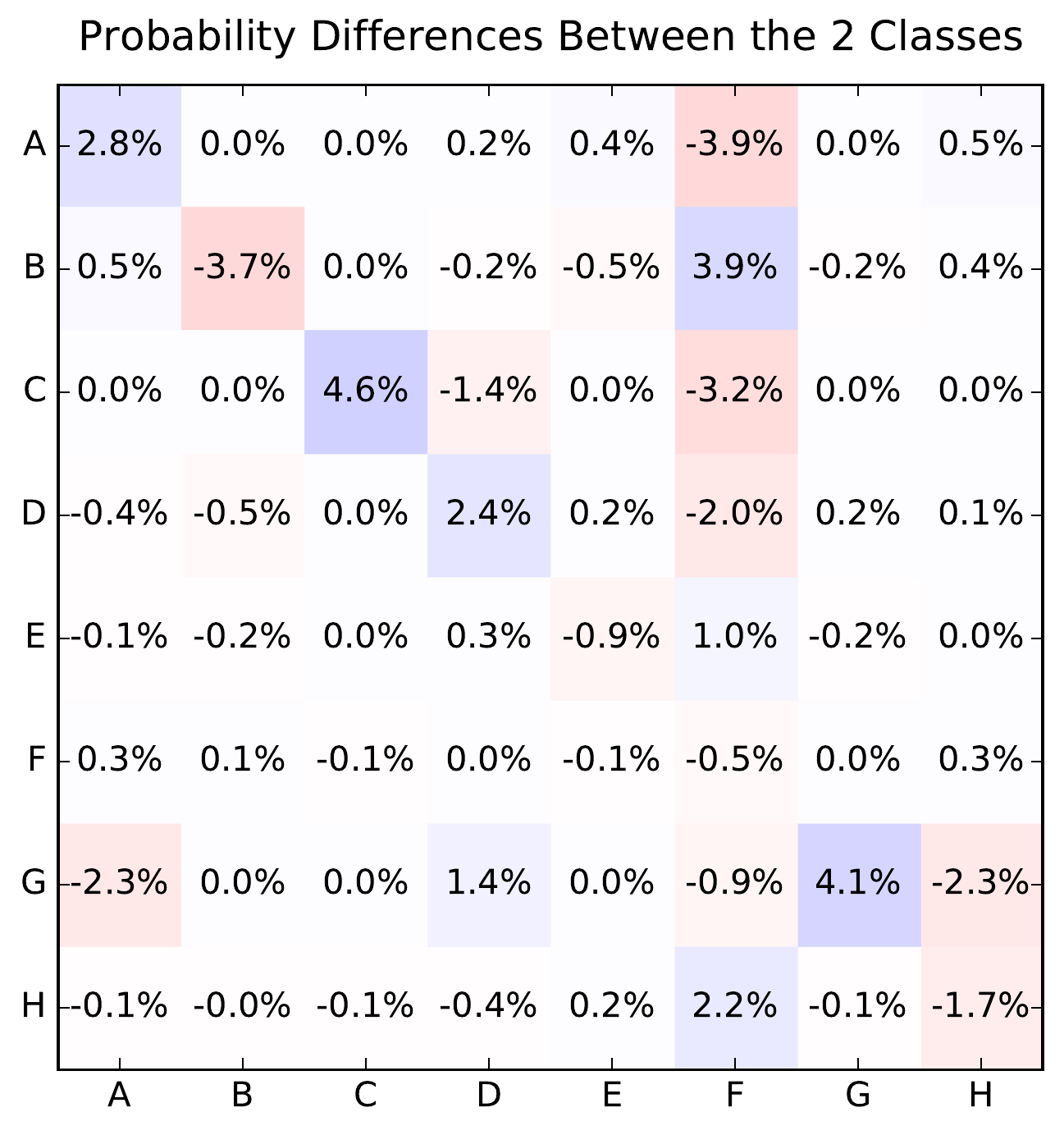}
      \caption{Transition matrices for Age (Young vs Mature). Classification accuracy: 58.3\%.}
      \label{fig:p07_age_02}
    \end{figure}
    
    \begin{figure}[!h]
      \transallfig{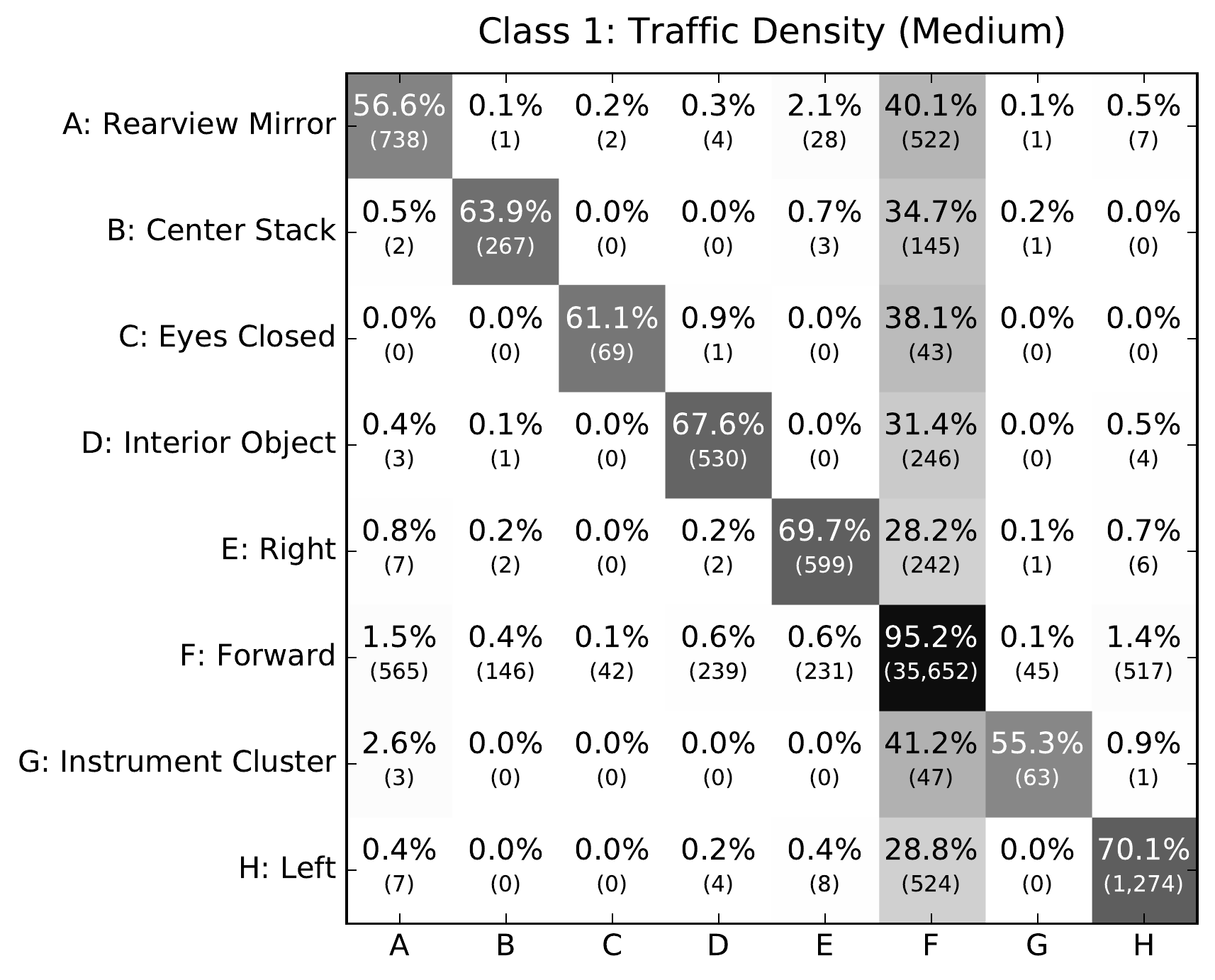}
      \transallfig{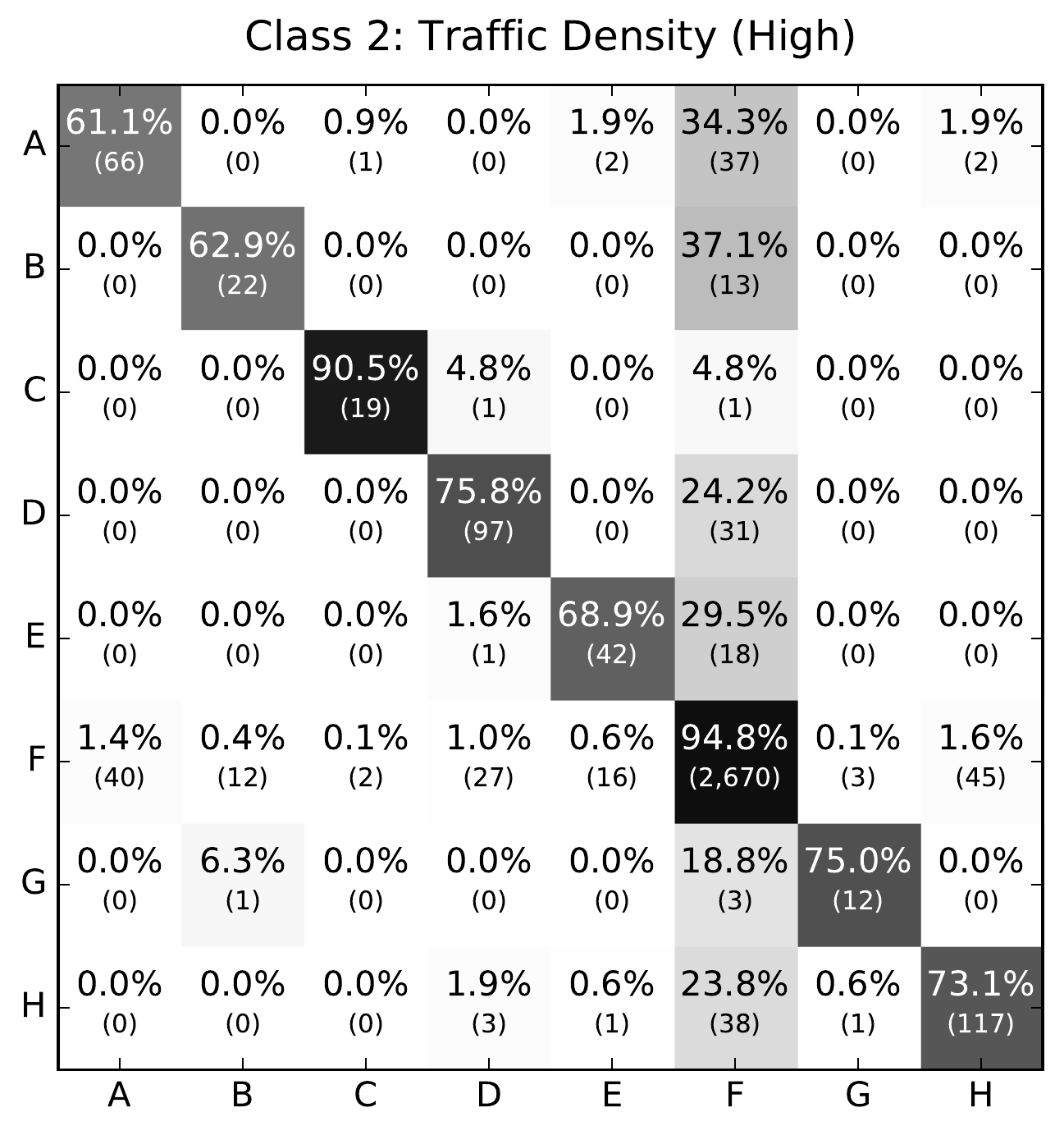}
      \transallfig{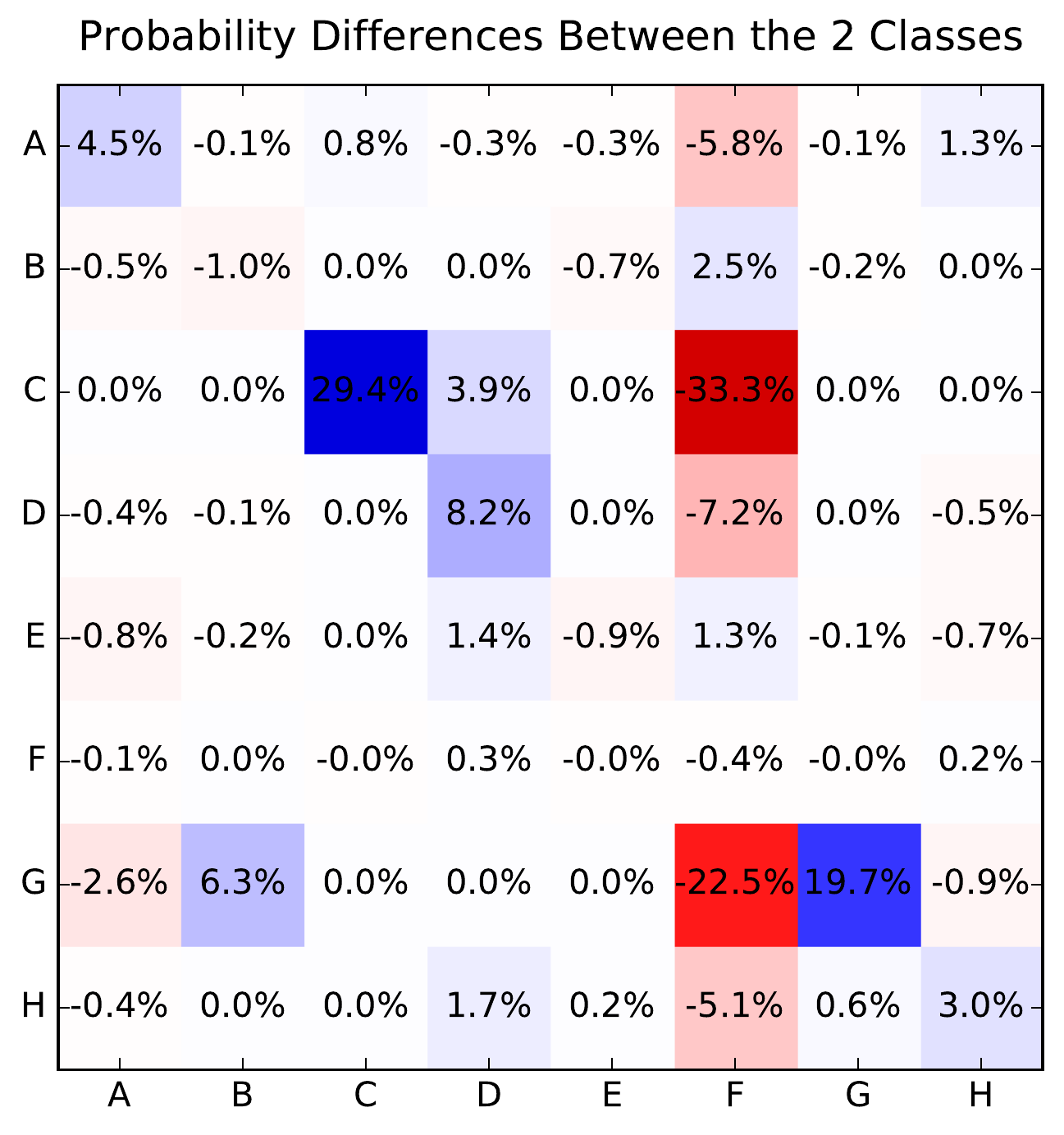}
      \caption{Transition matrices for Traffic Density (Medium vs High). Classification accuracy: 58.6\%.}
      \label{fig:p08_traffic_density_12}
    \end{figure}
    
    \begin{figure}[!h]
      \transallfig{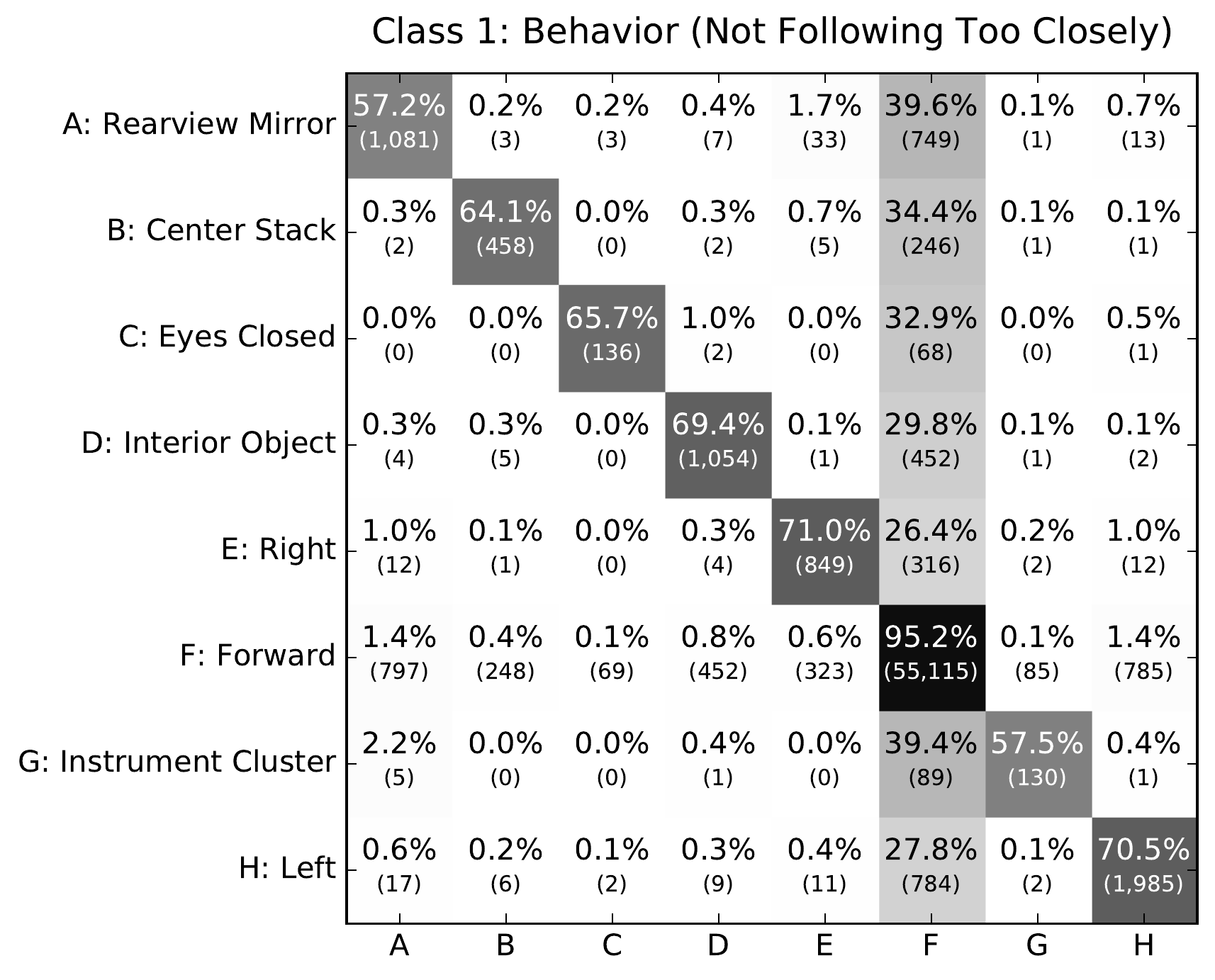}
      \transallfig{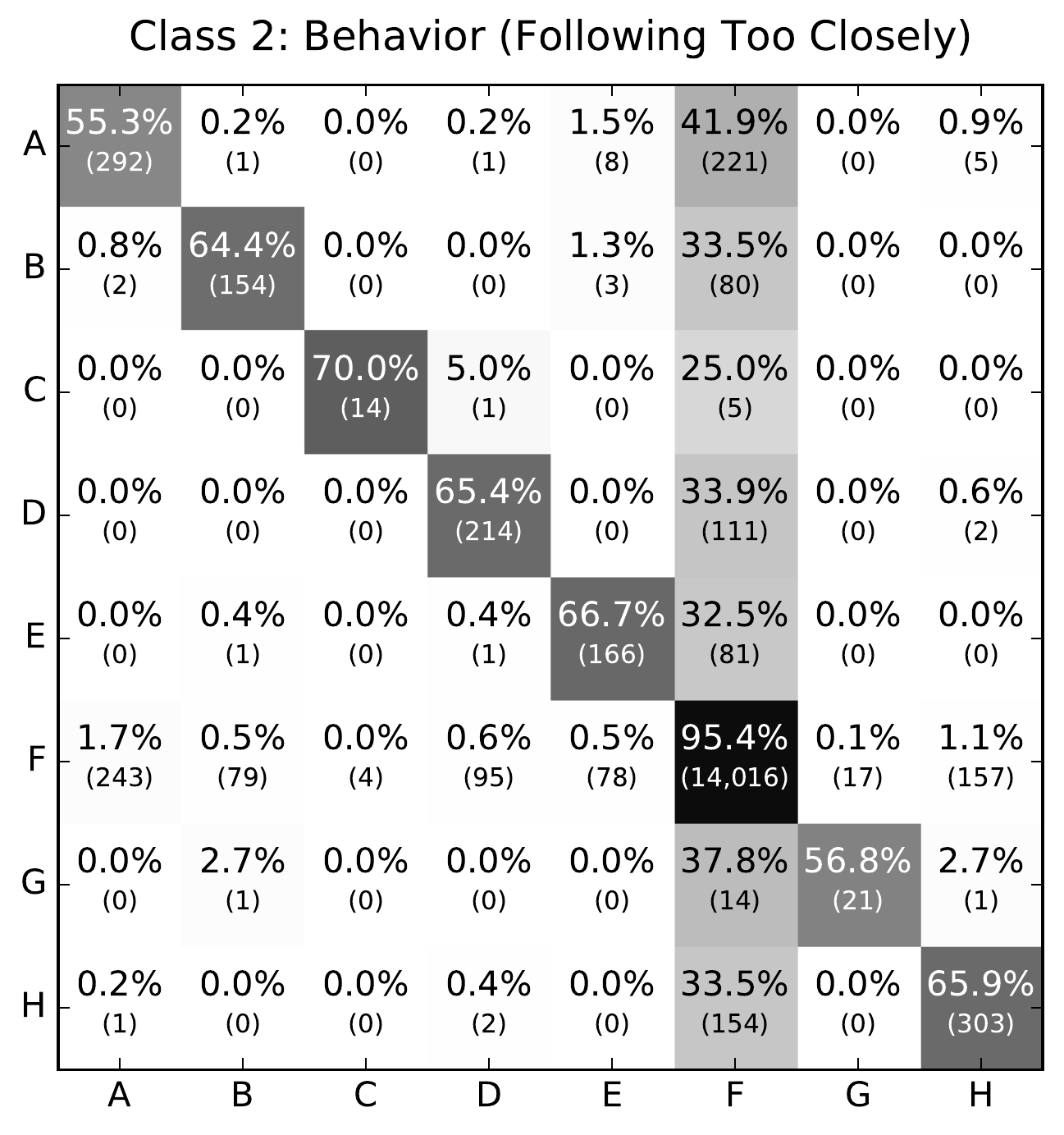}
      \transallfig{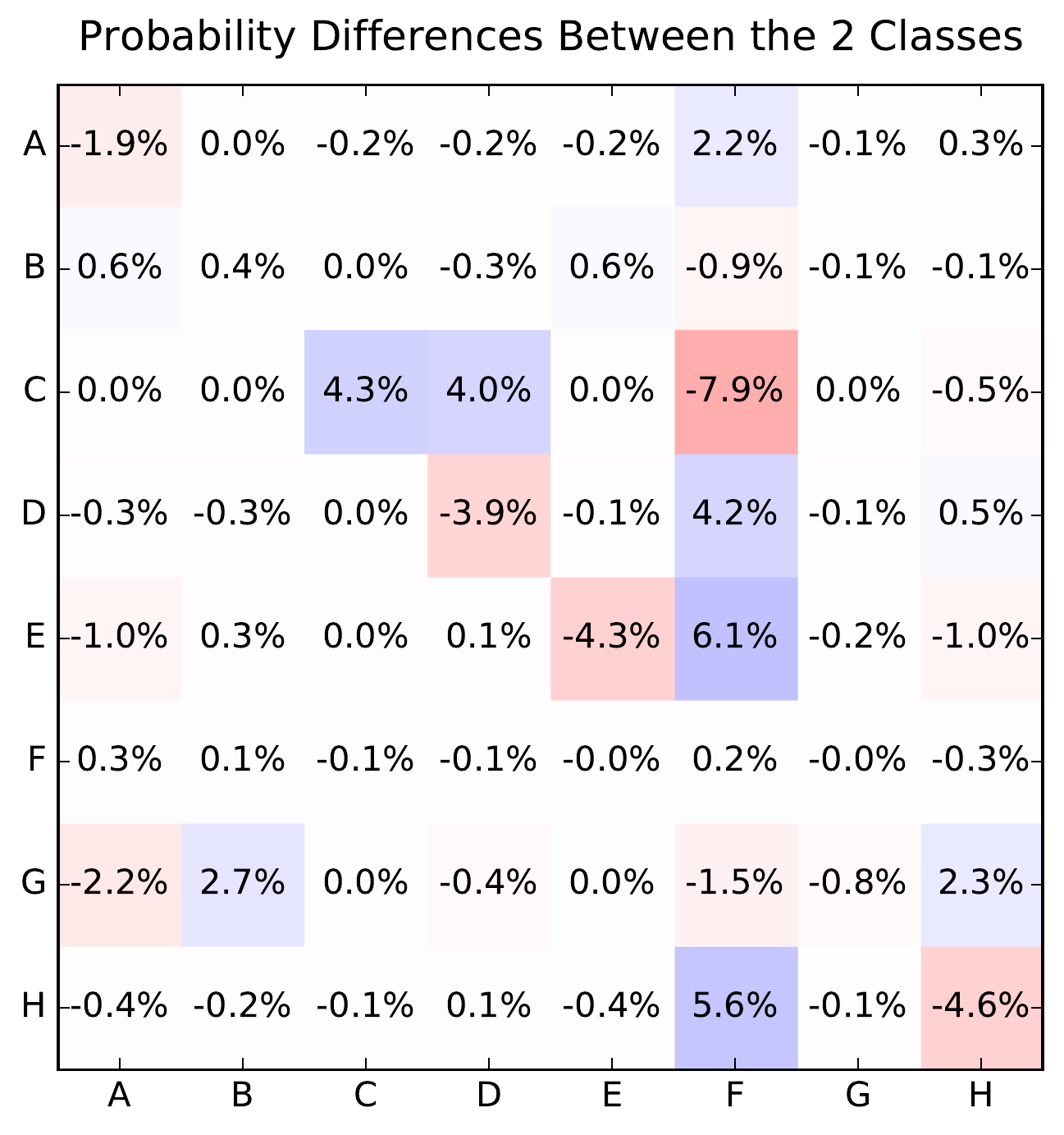}
      \caption{Transition matrices for Behavior (Following Too Closely). Classification accuracy: 59.1\%.}
      \label{fig:p09_behavior_01}
    \end{figure}
    
    \begin{figure}[!h]
      \transallfig{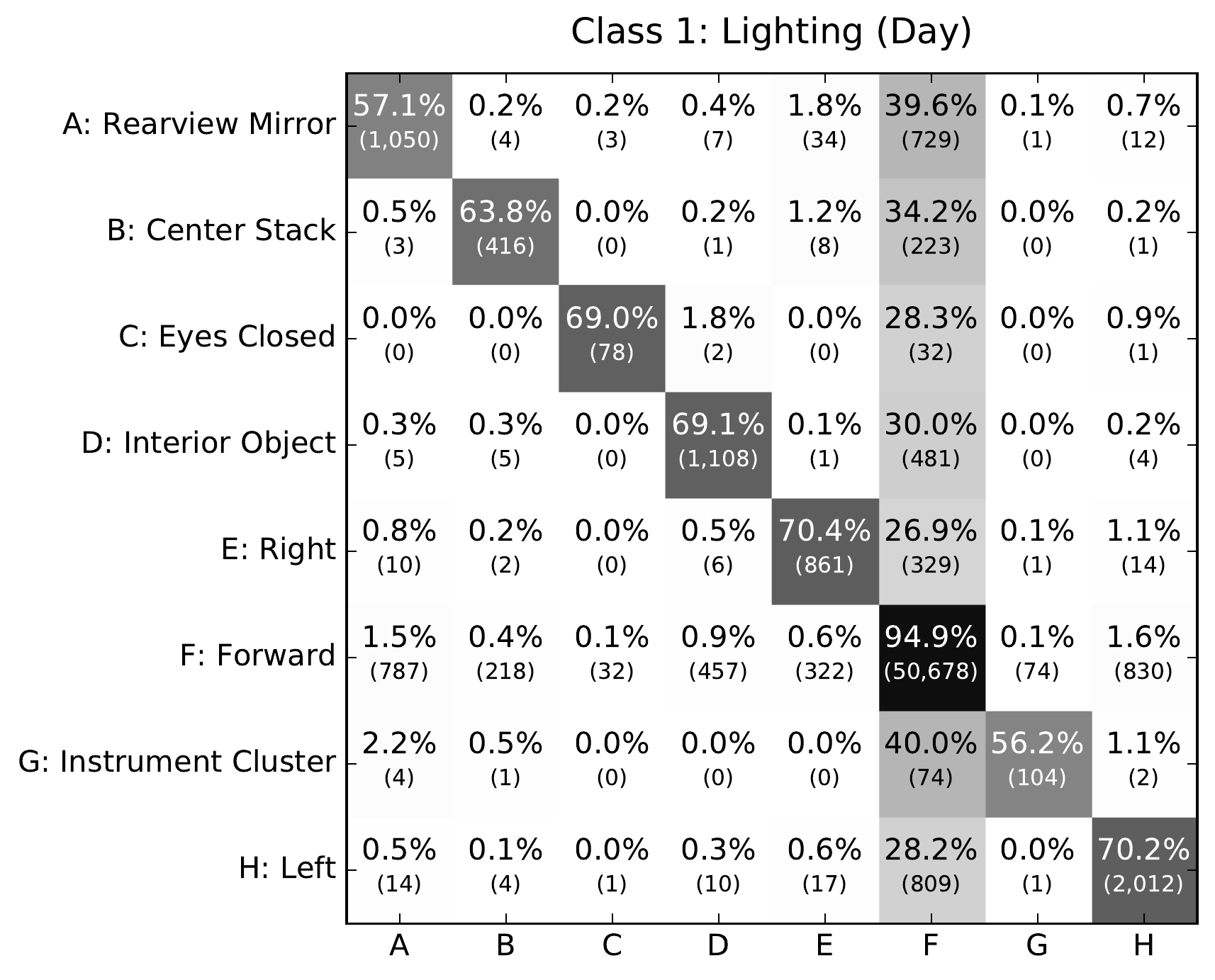}
      \transallfig{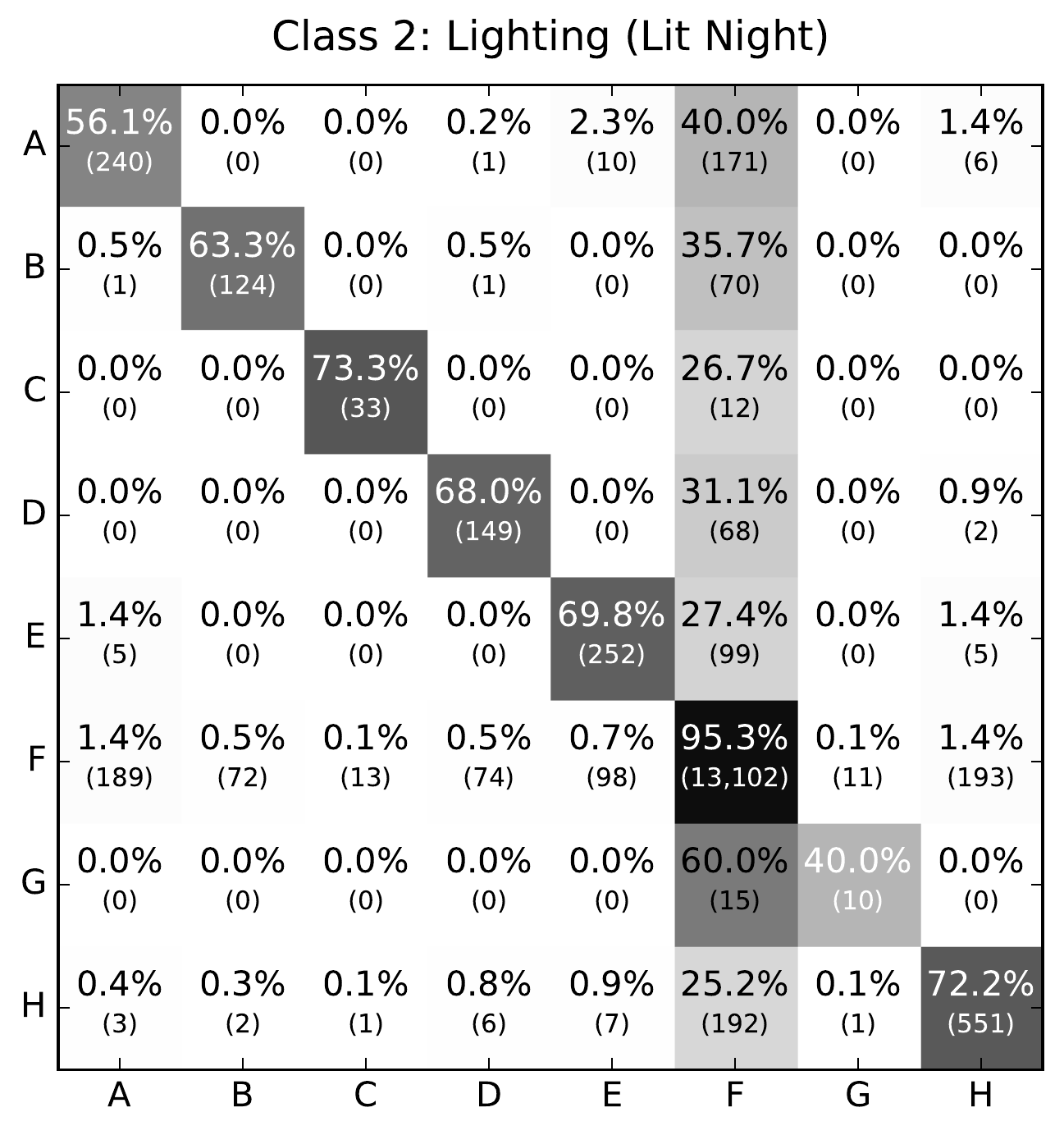}
      \transallfig{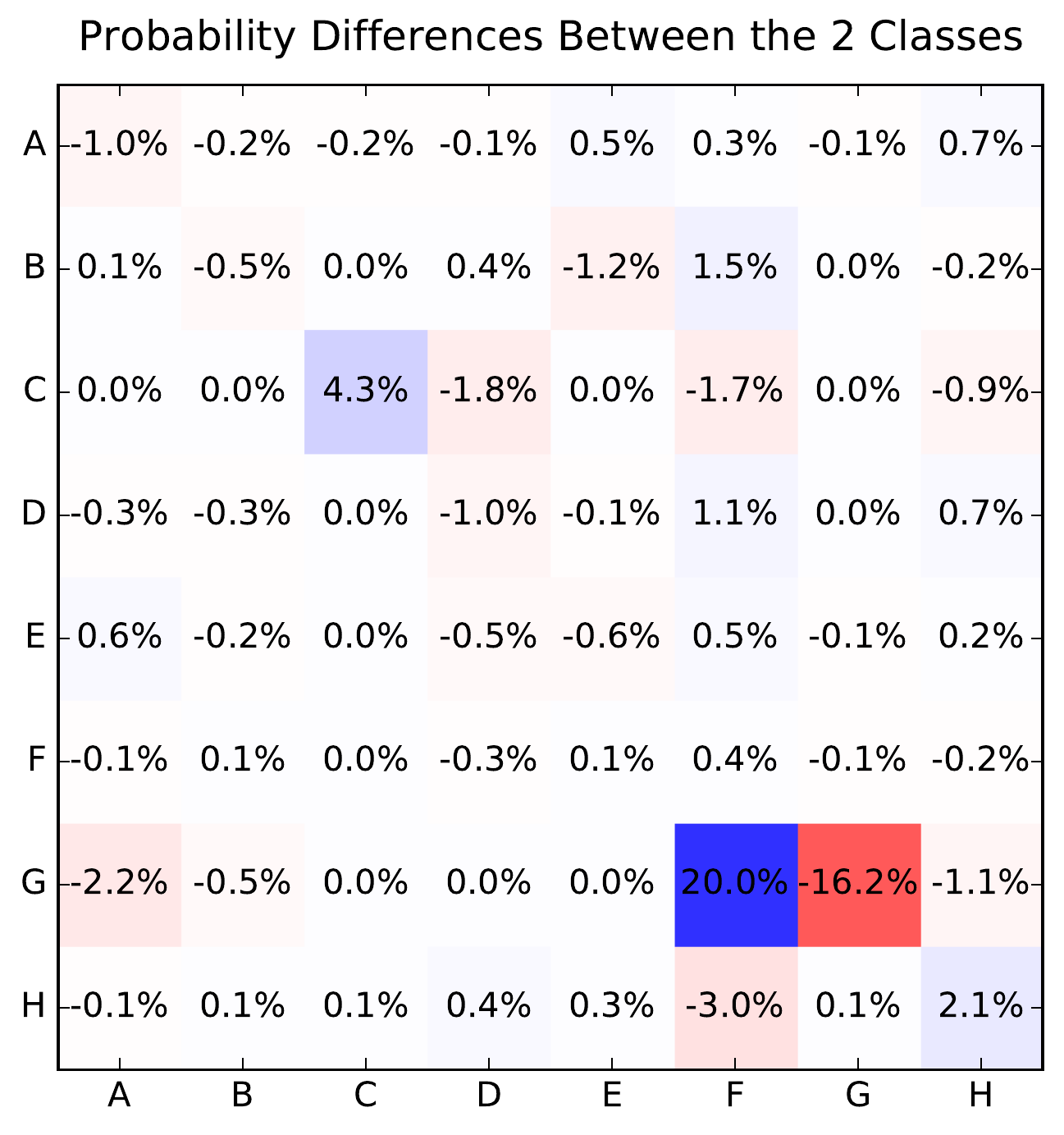}
      \caption{Transition matrices for Lighting (Day vs Lit Night). Classification accuracy: 59.2\%.}
      \label{fig:p10_lighting_01}
    \end{figure}
    
    \begin{figure}[!h]
      \transallfig{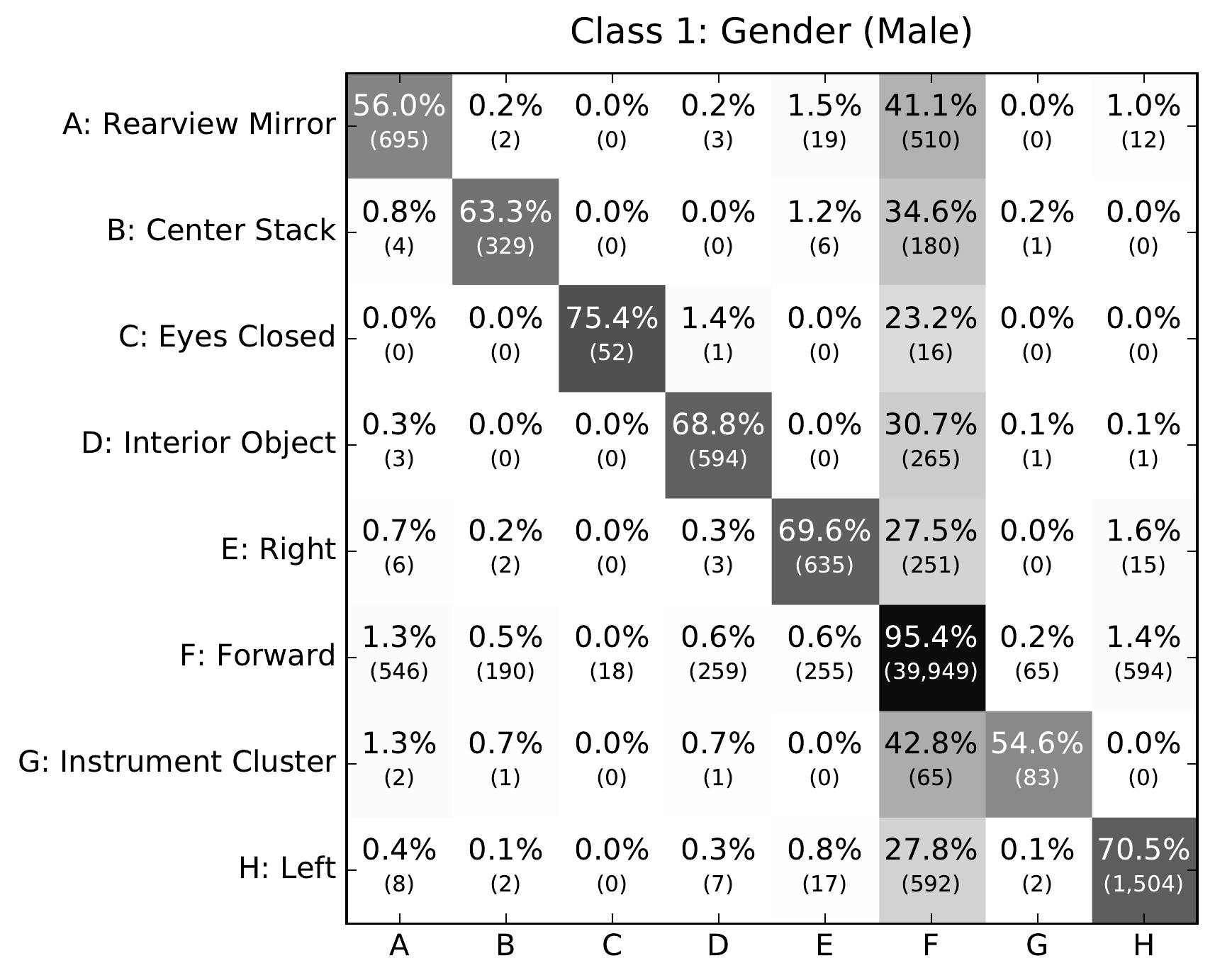}
      \transallfig{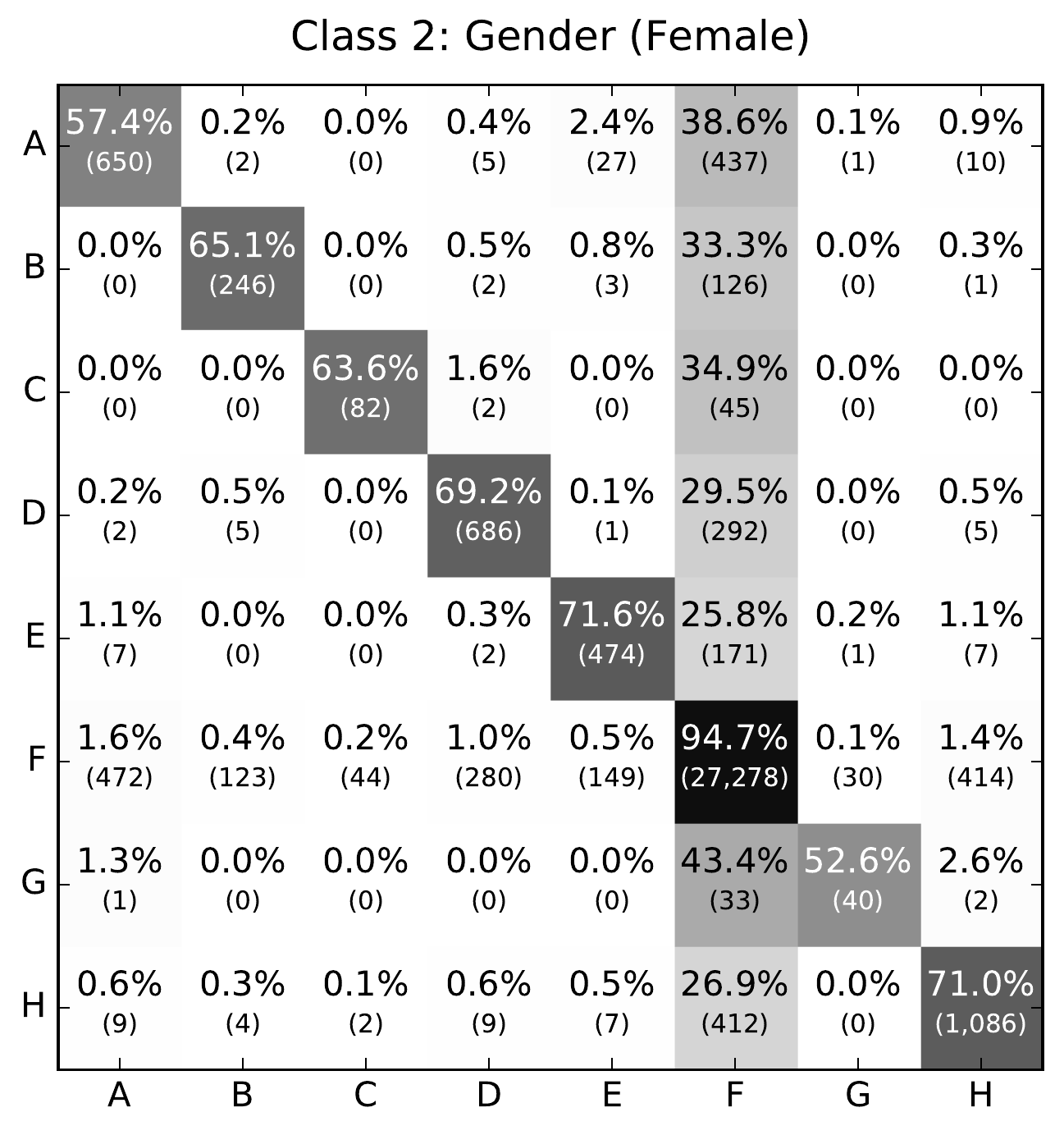}
      \transallfig{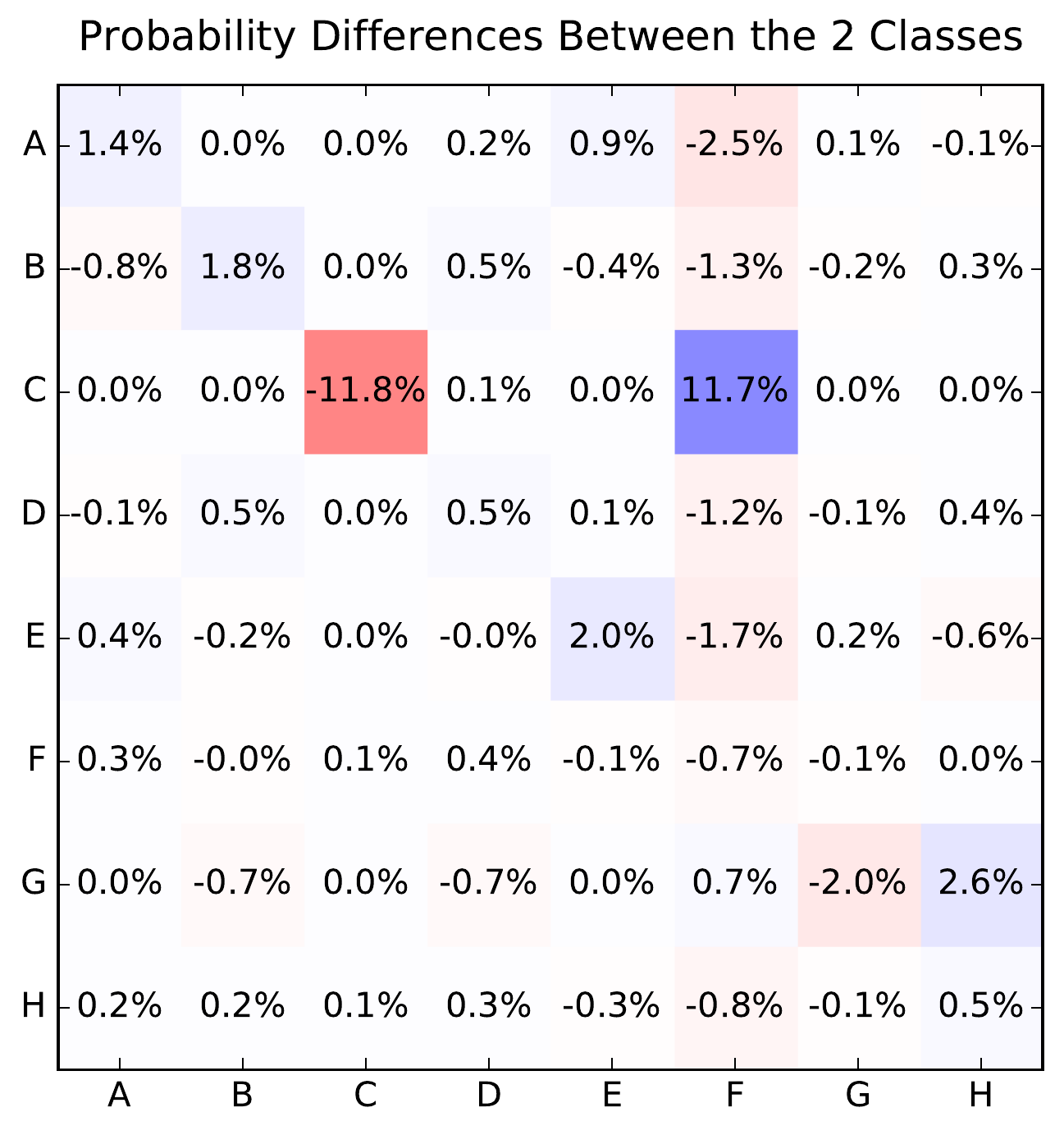}
      \caption{Transition matrices for Gender (Male vs Female). Classification accuracy: 59.7\%.}
      \label{fig:p11_gender_01}
    \end{figure}
    
    \begin{figure}[!h]
      \transallfig{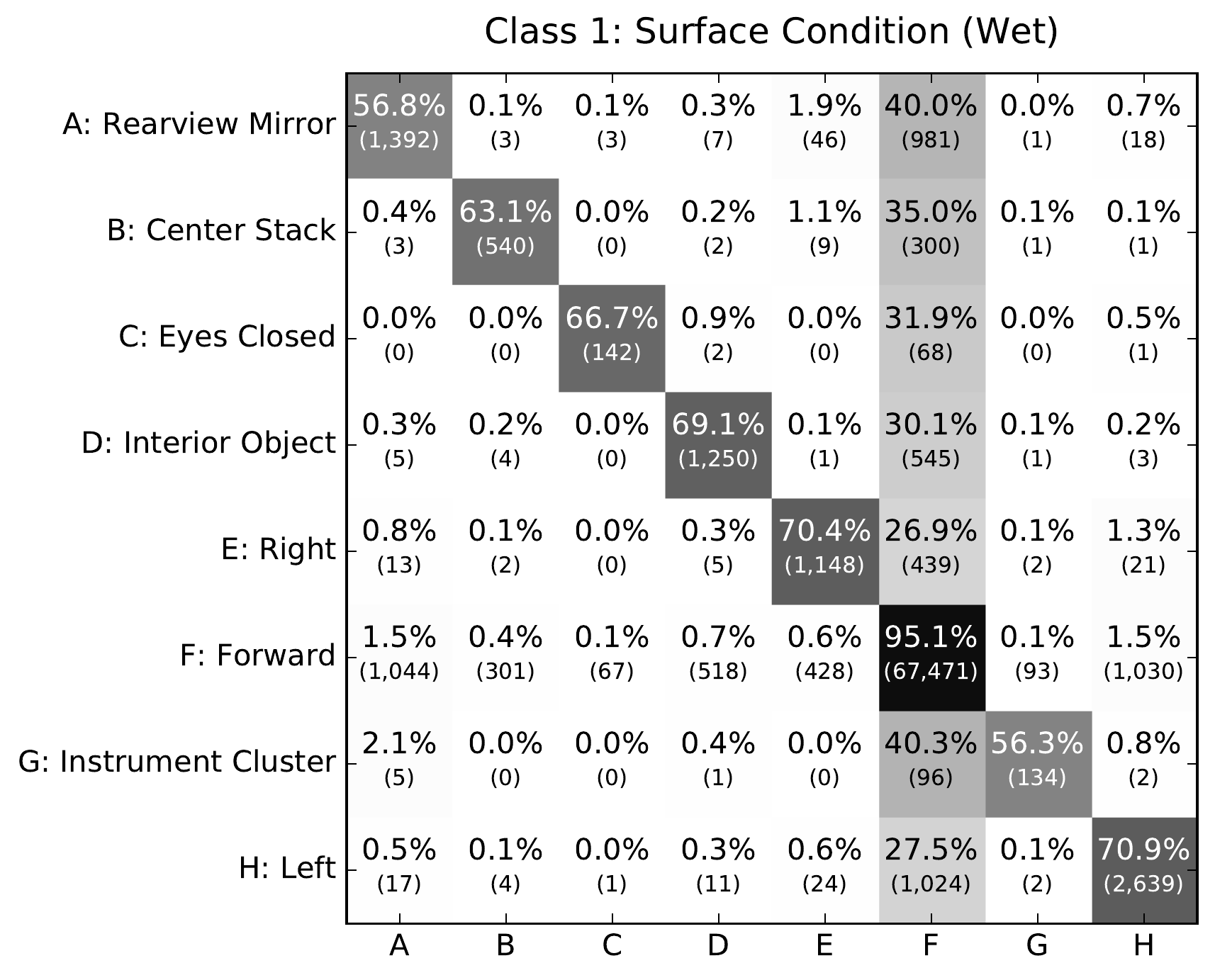}
      \transallfig{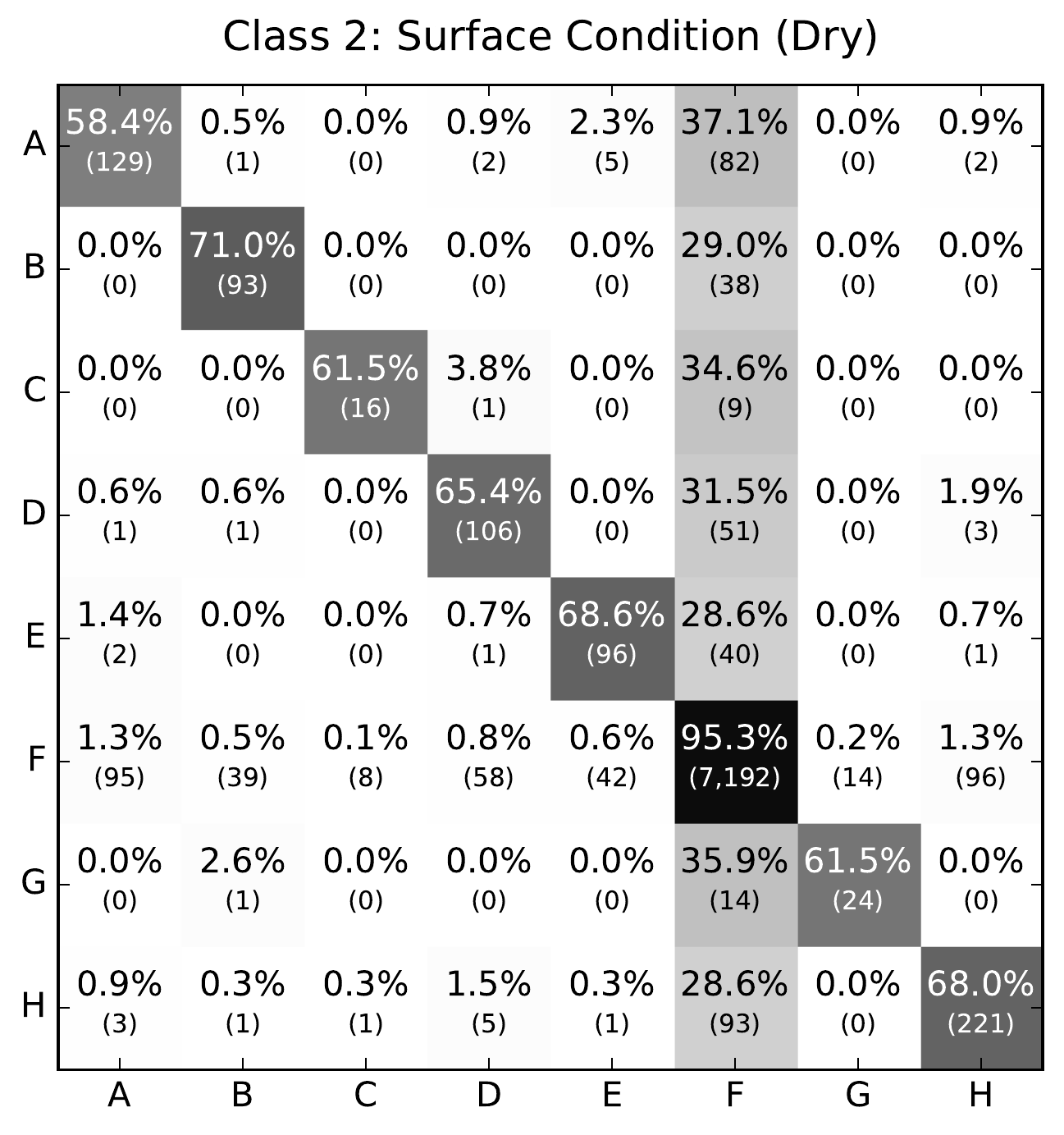}
      \transallfig{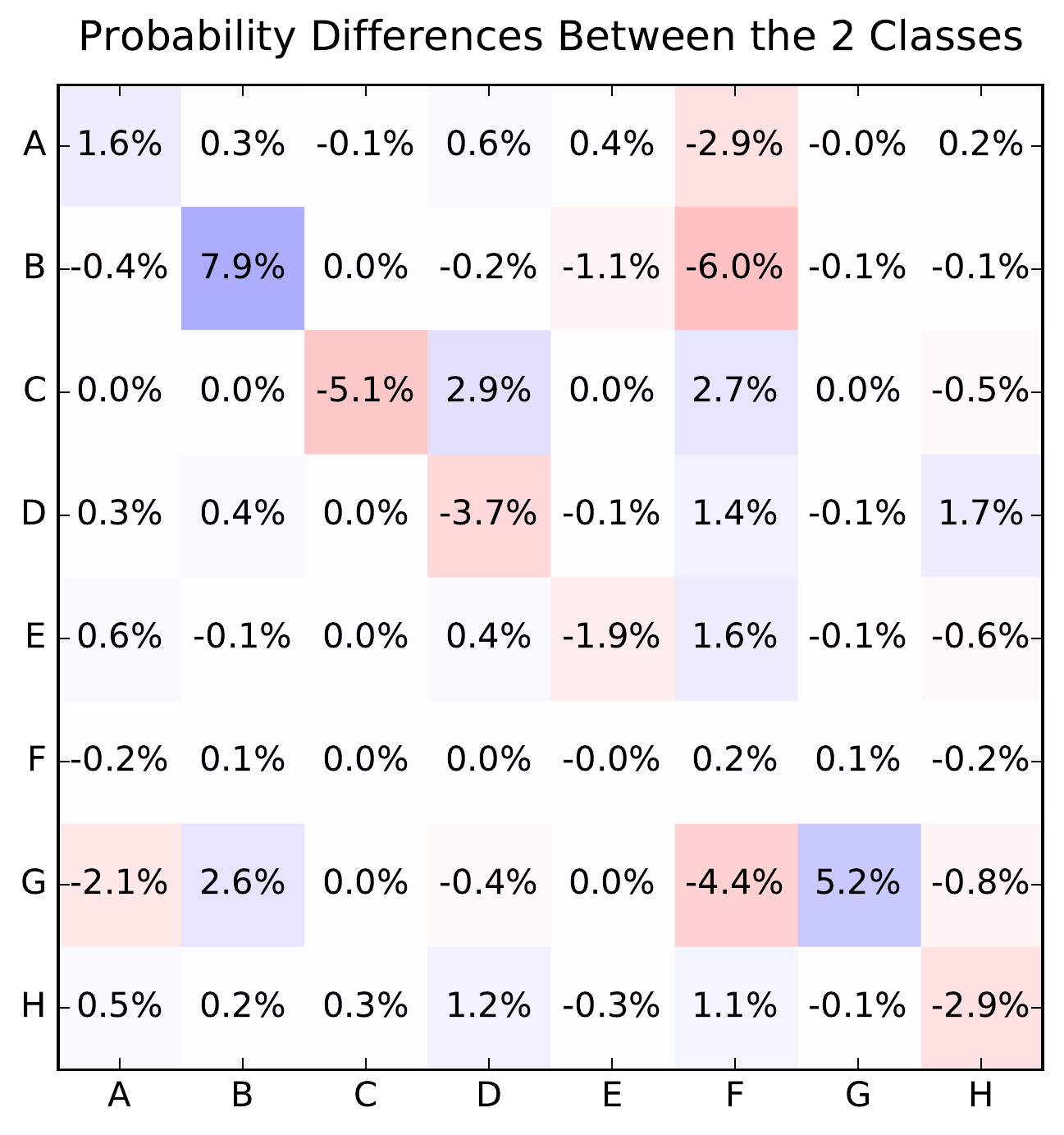}
      \caption{Transition matrices for Surface Condition (Wet vs Dry). Classification accuracy: 60.3\%.}
      \label{fig:p12_surface_condition_01}
    \end{figure}
    
    \begin{figure}[!h]
      \transallfig{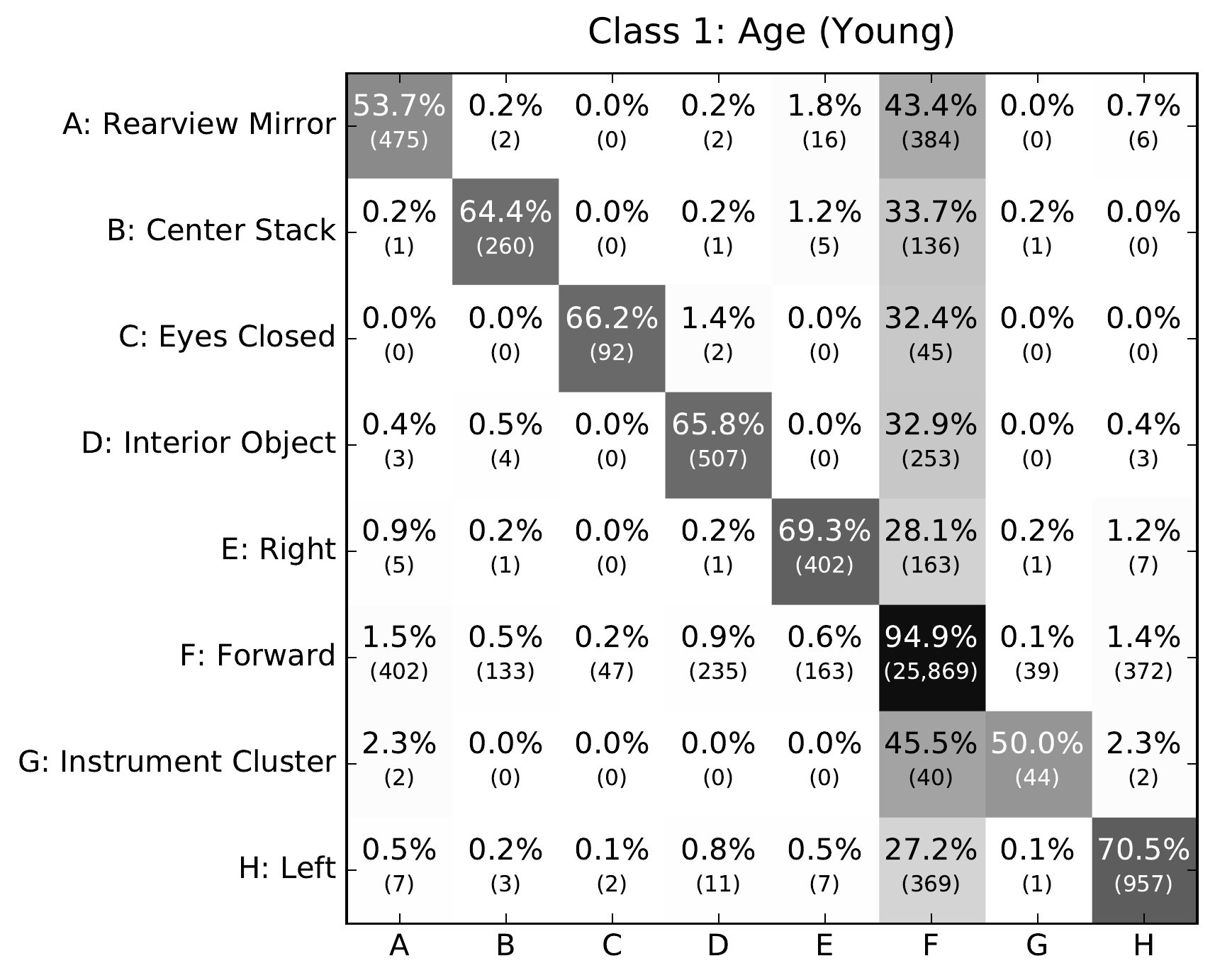}
      \transallfig{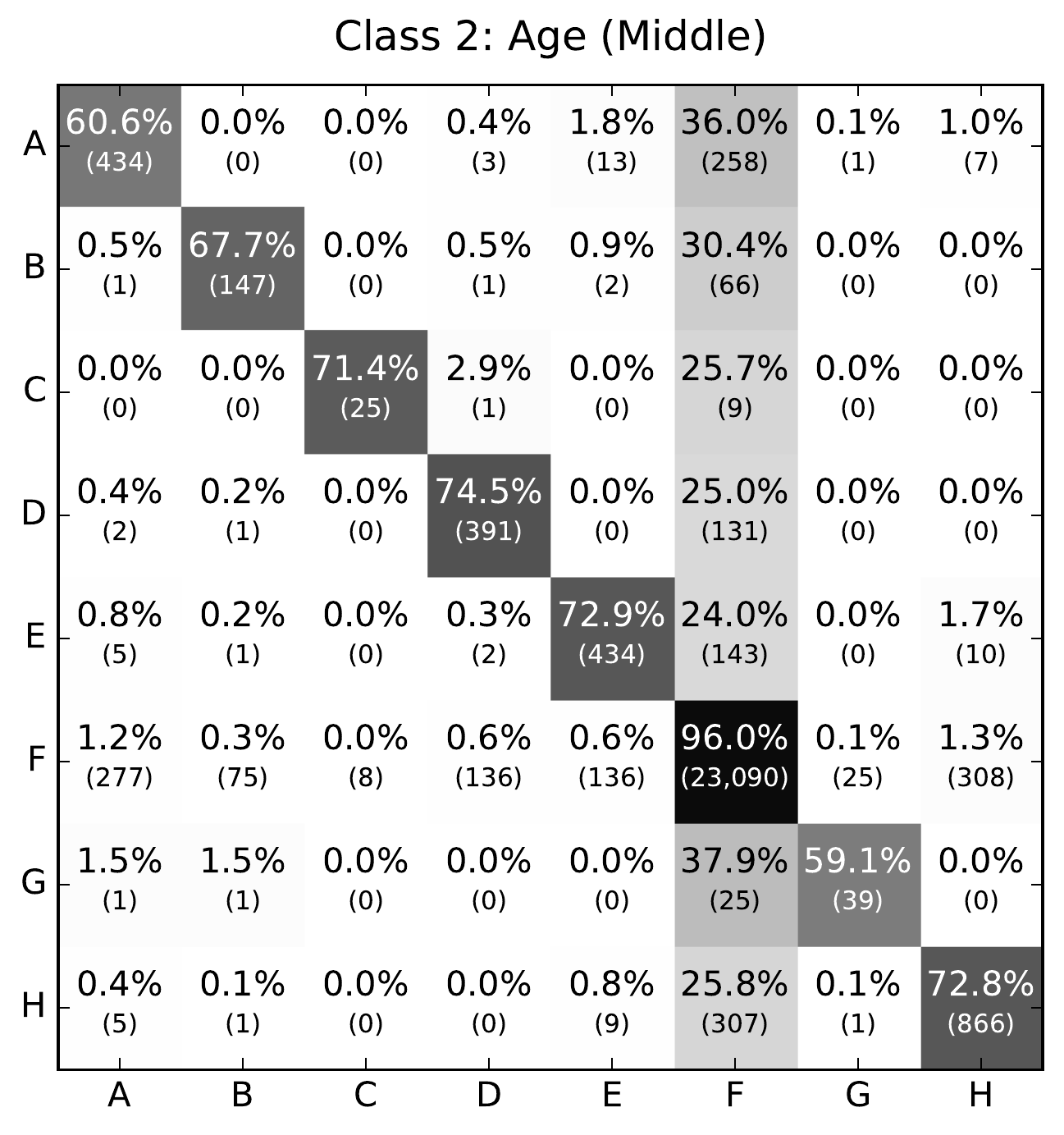}
      \transallfig{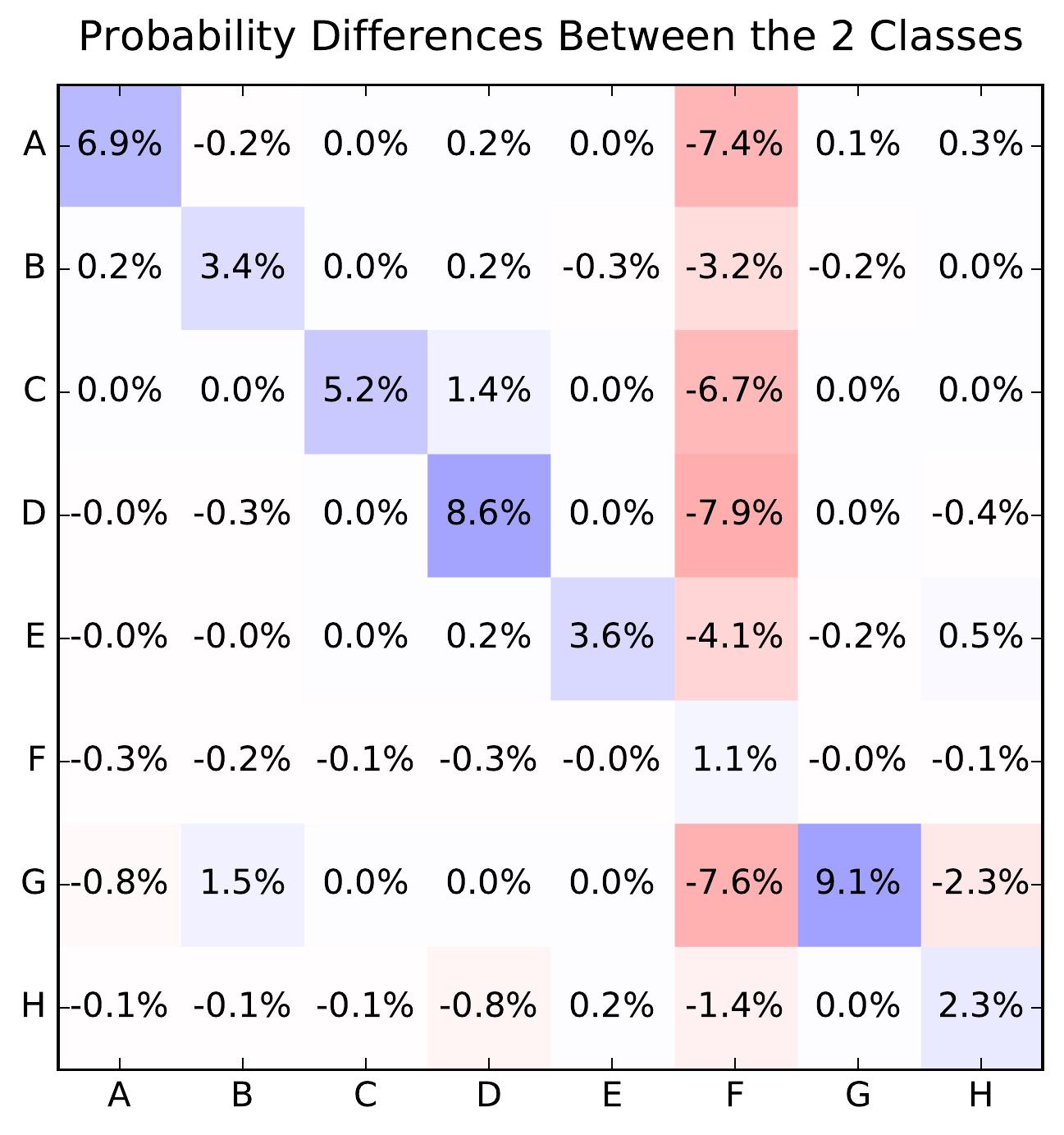}
      \caption{Transition matrices for Age (Young vs Middle). Classification accuracy: 60.7\%.}
      \label{fig:p13_age_01}
    \end{figure}
    
    \begin{figure}[!h]
      \transallfig{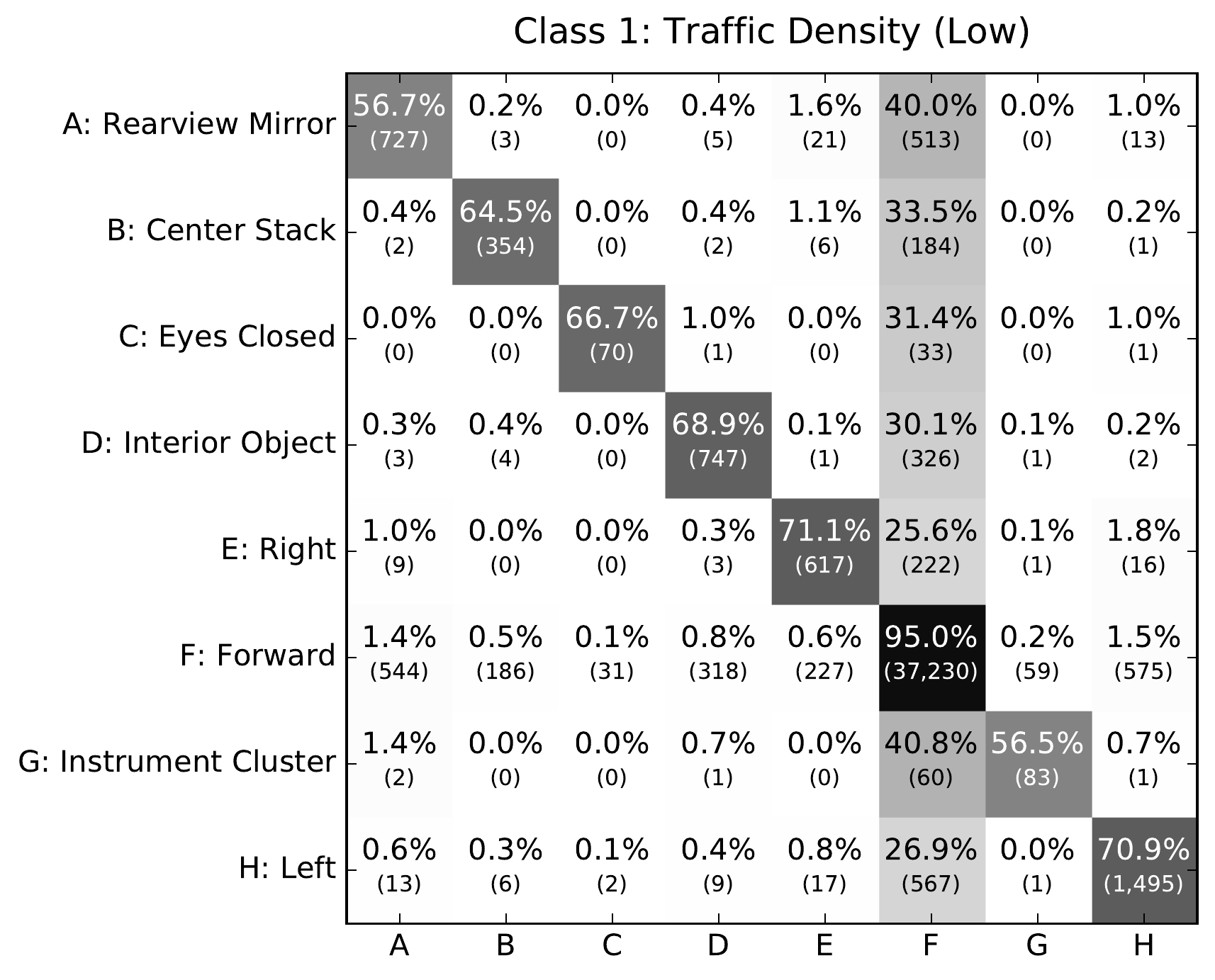}
      \transallfig{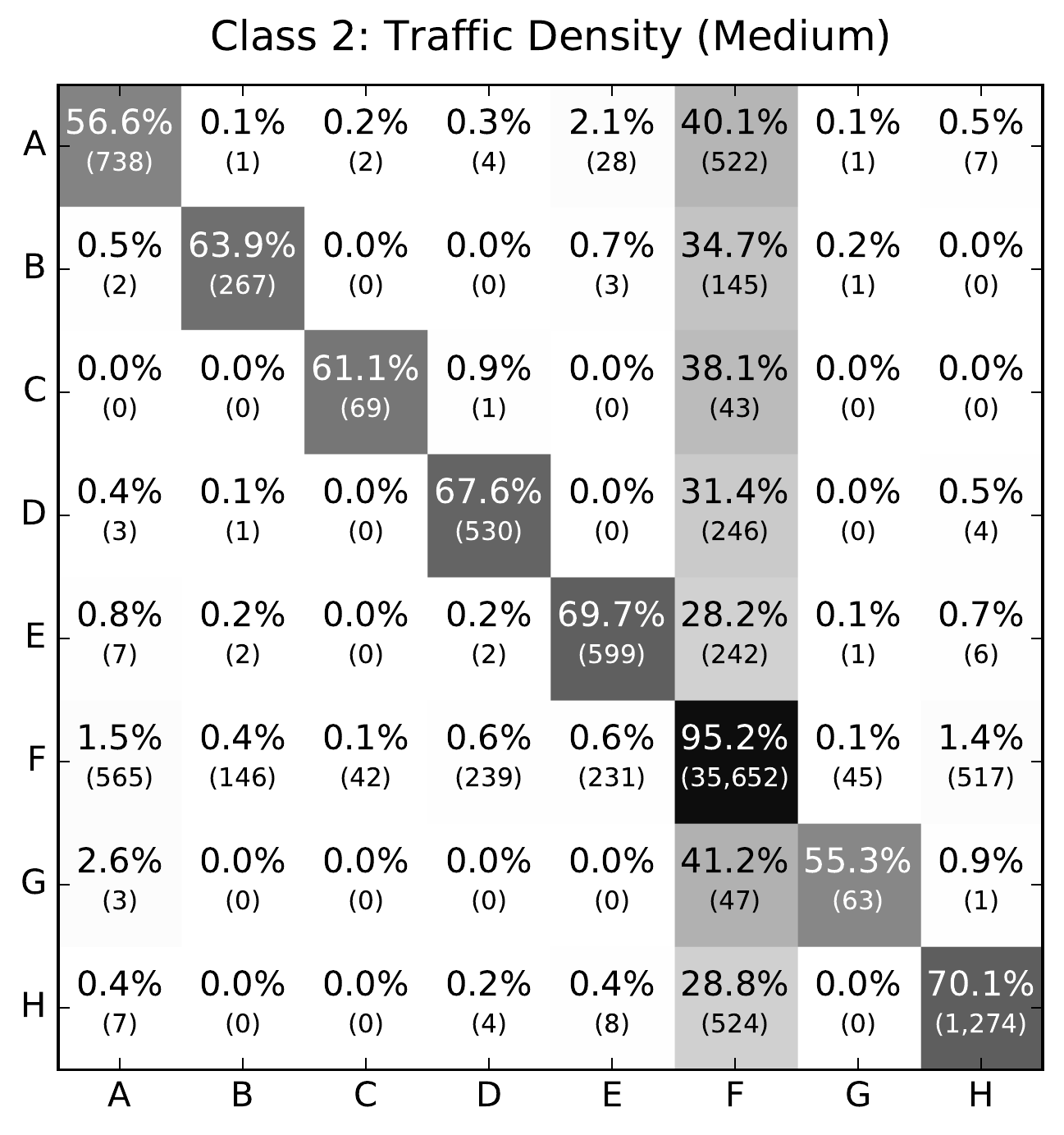}
      \transallfig{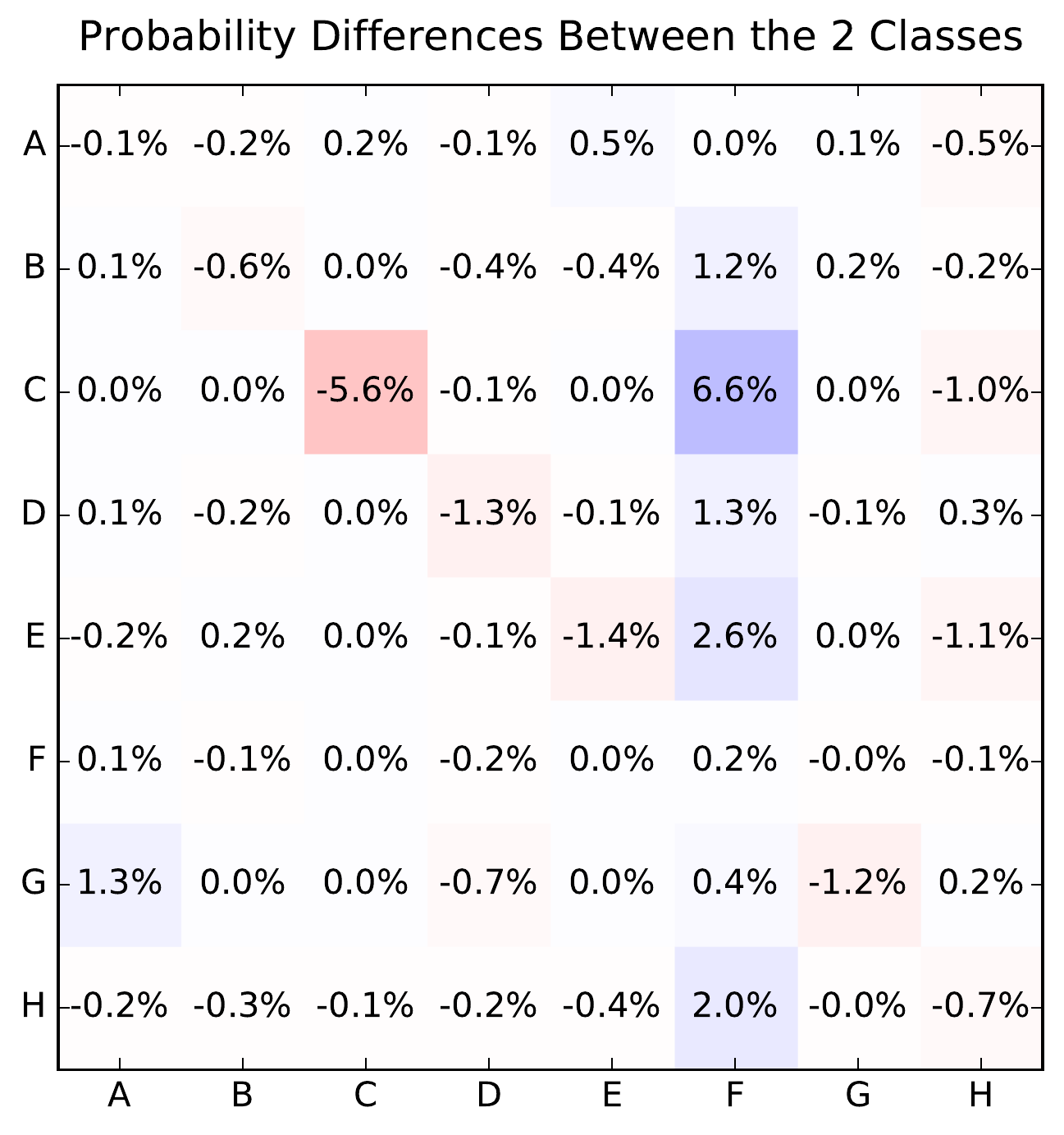}
      \caption{Transition matrices for Traffic Density (Low vs Medium). Classification accuracy: 61.2\%.}
      \label{fig:p14_traffic_density_01}
    \end{figure}
    
    \begin{figure}[!h]
      \transallfig{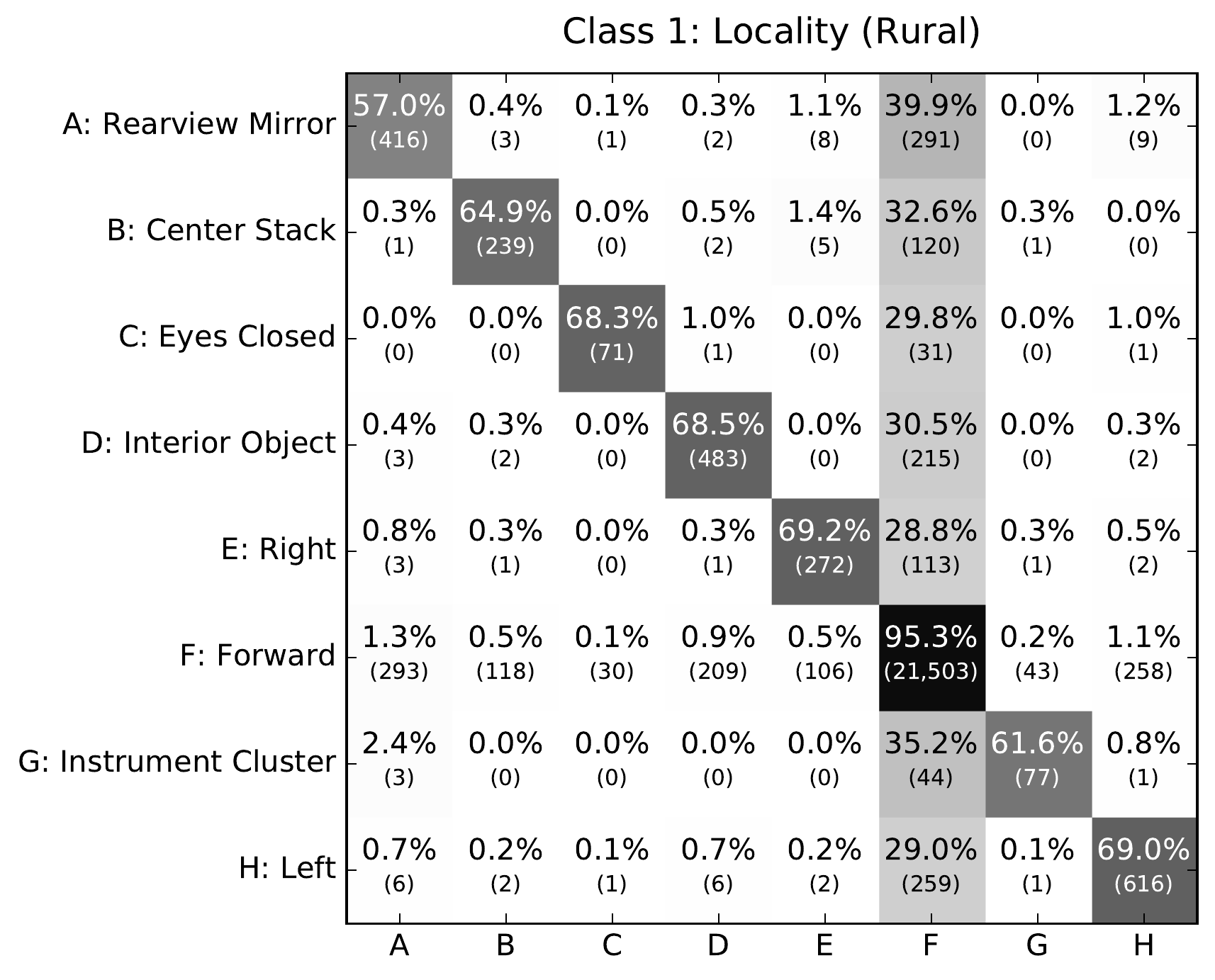}
      \transallfig{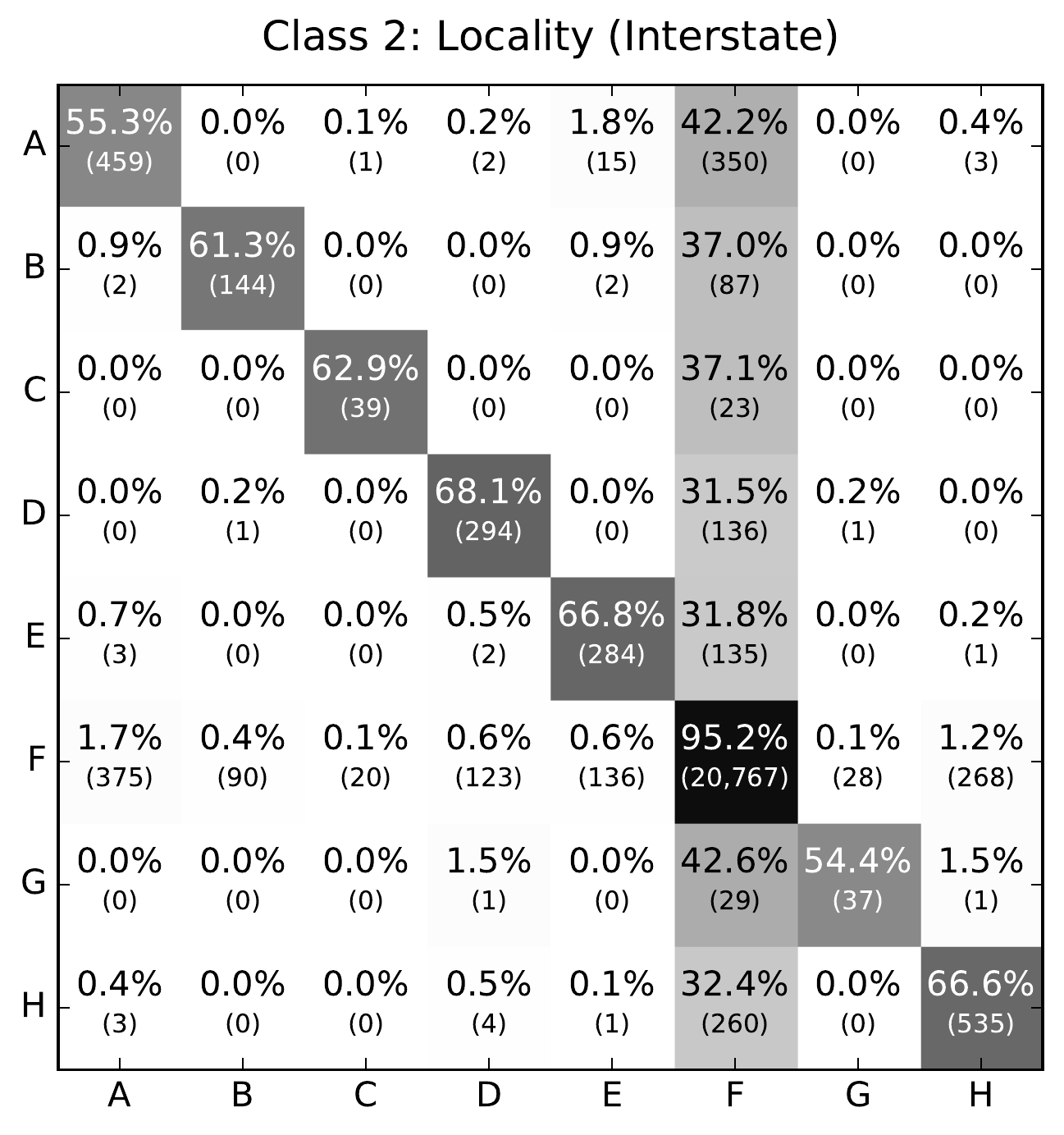}
      \transallfig{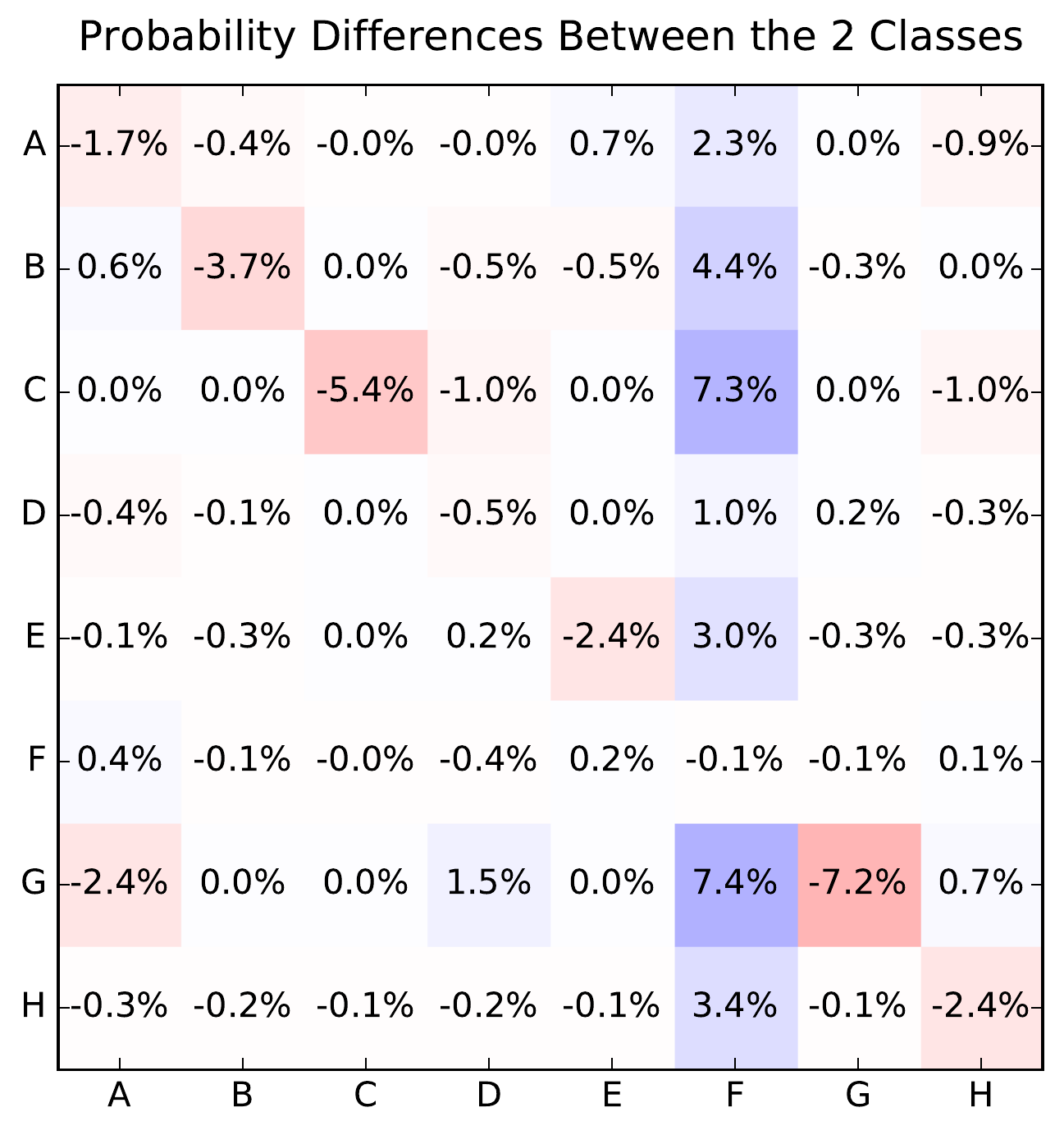}
      \caption{Transition matrices for Locality (Rural vs Interstate). Classification accuracy: 61.9\%.}
      \label{fig:p15_locality_12}
    \end{figure}
    
    \begin{figure}[!h]
      \transallfig{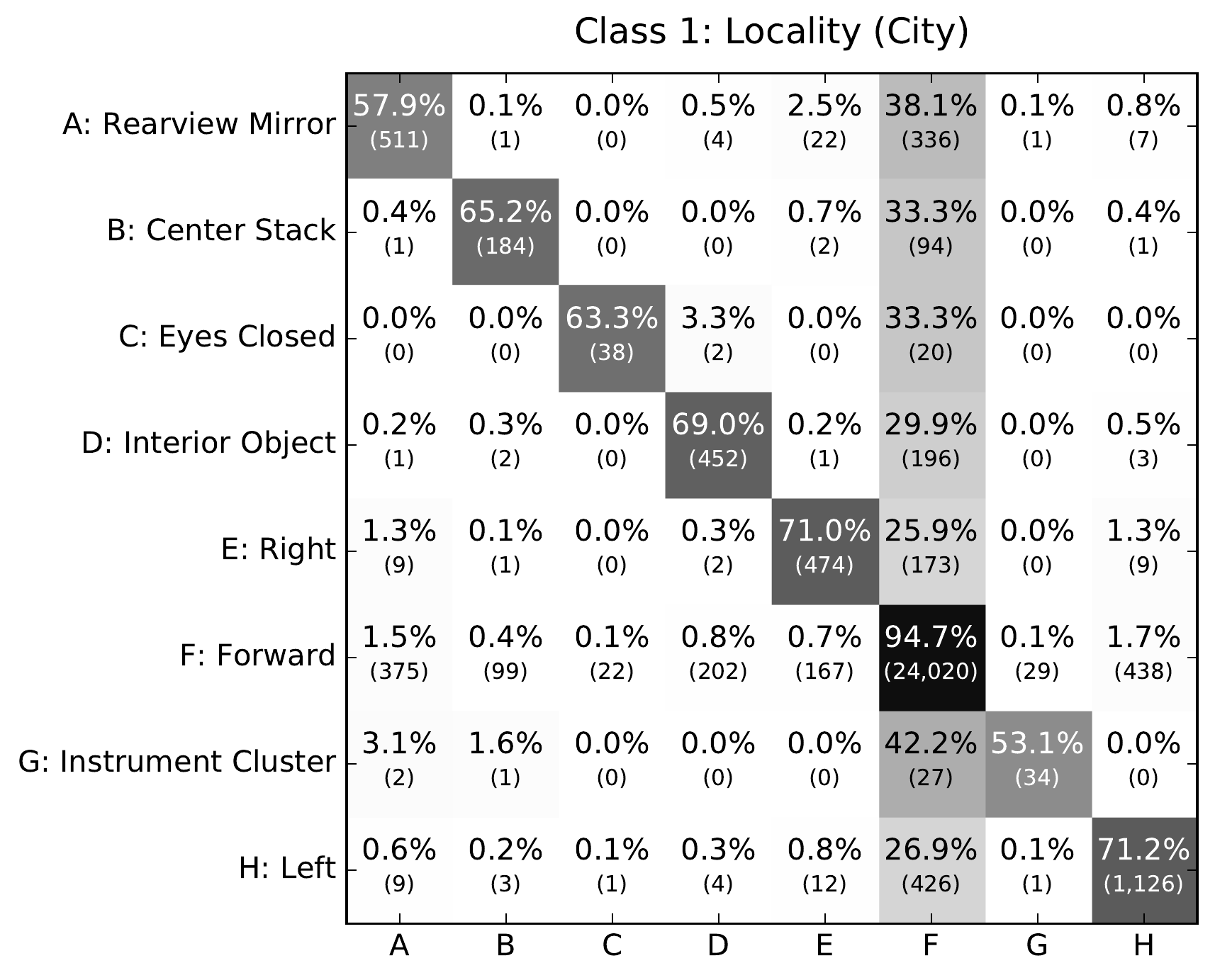}
      \transallfig{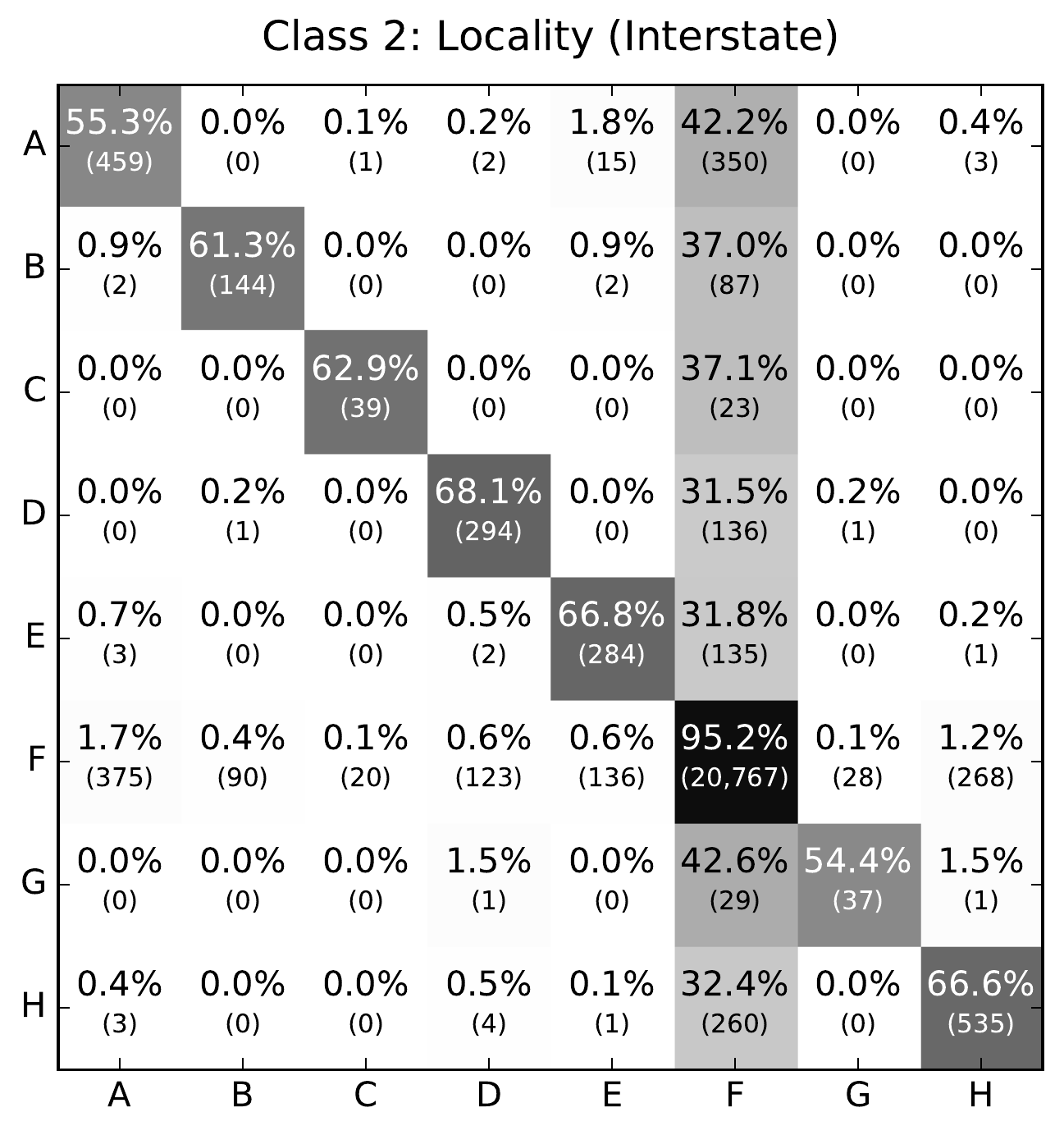}
      \transallfig{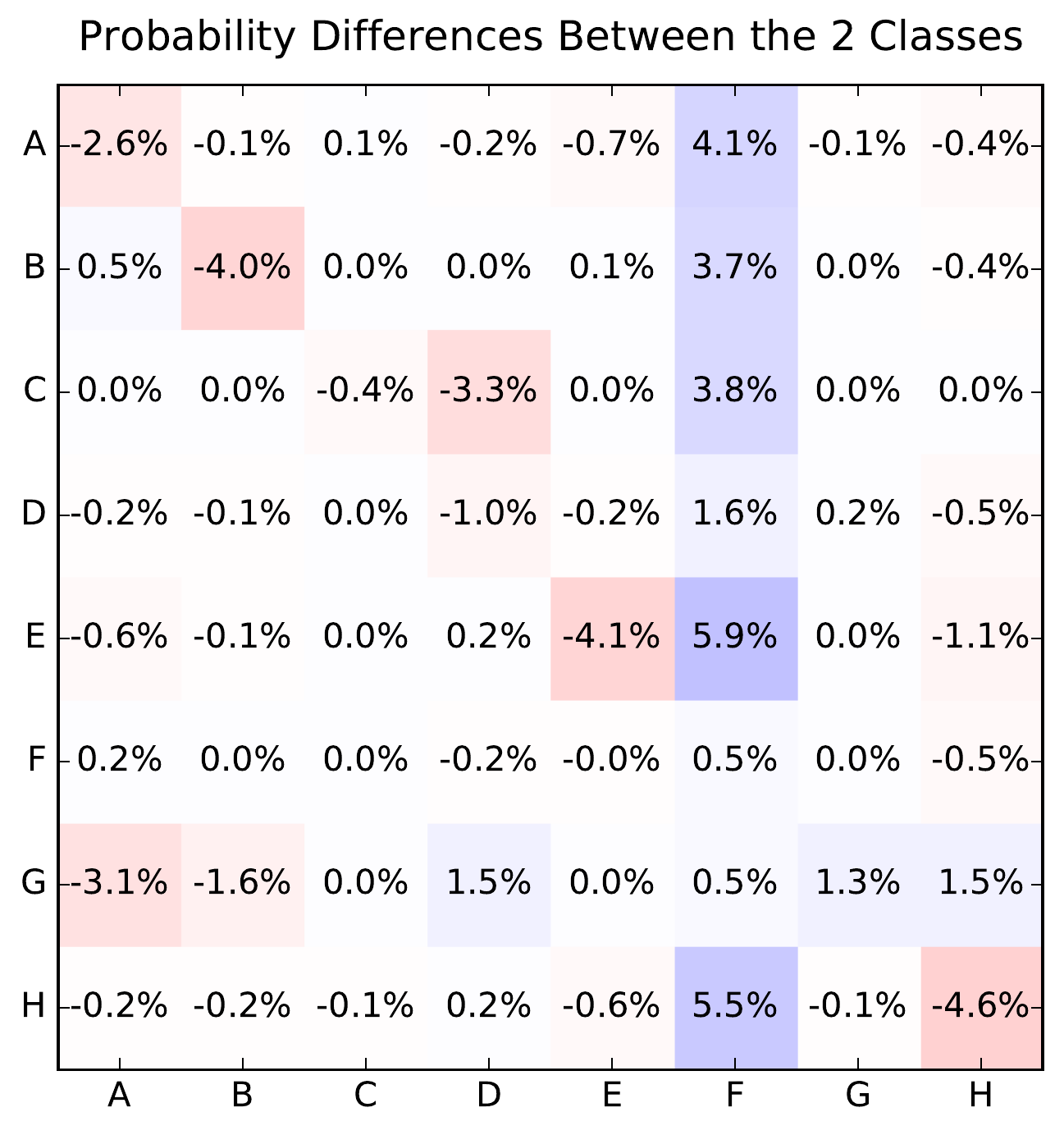}
      \caption{Transition matrices for Locality (City vs Interstate). Classification accuracy: 62.6\%.}
      \label{fig:p16_locality_02}
    \end{figure}
    
    \begin{figure}[!h]
      \transallfig{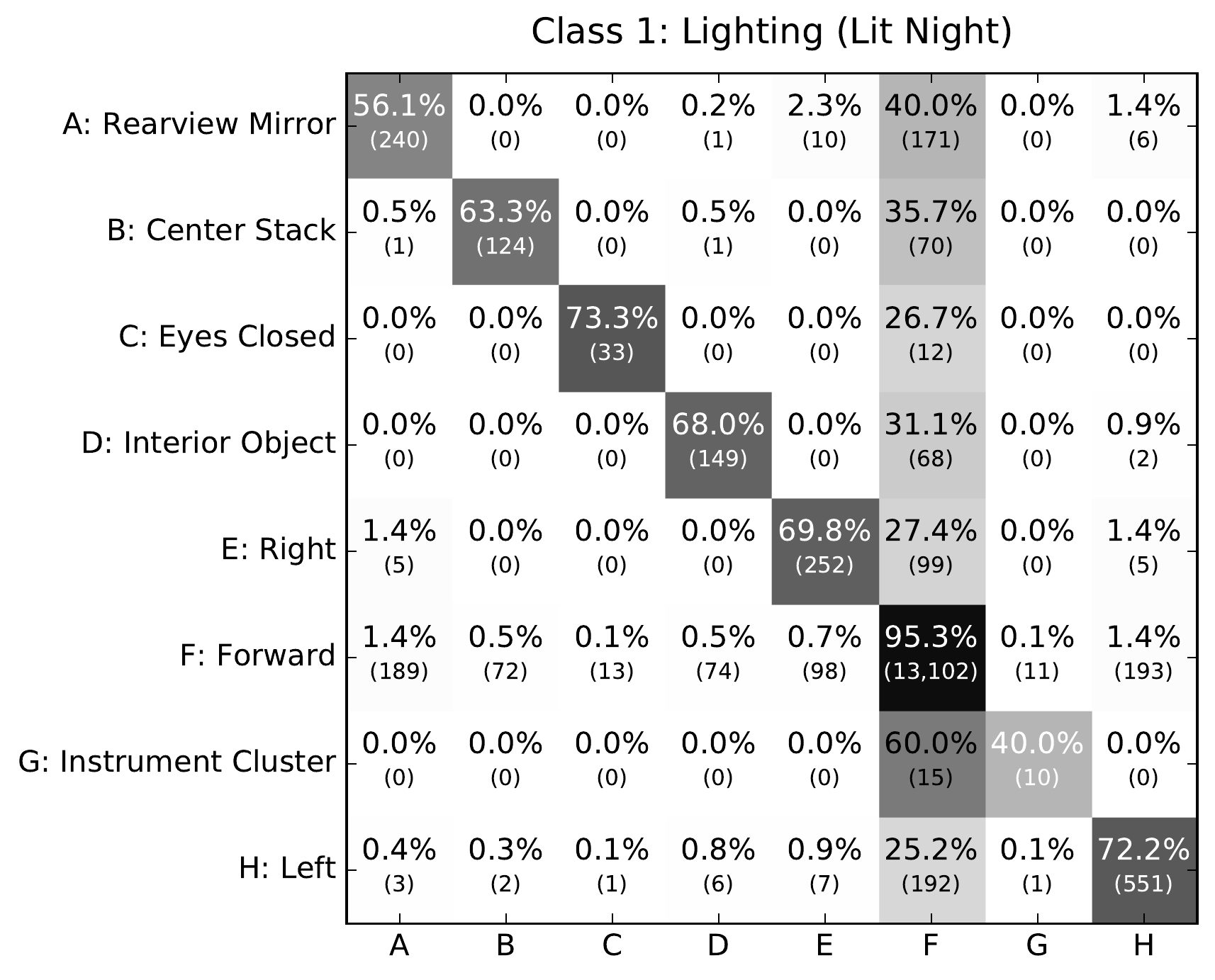}
      \transallfig{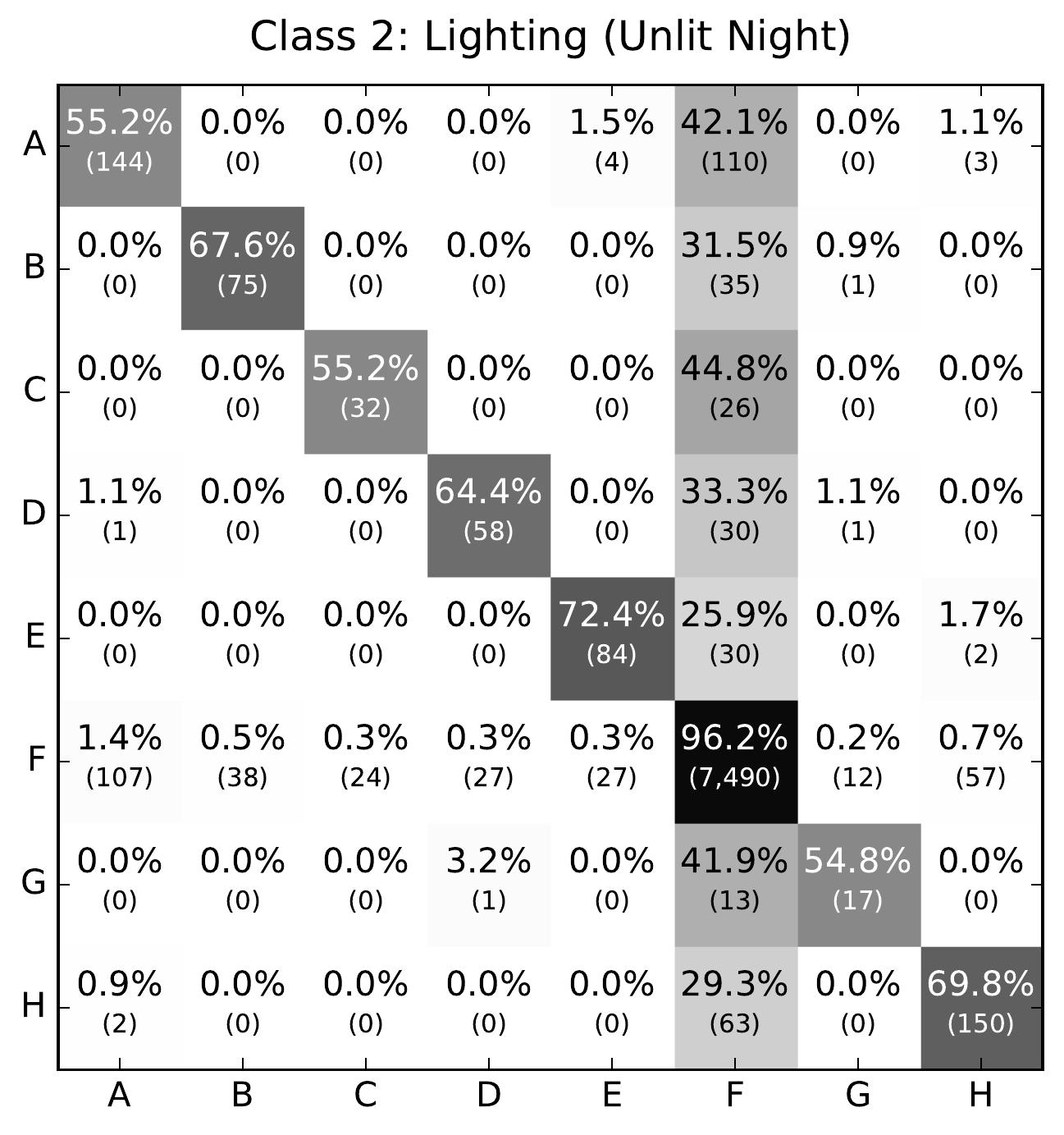}
      \transallfig{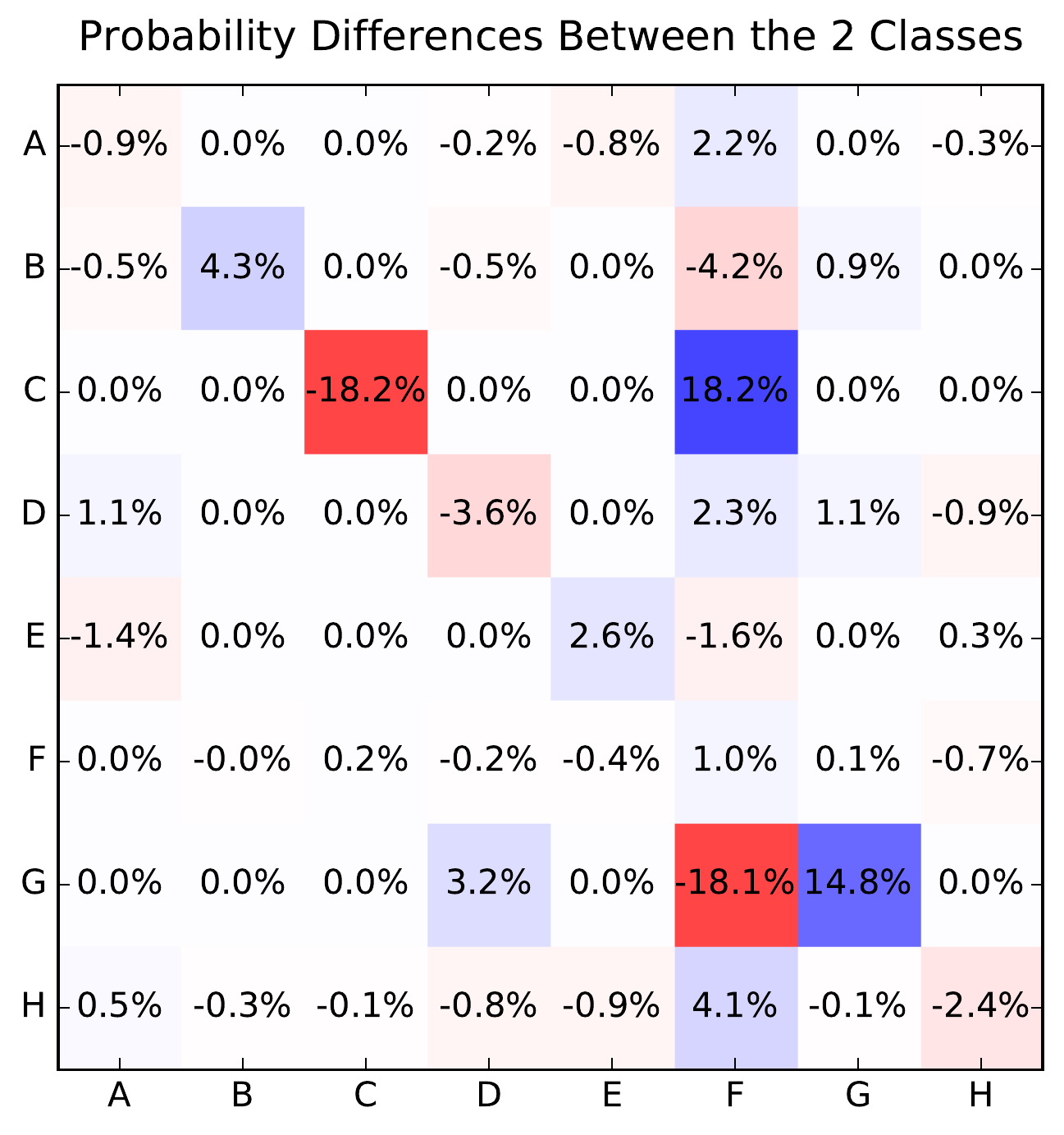}
      \caption{Transition matrices for Lighting (Lit Night vs Unlit Night). Classification accuracy: 63.8\%.}
      \label{fig:p17_lighting_12}
    \end{figure}
    
    \begin{figure}[!h]
      \transallfig{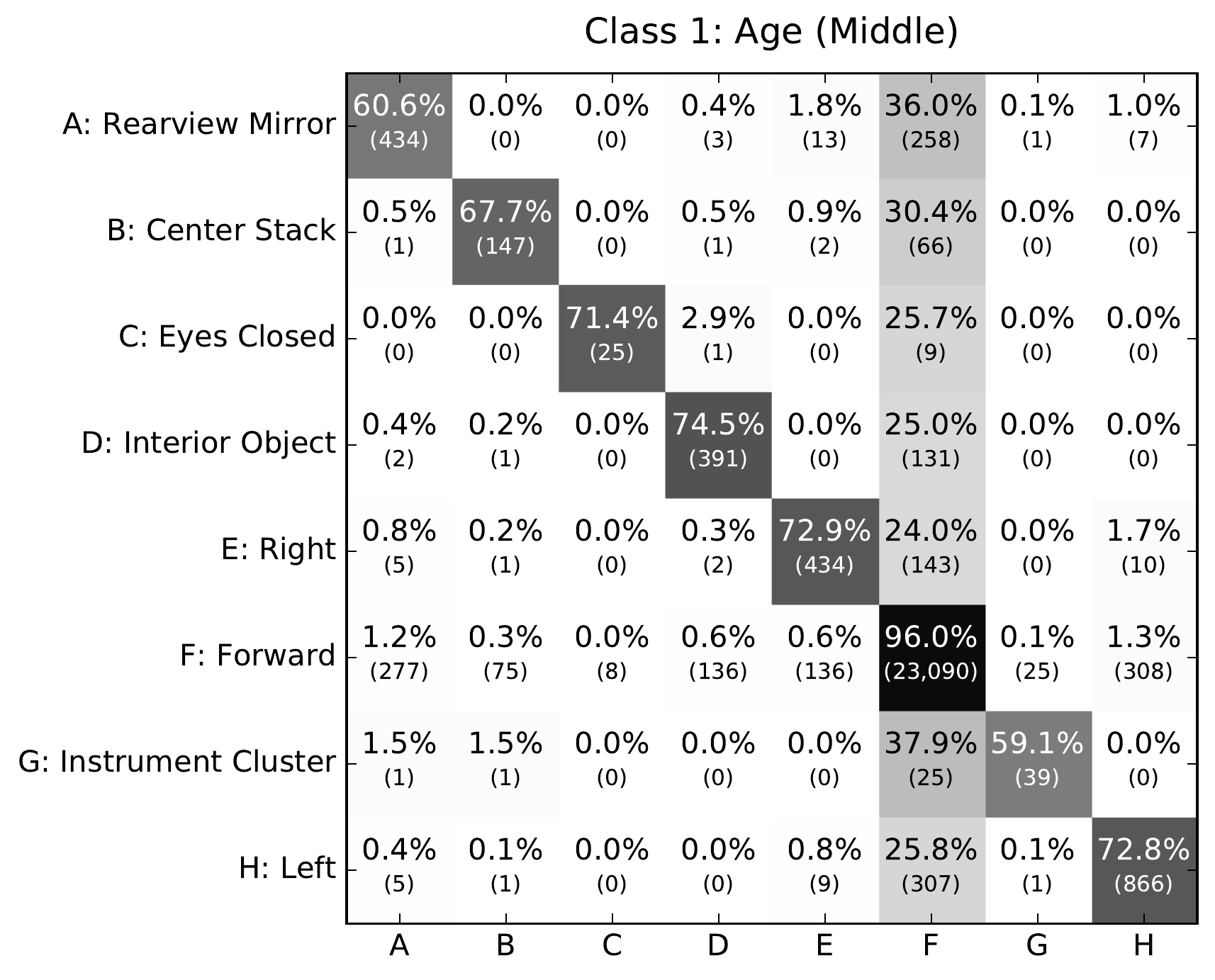}
      \transallfig{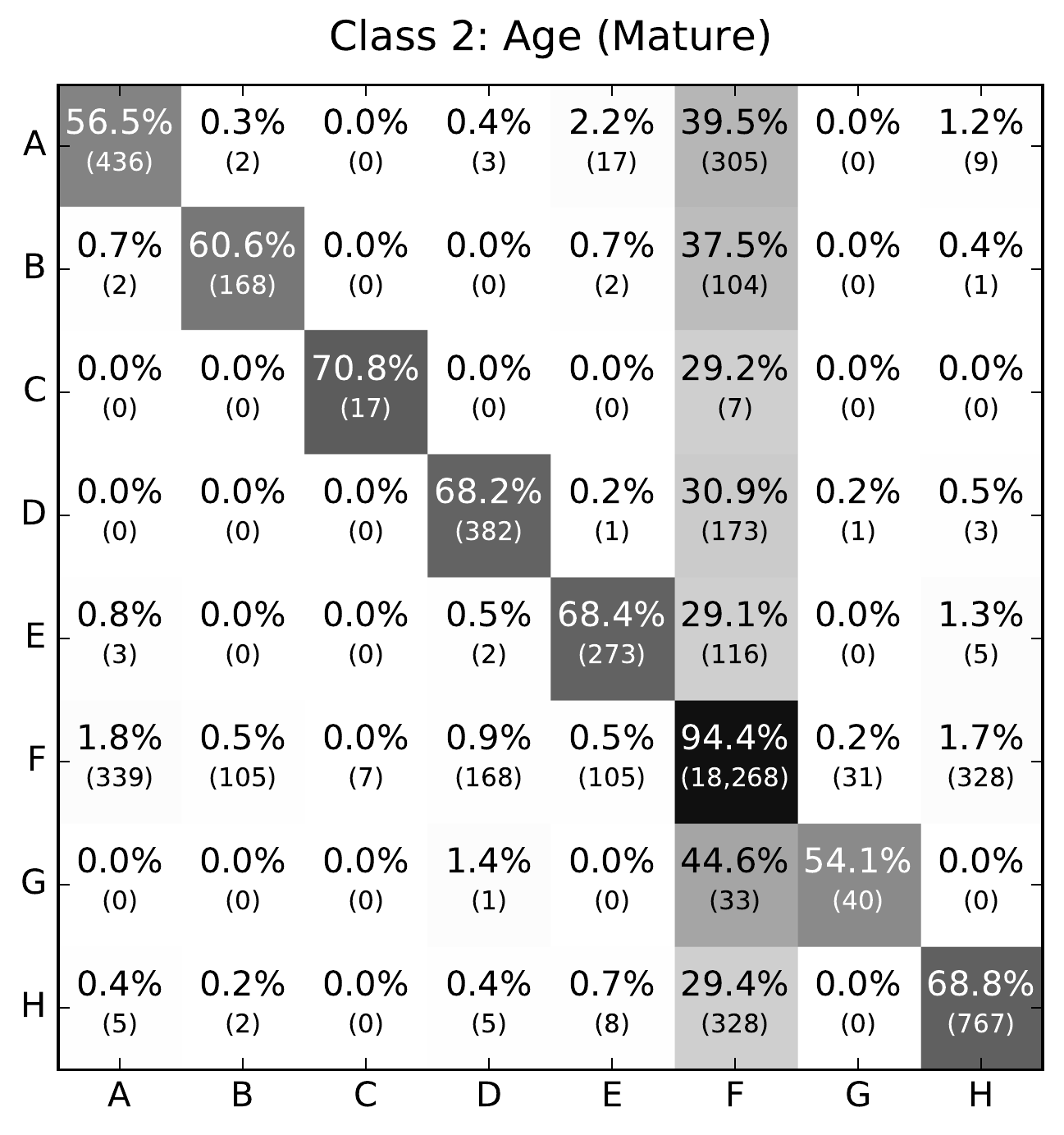}
      \transallfig{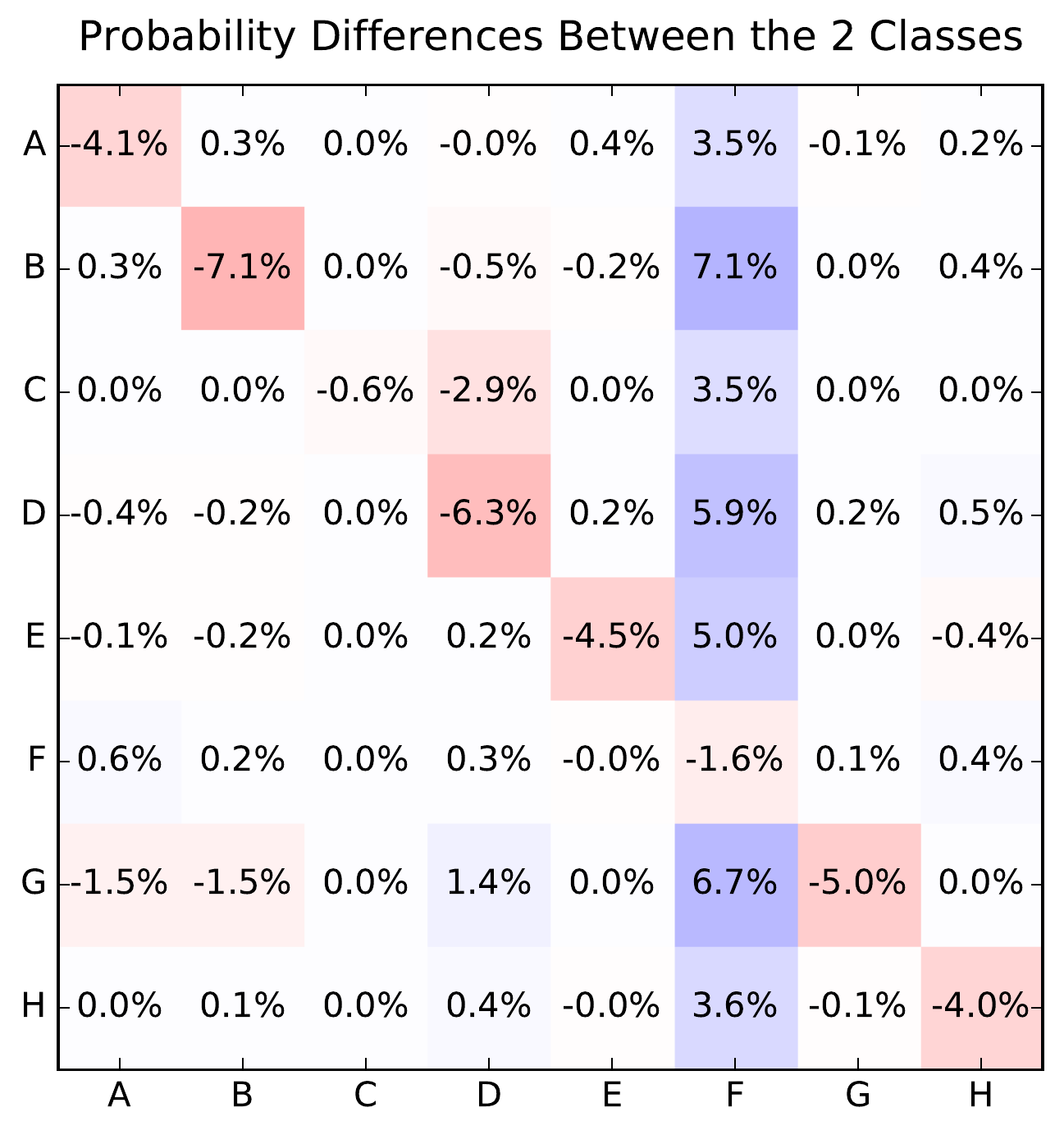}
      \caption{Transition matrices for Age (Middle vs Mature). Classification accuracy: 63.8\%.}
      \label{fig:p18_age_12}
    \end{figure}
    
    \begin{figure}[!h]
      \transallfig{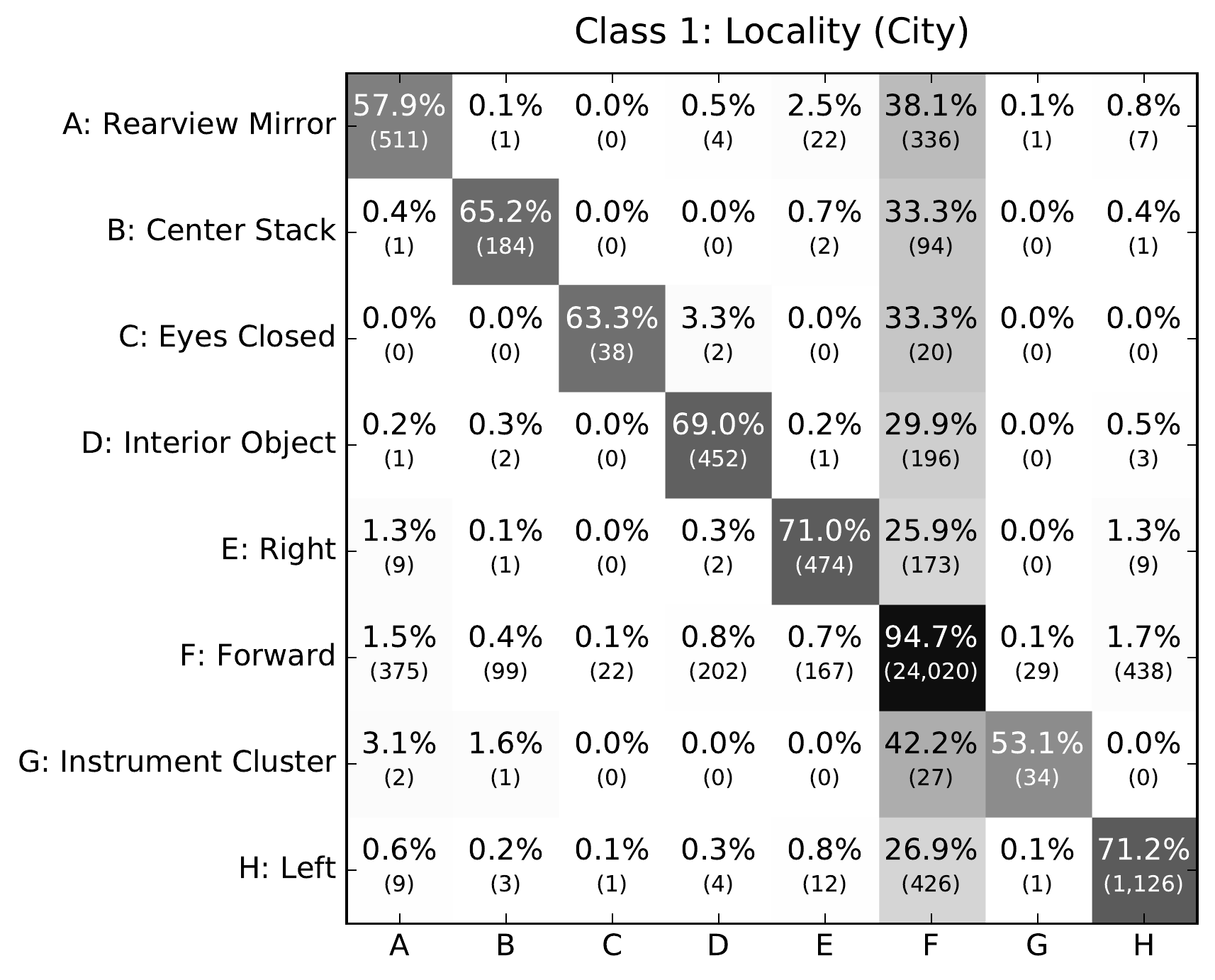}
      \transallfig{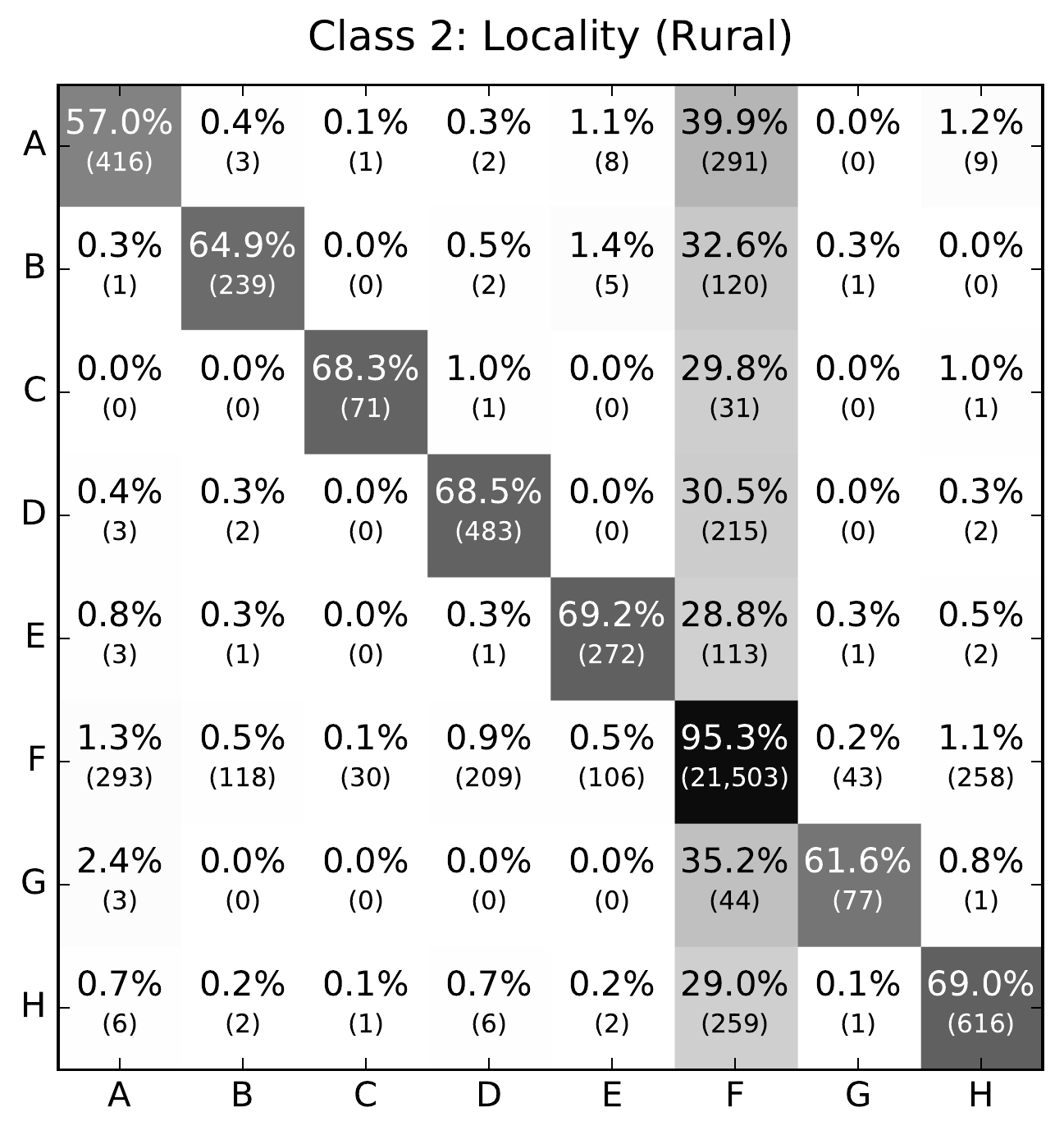}
      \transallfig{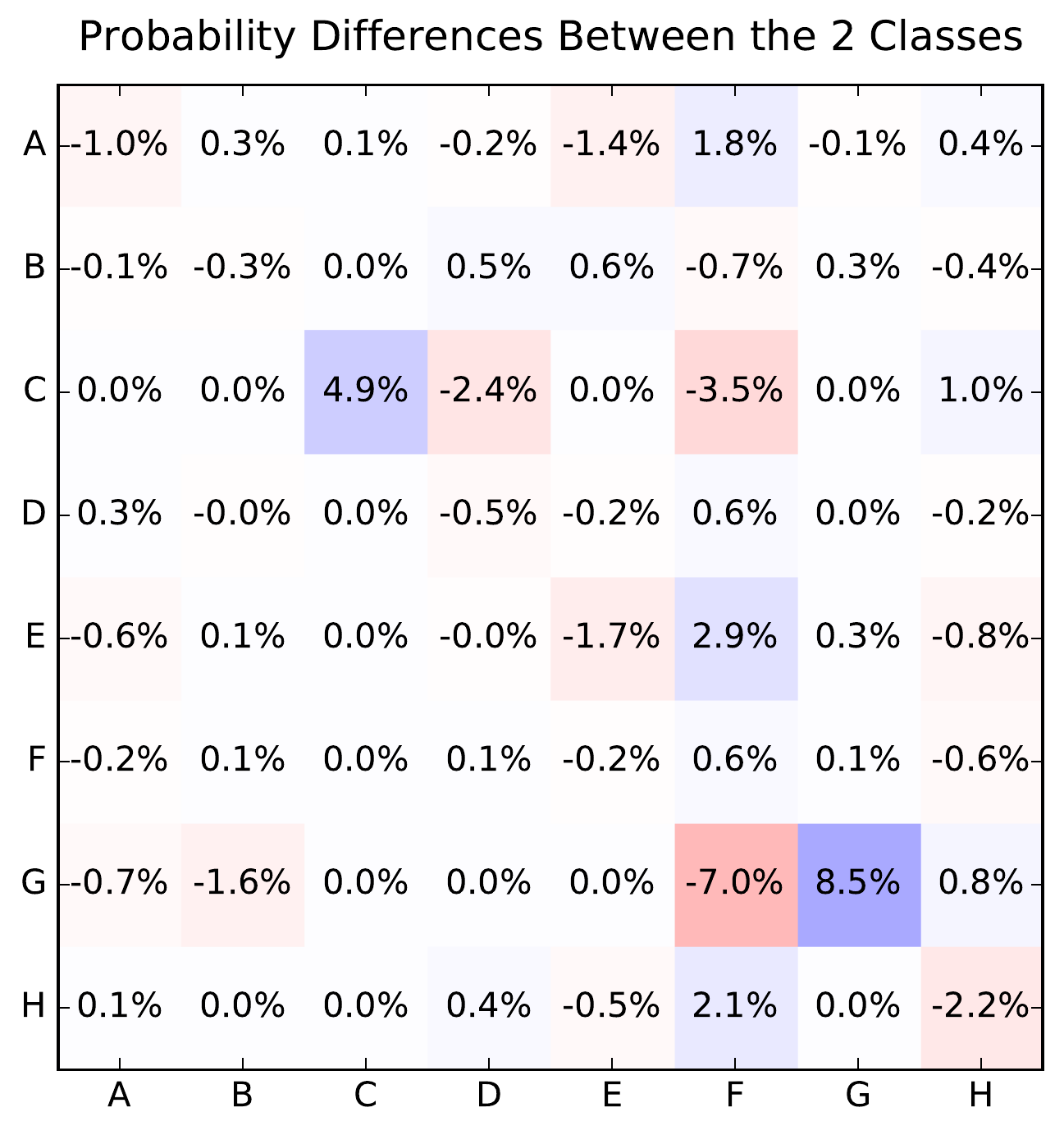}
      \caption{Transition matrices for Locality (City vs Rural). Classification accuracy: 63.8\%.}
      \label{fig:p19_locality_01}
    \end{figure}
    
    \begin{figure}[!h]
      \transallfig{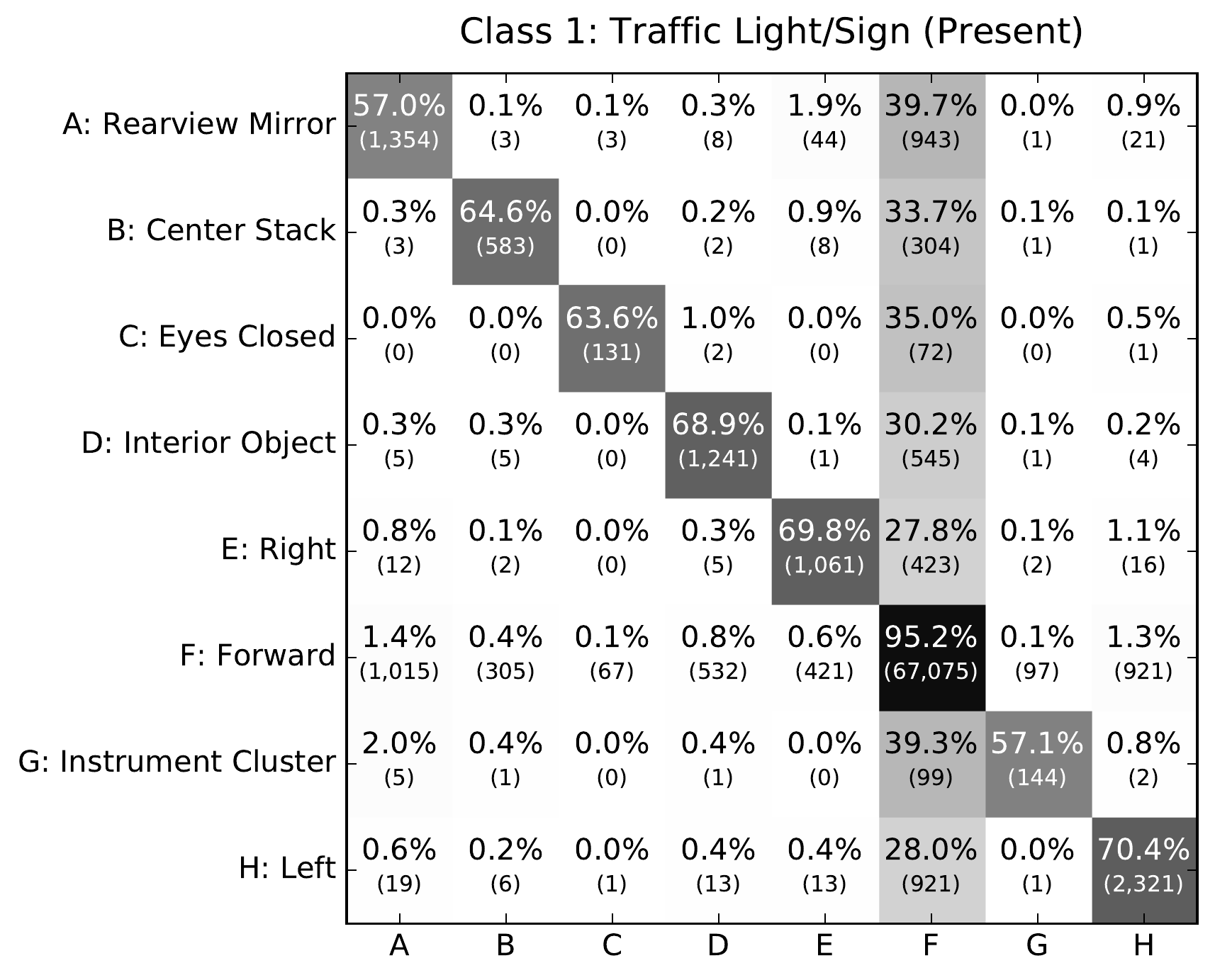}
      \transallfig{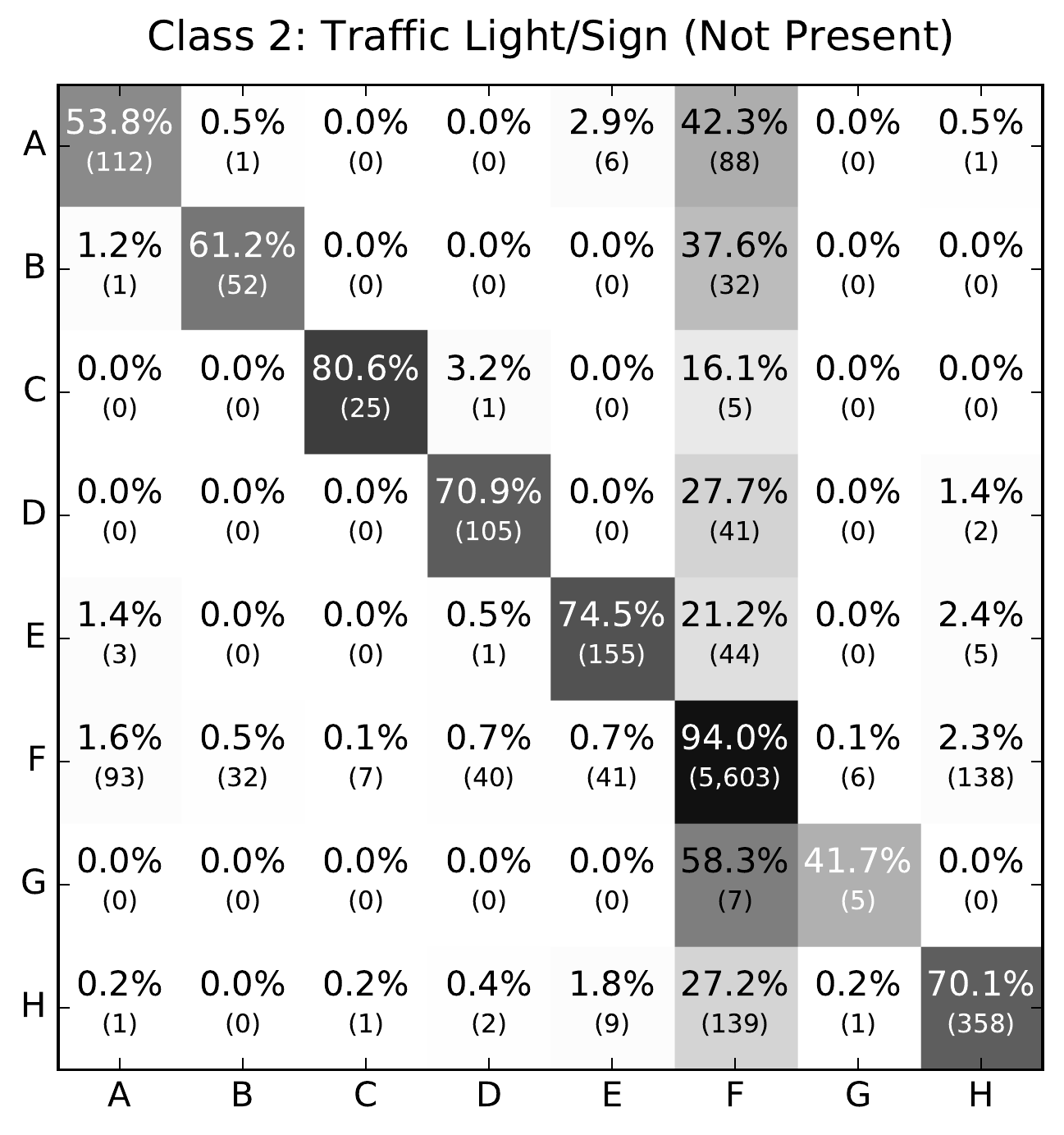}
      \transallfig{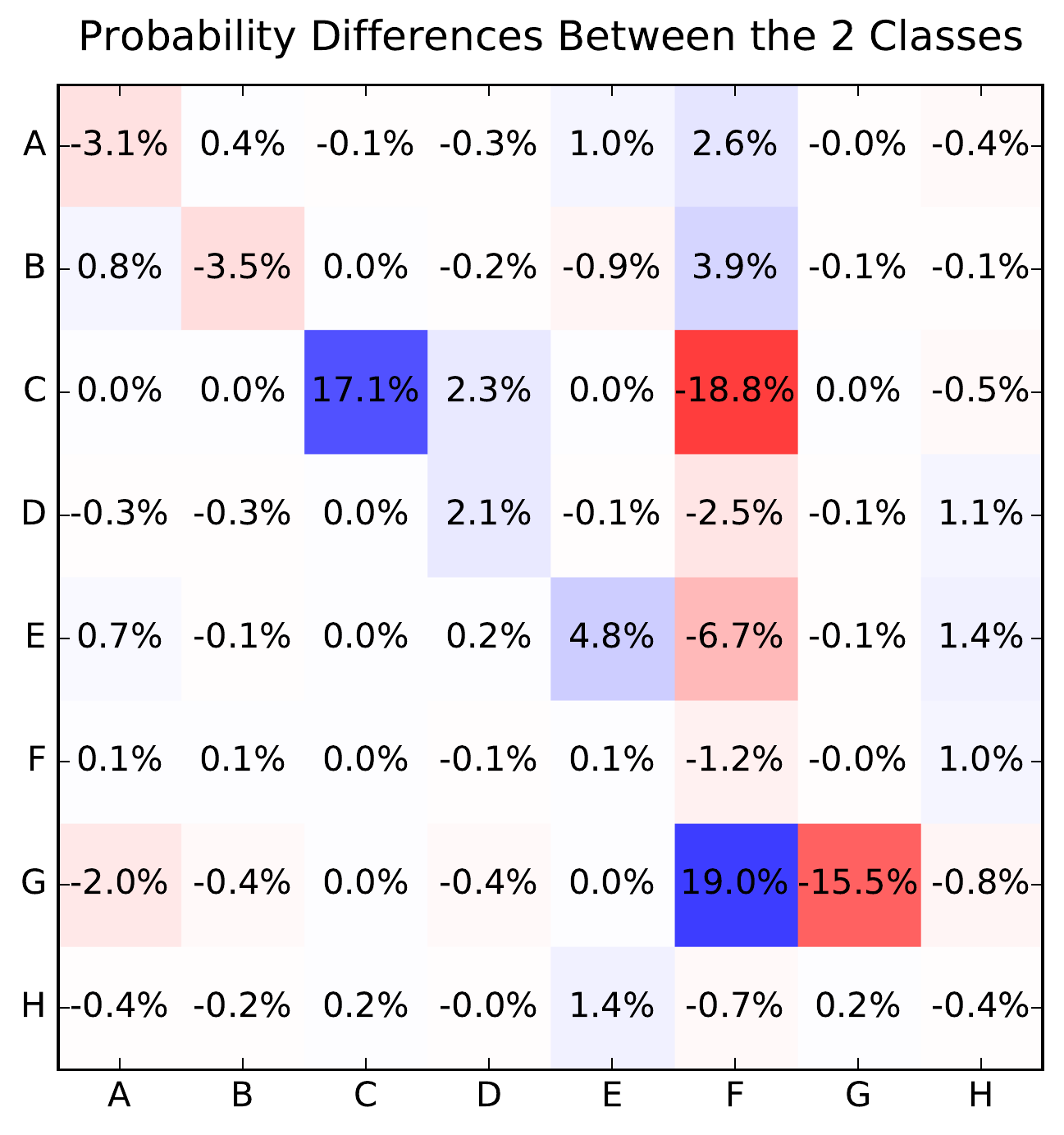}
      \caption{Transition matrices for Traffic Light/Sign (Present vs Not Present). Classification accuracy: 64.0\%.}
      \label{fig:p20_traffic_control_01}
    \end{figure}
    
    \begin{figure}[!h]
      \transallfig{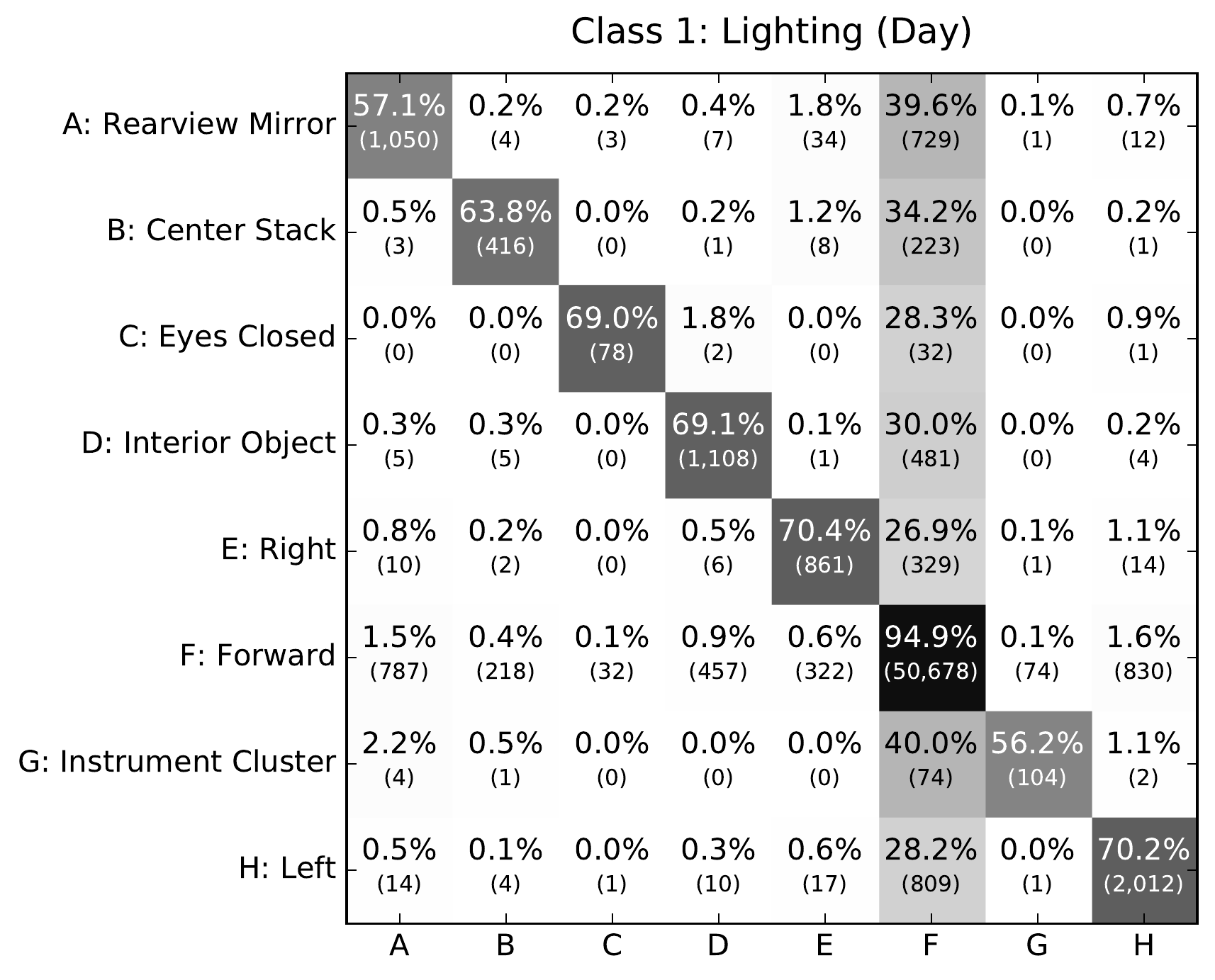}
      \transallfig{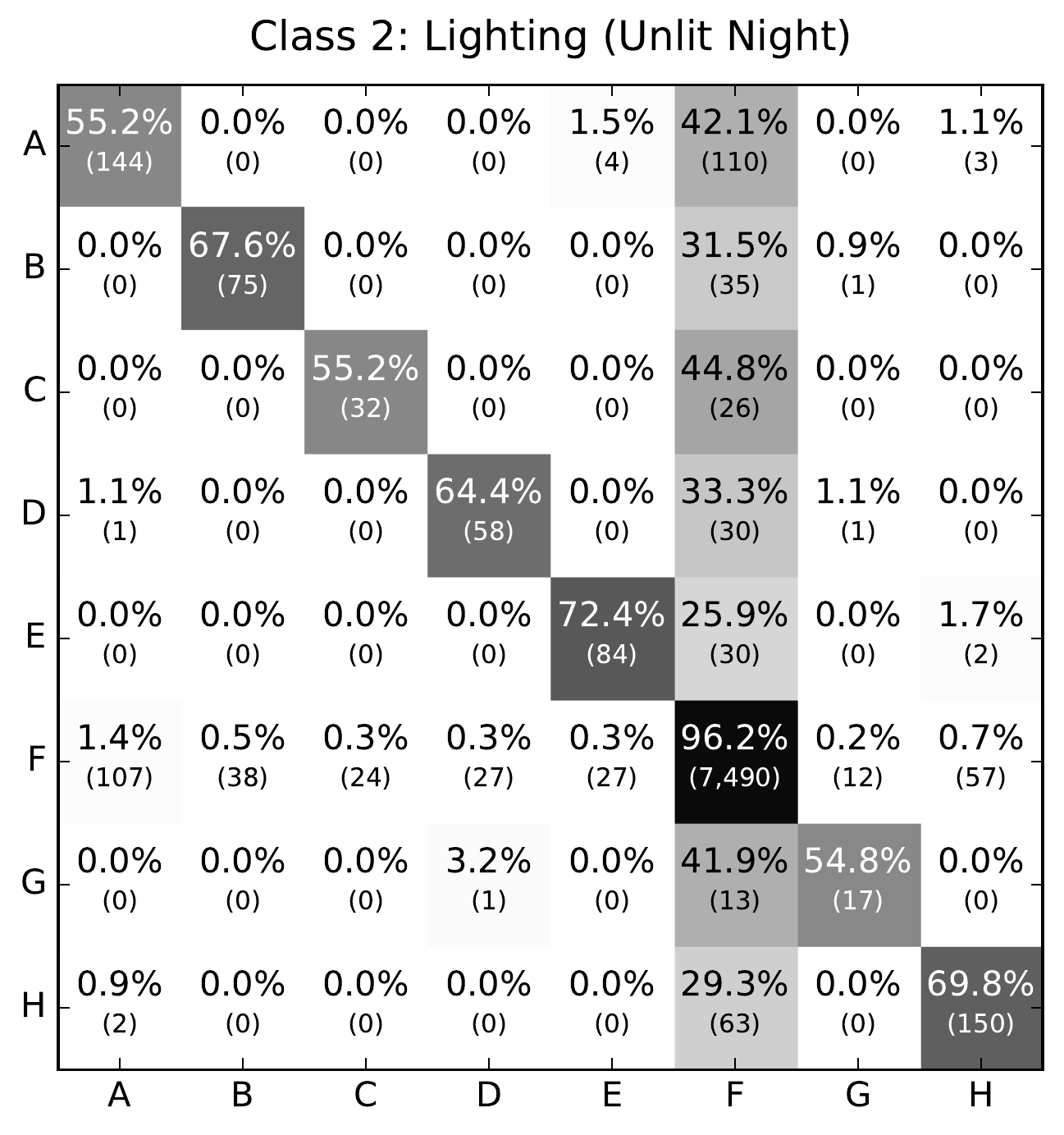}
      \transallfig{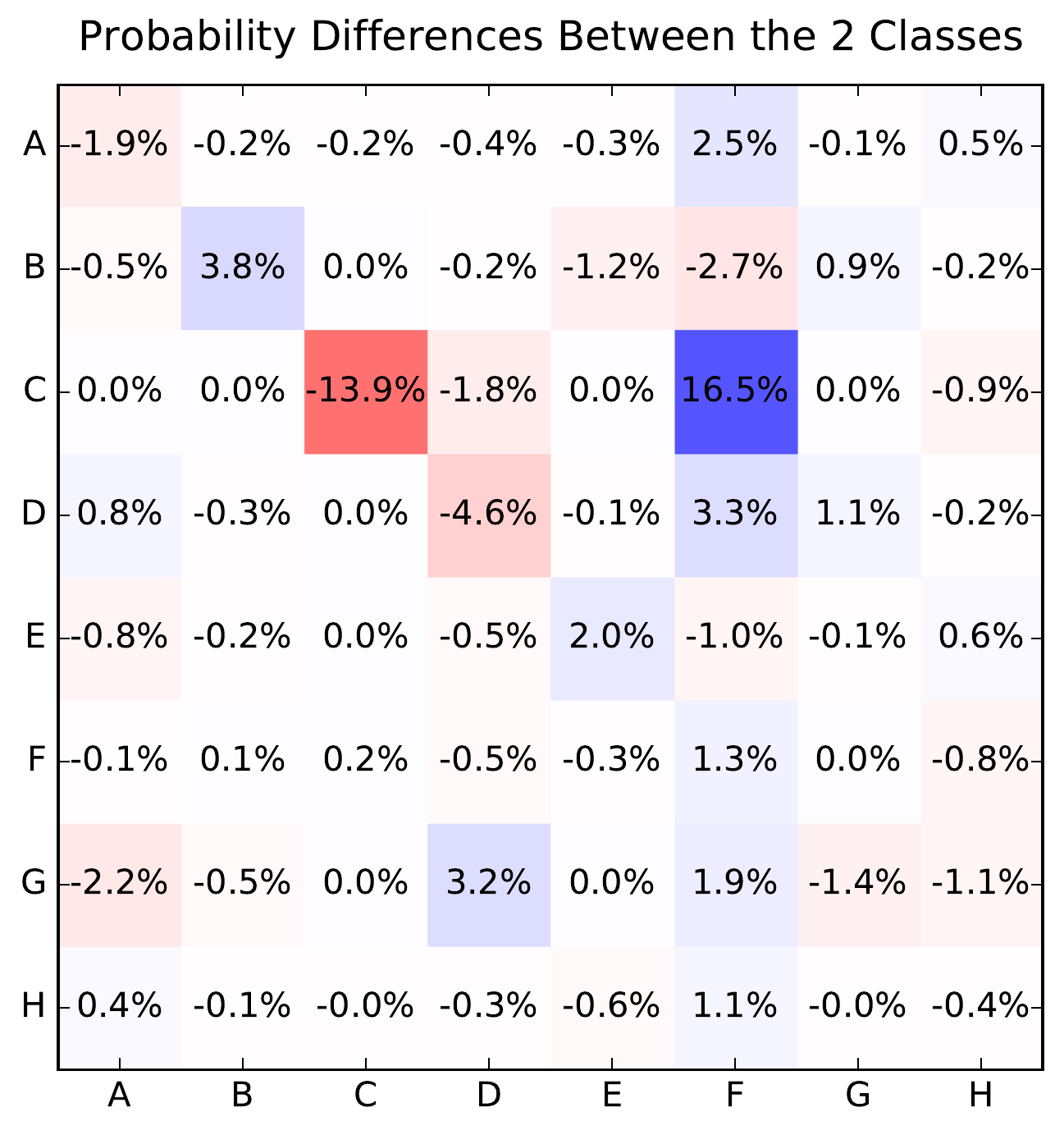}
      \caption{Transition matrices for Lighting (Day vs Unlit Night). Classification accuracy: 66.6\%.}
      \label{fig:p21_lighting_02}
    \end{figure}
    
    \begin{figure}[!h]
      \transallfig{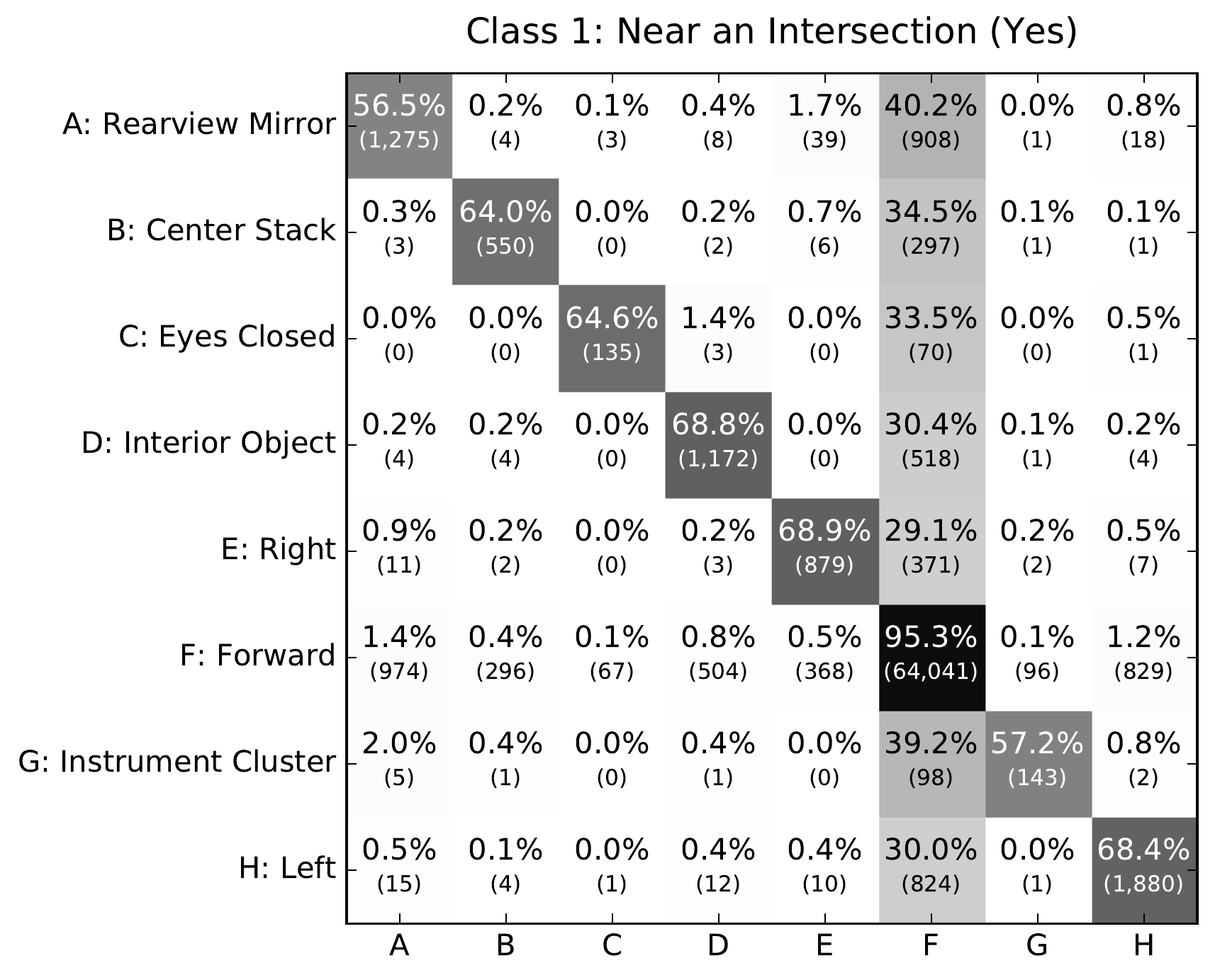}
      \transallfig{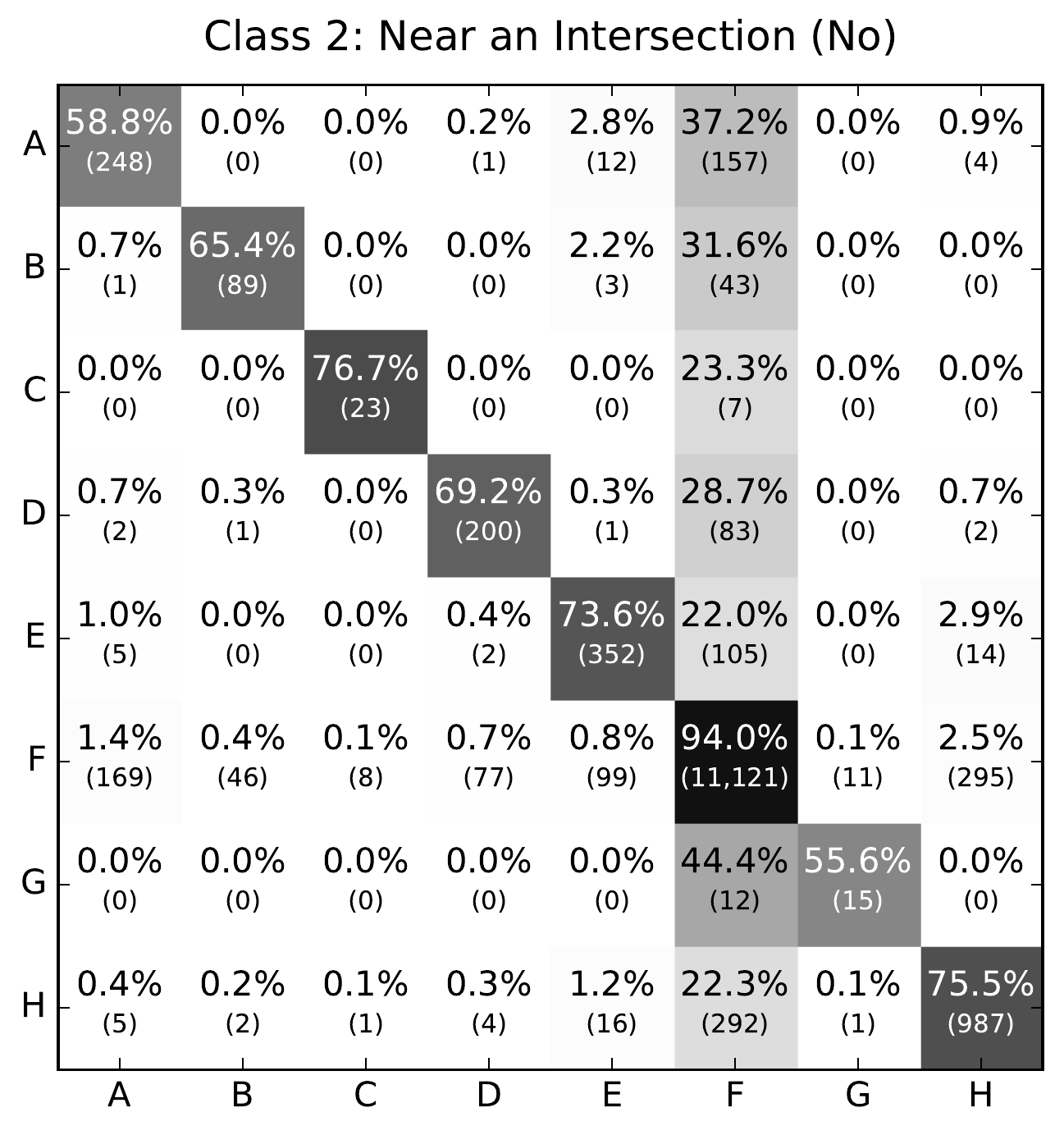}
      \transallfig{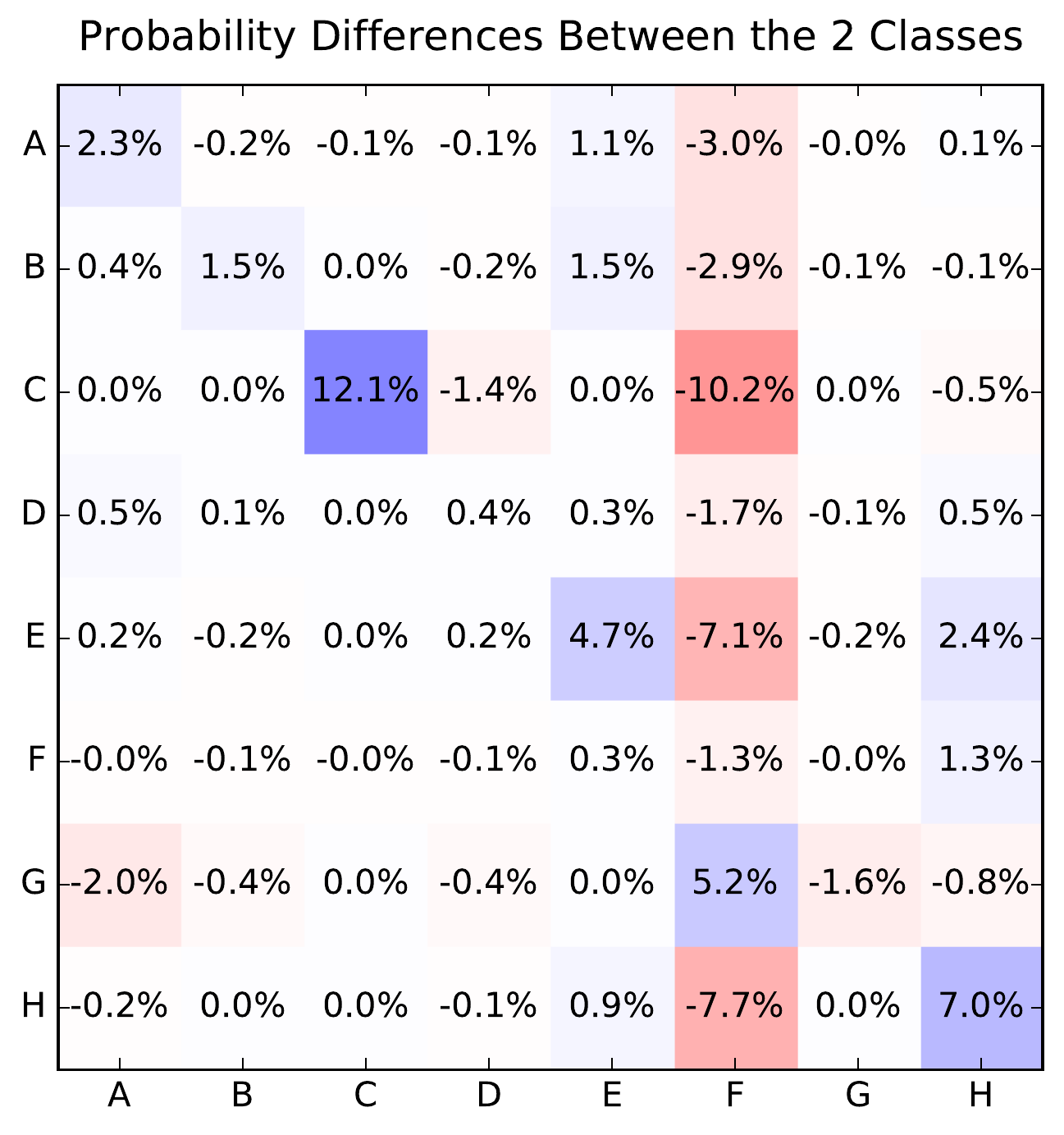}
      \caption{Transition matrices for Near an Intersection (Yes vs No). Classification accuracy: 70.9\%.}
      \label{fig:p22_relation_to_junction_01}
    \end{figure}
    
    \begin{figure}[!h]
      \transallfig{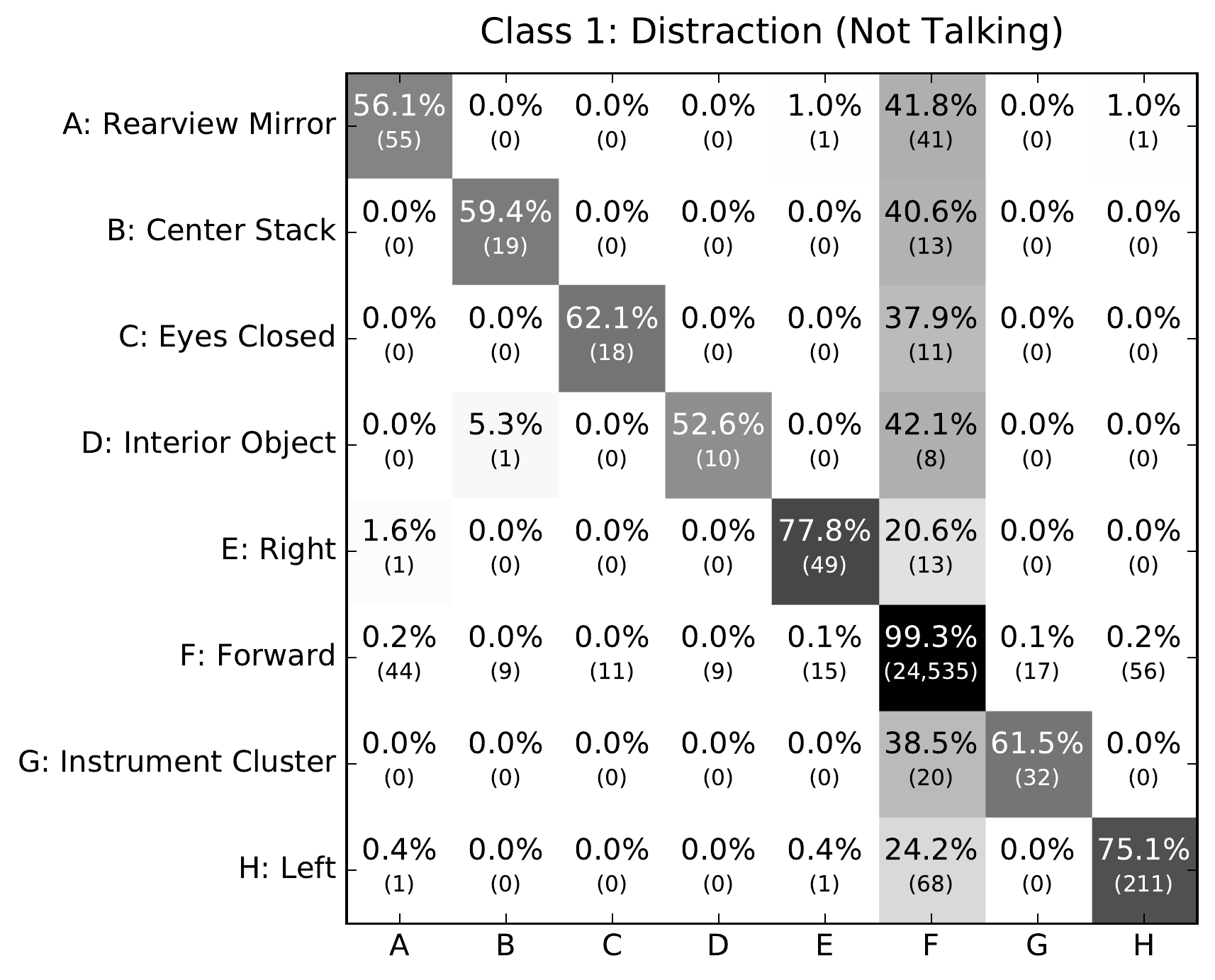}
      \transallfig{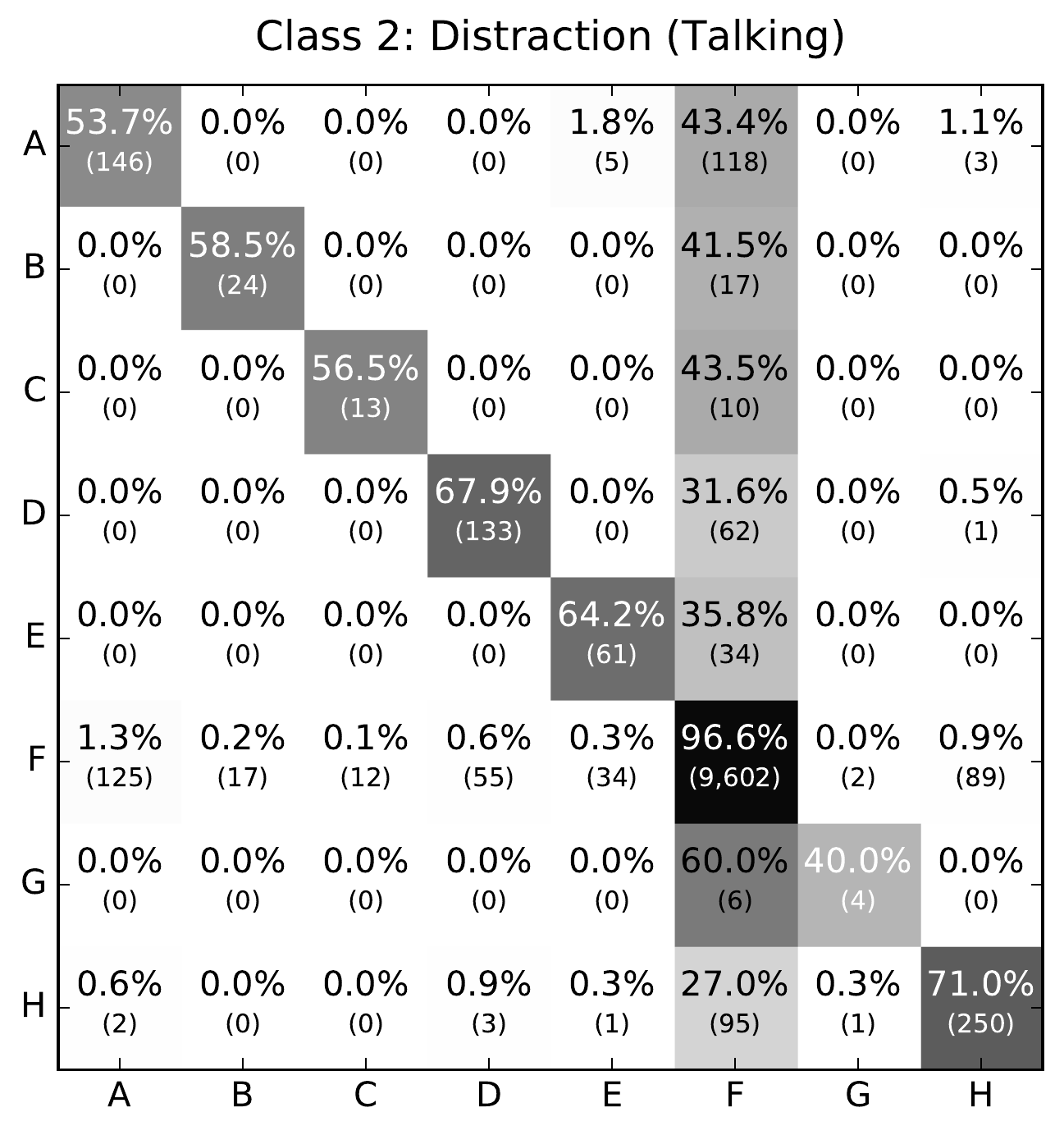}
      \transallfig{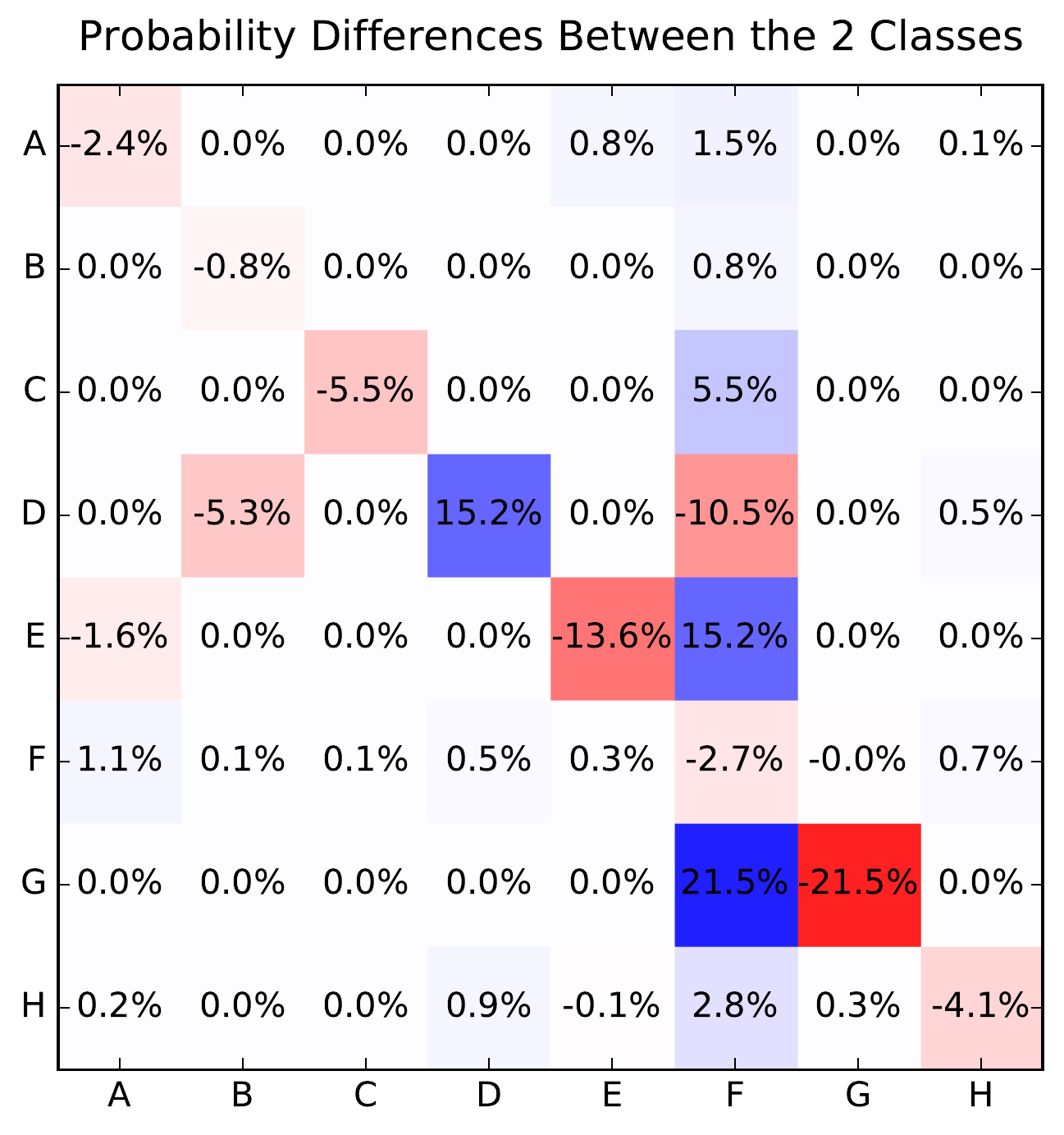}
      \caption{Transition matrices for Distraction (Talking). Classification accuracy: 75.4\%.}
      \label{fig:p23_distraction_01}
    \end{figure}
    
    \begin{figure}[!h]
      \transallfig{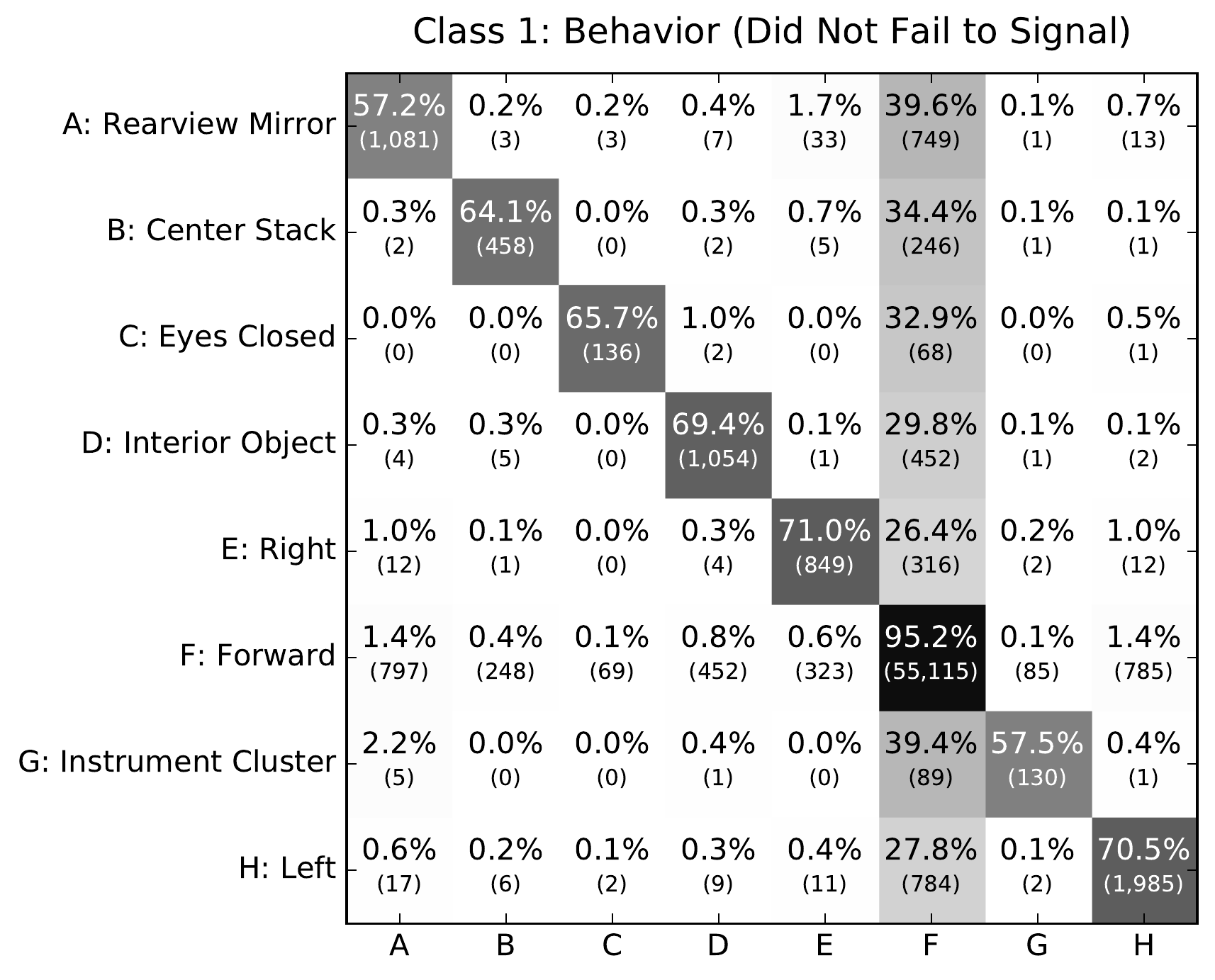}
      \transallfig{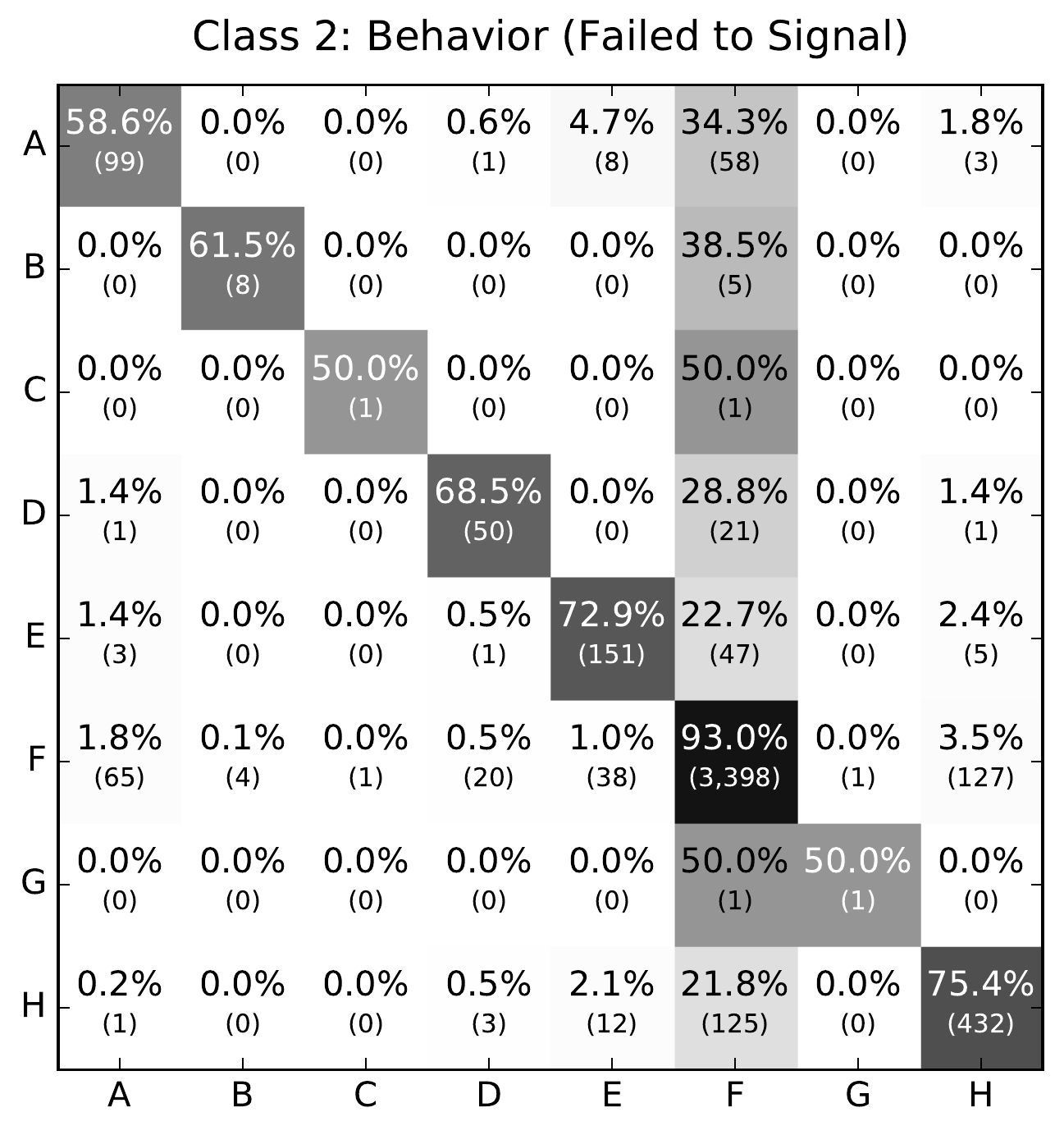}
      \transallfig{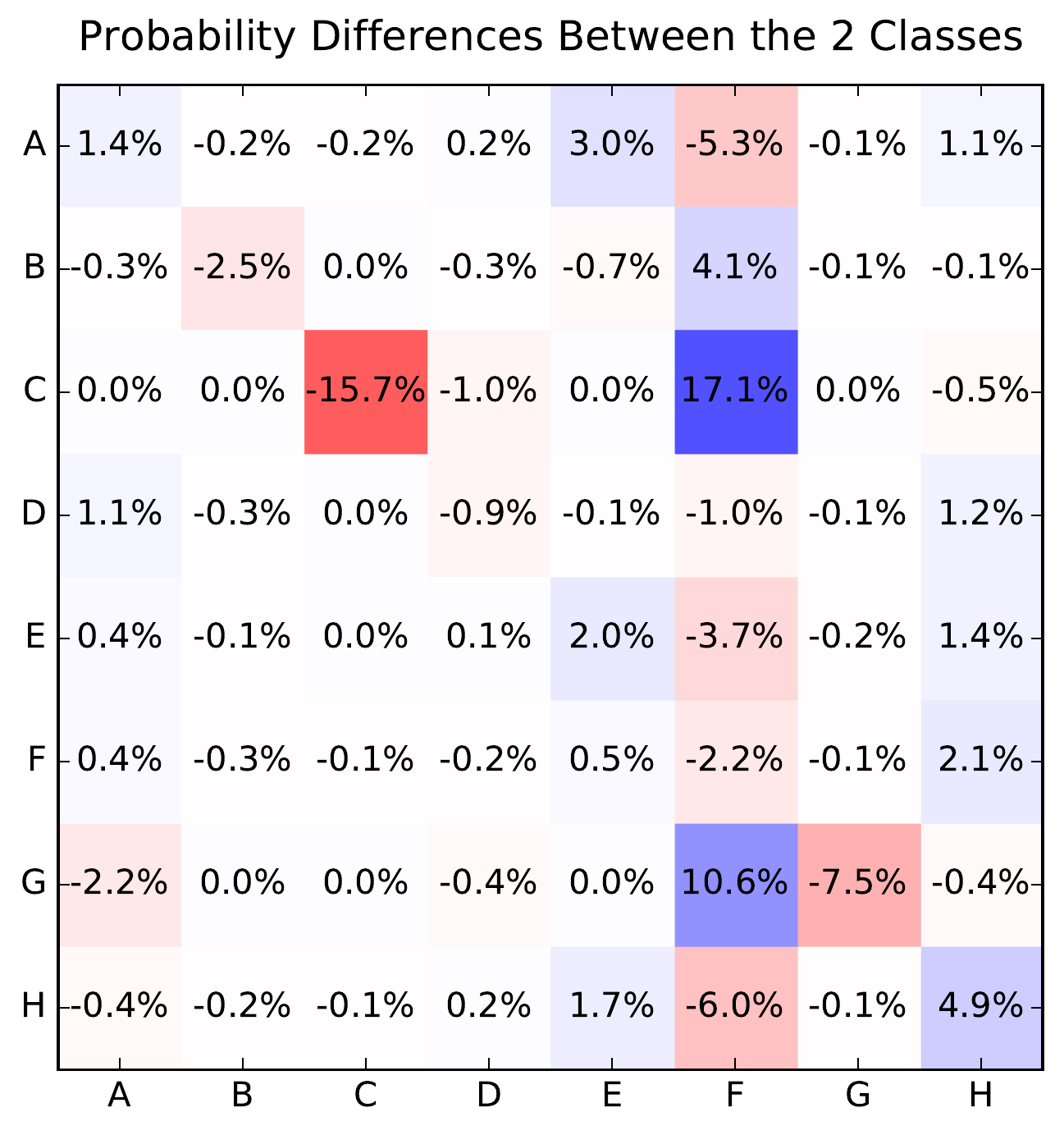}
      \caption{Transition matrices for Behavior (Failed to Signal). Classification accuracy: 75.5\%.}
      \label{fig:p24_behavior_02}
    \end{figure}
    
    \begin{figure}[!h]
      \transallfig{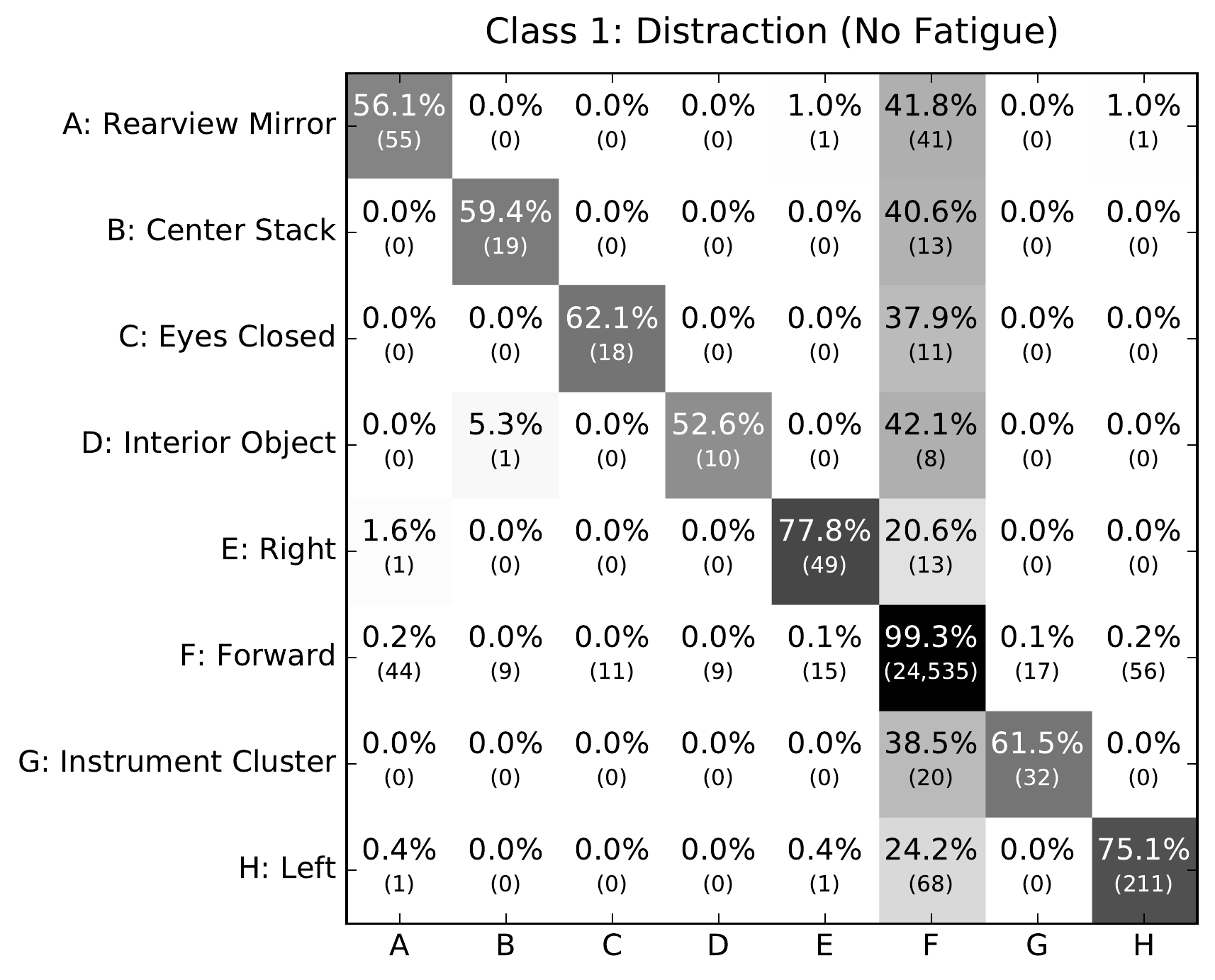}
      \transallfig{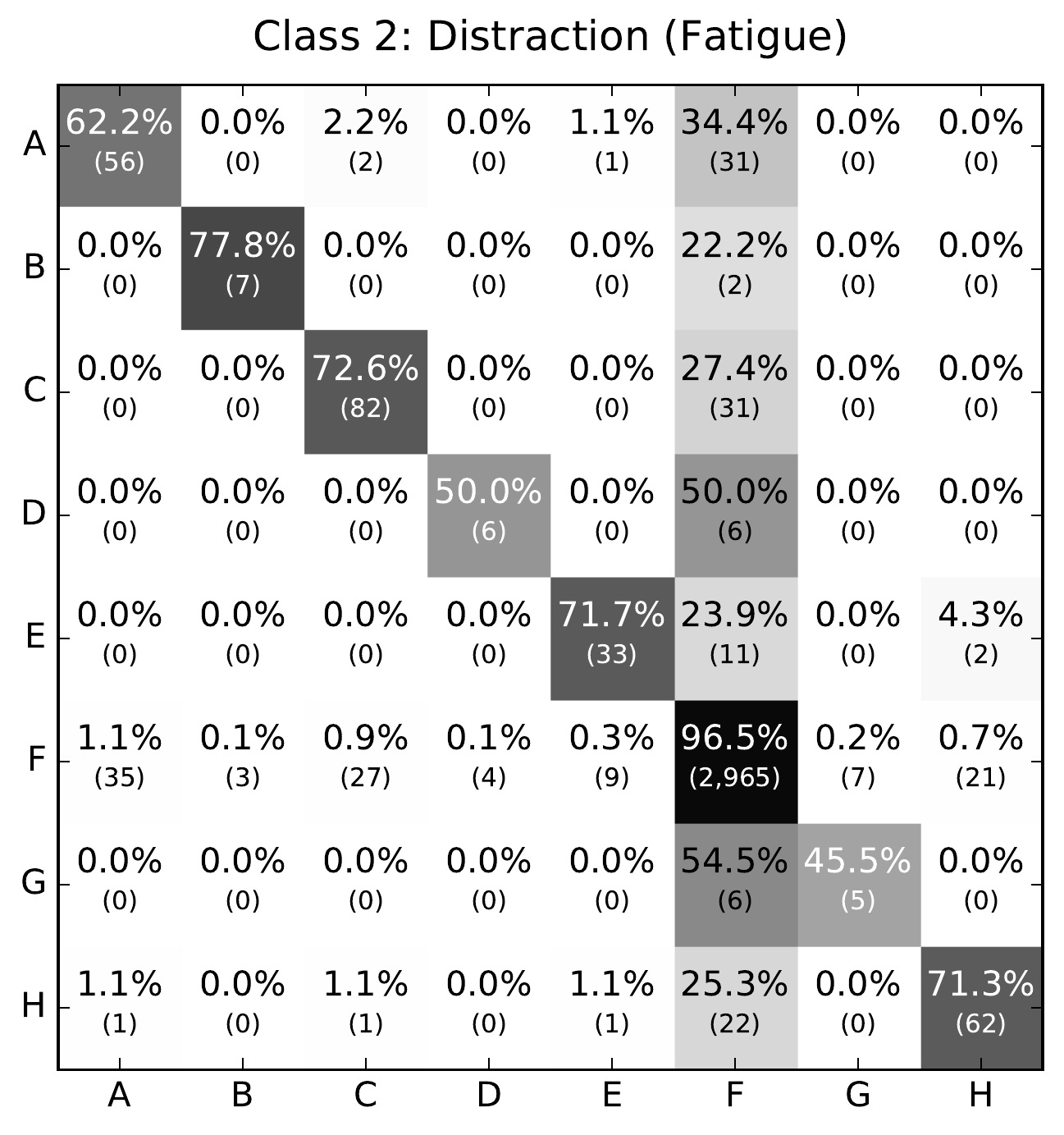}
      \transallfig{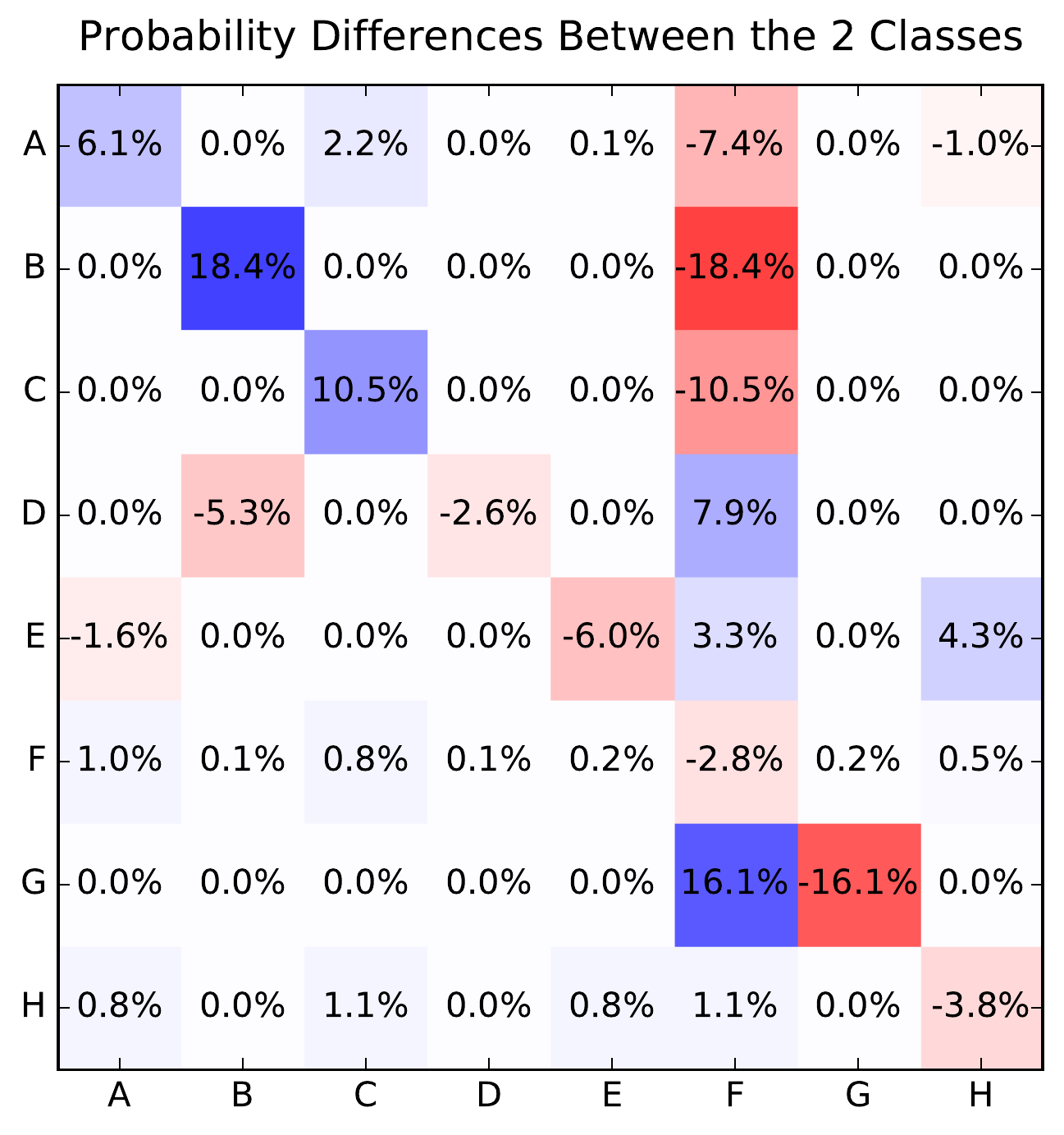}
      \caption{Transition matrices for Distraction (Fatigue). Classification accuracy: 80.4\%.}
      \label{fig:p25_distraction_03}
    \end{figure}
    \clearpage

    \begin{figure}[!h]
      \transallfig{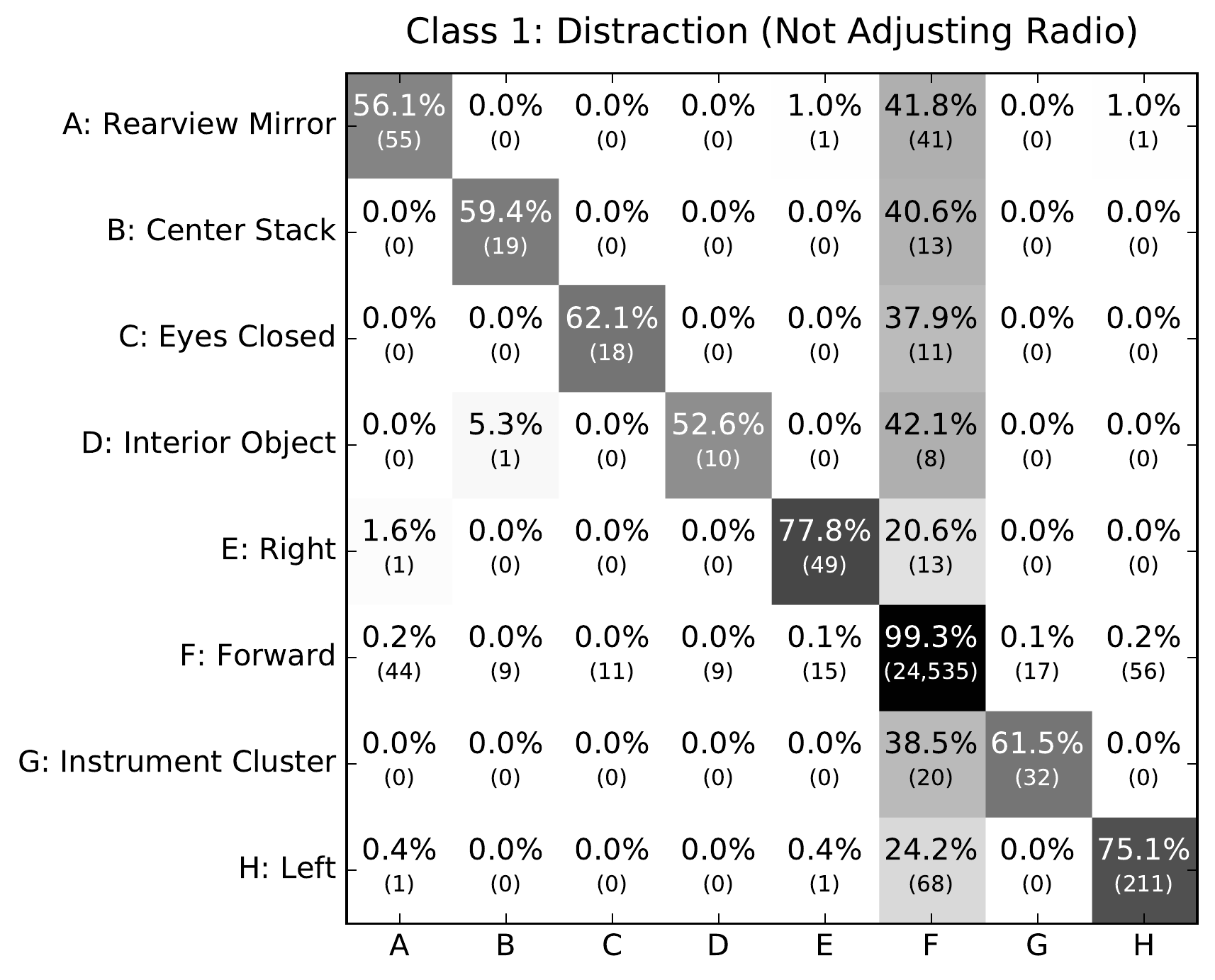}
      \transallfig{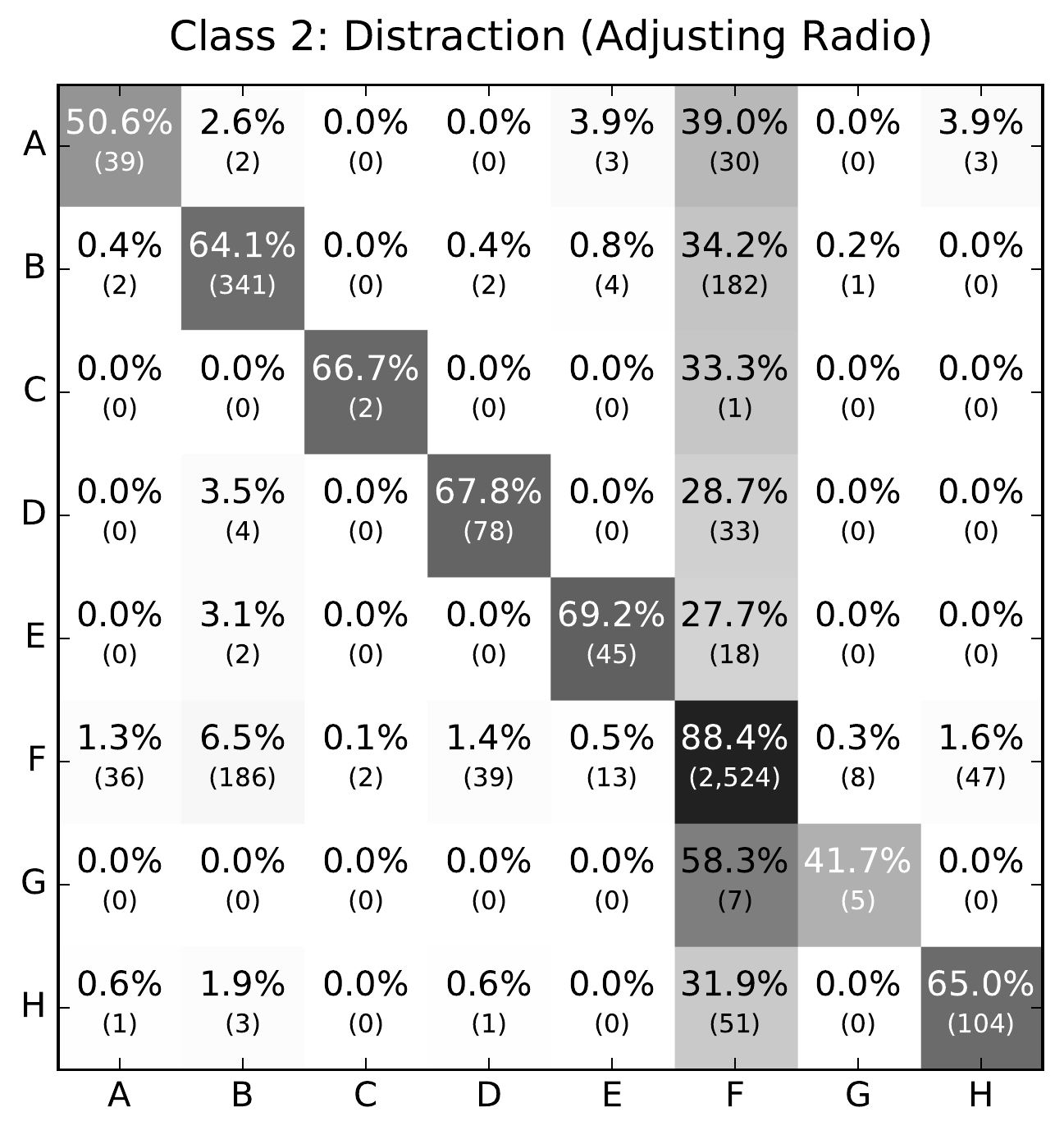}
      \transallfig{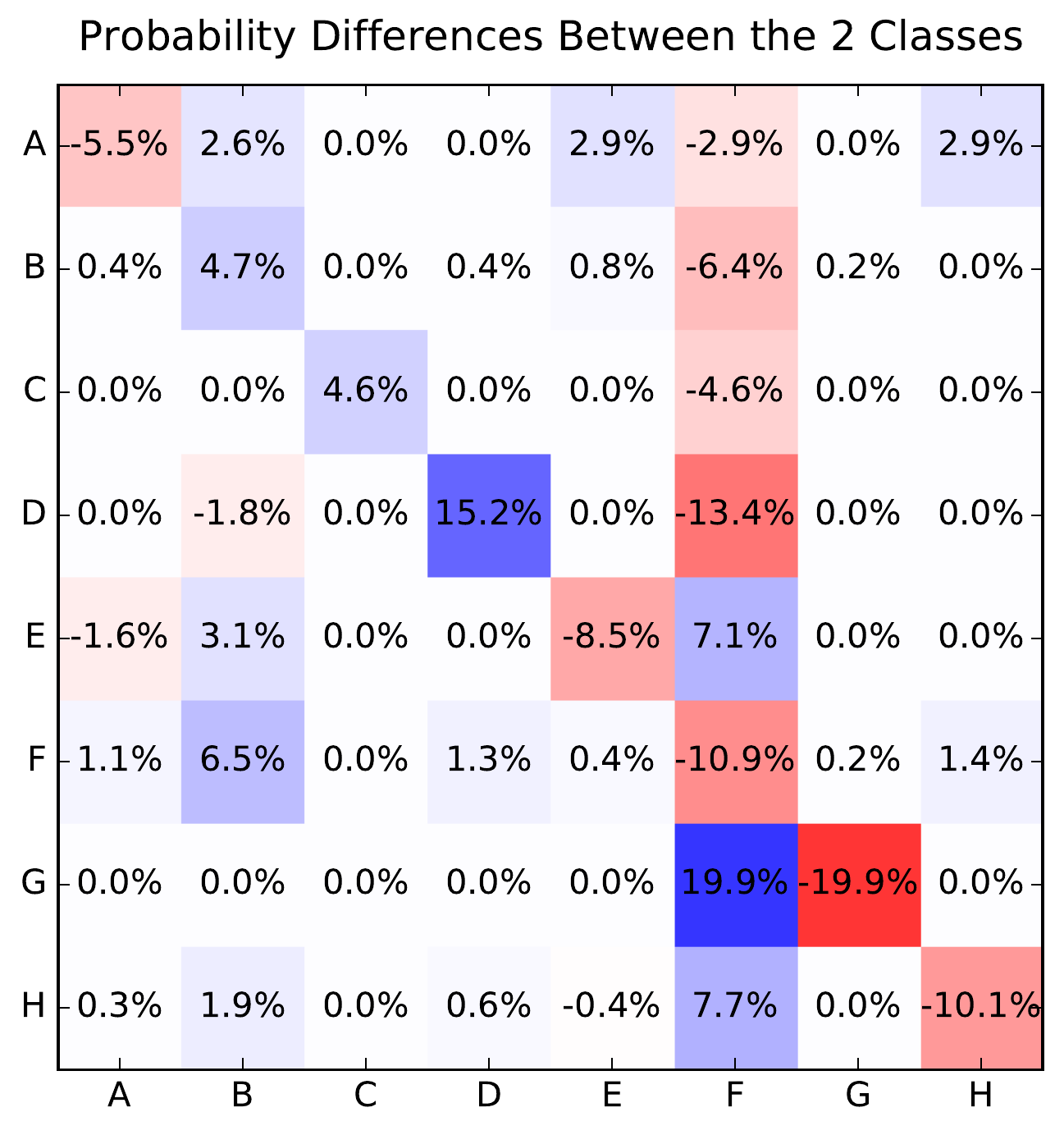}
      \caption{Transition matrices for Distraction (Adjusting Radio). Classification accuracy: 88.3\%.}
      \label{fig:p26_distraction_02}
    \end{figure}

\null
\vfill

\end{document}